\pdfoutput=1
\documentclass[10pt,twocolumn,letterpaper]{article}
\usepackage[pagenumbers]{cvpr}
\usepackage{microtype}

\usepackage{algorithm}      %
\usepackage{algorithmic}
\newcommand{\agro}{AgroGround}
\definecolor{cvprblue}{rgb}{0.21,0.49,0.74}
\usepackage[pagebackref,breaklinks,colorlinks,allcolors=cvprblue,bookmarksnumbered]{hyperref}
\makeatletter
\patchcmd{\@maketitle}{\vskip .375in}{\vskip 12pt}{}{\PackageError{main}{title patch 1 failed}{}}
\patchcmd{\@maketitle}{\vspace*{24pt}}{\vspace*{11.102pt}}{}{\PackageError{main}{title patch 2 failed}{}}
\makeatother

\title{\agro: What Grounding Fine-Tuning Changes\\ in Agricultural Vision-Language Models}
\author{Abdulla Alshehhi \qquad Zongyan Han \qquad Rao Muhammad Anwer\\
Mohamed bin Zayed University of Artificial Intelligence, Abu Dhabi, United Arab Emirates\\
{\tt\small \{abdulla.alshehhi, zongyan.han, rao.anwer\}@mbzuai.ac.ae}}

\ifdefined\pdfsuppressptexinfo \pdfsuppressptexinfo=-1 \fi
\begin{document}
\maketitle
\begin{abstract}
Large-scale agricultural grounding data can now be generated automatically, but how training on such data affects a vision-language model's localization, recognition, and refusal capabilities remains unclear. To study these effects, we introduce AgroGround, containing 794,850 automatically grounded agricultural instructions for training and a separate benchmark of 1,478 human-verified images for evaluation. Among these images, 500 are independently re-annotated, yielding an inter-annotator instance F1 of 0.789. We use controlled comparisons to isolate the effects of different training choices on localization, recognition, their joint correctness, and refusal.
Grounding-only fine-tuning reduces four-way recognition from 51.7\% to 29.2\%, with predictions showing a strong candidate-position bias. However, changing only the instruction format raises recognition to 72.7\% while maintaining similar grounding performance. Training with healthy-image negatives achieves 94.1\% refusal on healthy images, but only 28.6\% refusal of incorrect target names on positive images, showing that these two refusal behaviors do not necessarily improve together. Reinforcement learning reveals a further trade-off, improving localization by 0.014 F1 over a matched supervised control while reducing recognition and refusal. Across model families and held-out training sources, localization transfers more reliably than recognition and target-specific refusal. External benchmarks further show that the gains depend on the domain and task. Together, these findings show that grounding fine-tuning changes the balance between localization, recognition, and refusal rather than improving all three uniformly.
 
\end{abstract}

\section{Introduction}

\begin{figure}[t]
\centering
\includegraphics[width=\linewidth]{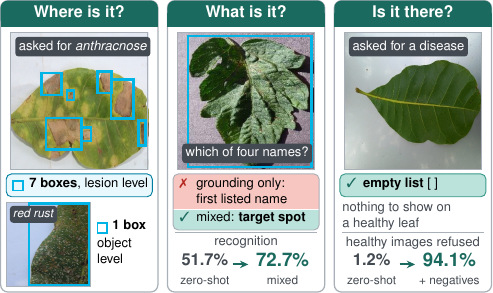}
\caption{\textbf{Grounded agricultural recognition requires three capabilities: localization, recognition, and target-presence assessment.} %
Examples are from our 2B models  (Quantitative results are in Tables~\ref{tab:main},~\ref{tab:ablation}).} \label{fig:teaser}
\end{figure}

Agricultural visual understanding requires answering three related questions: where is the relevant evidence, what target does it support, and is that target present at all (Figure~\ref{fig:teaser})? A grounded model should localize the relevant visual evidence, identify the corresponding agricultural target, and avoid producing evidence when the queried target is absent. Existing agricultural classifiers and visual question answering (VQA) systems primarily predict semantic labels \citep{mohanty2016plantvillage,liu2024cddm}, while detection, segmentation, and grounding systems typically localize a known target supplied in the query \citep{dubois2026textguidance,li2026agrovg,haghighat2026gvg}. These capabilities are often trained and evaluated separately.

Automatic grounding raises a new scientific question. Open-vocabulary detectors and segmentation models can convert the text labels of existing datasets into boxes and masks at large scale \citep{li2022glip,peng2024kosmos2,xiao2024florence2,xin2024dart,li2024pdcvld,xie2025medtrinity}, and recent agricultural corpora built with such pipelines reach hundreds of thousands of images \citep{boudiaf2026agriscope}.
Such pipelines reduce annotation cost but leave an important question unresolved: \textbf{What capabilities does a vision-language model (VLM) gain or lose when it is fine-tuned on automatically grounded supervision?} The answer is unclear: automatic labels are rarely compared with independent human annotation of measured reliability, while training recipes often change instruction format, negative supervision, and optimization simultaneously. Moreover, better localization need not improve recognition, refusal, or transfer.

We introduce \agro{}\footnote{Unrelated to the concurrent AgriGround dataset \citep{boudiaf2026agriscope}.} to answer this question for plant diseases and other agricultural targets. The training corpus is automatic: detector boxes for the targets named in eight VQA datasets become 794,850 examples at two granularities, individual lesions and whole affected objects, with healthy images as negatives (Figure~\ref{fig:data}). The benchmark is human-verified: it contains 1,478 images, separated from training by file and perceptual near-duplicate checks, and evaluates localization, recognition, their joint correctness, and refusal. A second annotator independently re-annotated 500 of them, half without detector proposals (instance F1 0.789, $\kappa$ 0.981 on target presence). Compared with existing agricultural localization resources, \agro{} combines language-conditioned training supervision, independently measured annotation agreement, and evaluation of both healthy-image and target-specific absence (Table~\ref{tab:compare}).

Our aim is not to propose a new multi-task objective or reinforcement learning (RL) algorithm. We use \agro{} to isolate how individual training choices redistribute these capabilities. Within each controlled comparison, we hold the relevant training setup fixed (Figure~\ref{fig:method}) and vary one ingredient at a time: the instruction format (the target name is given, or must be recognized among candidates), the negative supervision (healthy images versus incorrect target names), and the post-training stage, comparing RL, whose reward explicitly credits spatial coverage, against a supervised control matched in prompts, image resolution, and training exposure.

Three main findings emerge. First, grounding-only fine-tuning preserves localization but degrades recognition through a bias toward the first listed candidate, and changing only the instruction format restores and improves it. Second, healthy-image negatives teach refusal on healthy images but transfer only partially to incorrect target names, showing that the two refusals are distinct behaviors. Third, RL improves localization only slightly beyond a matched control and reduces recognition and refusal. The instruction-format effect replicates in a second model family and on external images. With held-out sources, localization and healthy-image refusal carry over, whereas the gains in recognition and in separating wrong from true names largely do not. External evaluation on PlantSeg, PlantDoc, AgroVG, and GroundingME shows that the gains remain domain- and task-specific.

Our contributions are threefold: (1) \agro{}, a training corpus of 794,850 examples and a human-verified benchmark of 1,478 images, including an independent re-annotation of 500 images;
(2) a controlled study of grounding fine-tuning that isolates instruction format, negative supervision, and RL under a fixed recipe; and (3) an analysis of transfer and failure modes across model families, held-out sources, and external benchmarks. Annotations, code, and predictions will be released.

\section{Related Work}
\label{sec:related}
\textbf{Agricultural grounding resources.} Existing agricultural multimodal corpora provide rich text supervision but little region-level annotation \citep{liu2024cddm,shinoda2025agrobench,gauba2025agmmu,dongre2025mirage,wen2025agrocot,li2025agromind,nguyen2026leafnet,boudiaf2026agrichat,sakib2026plantexpertvqa,mahmood2026agrichain}.
Expert-reviewed localization sets such as PlantSeg \citep{wei2024plantseg} and PD40 \citep{zhang2026pd40} have fixed classes and no language, while language-conditioned grounding data exists for pests, crops, and weeds \citep{yang2026pestscope,haghighat2026gvg}. AgroVG \citep{li2026agrovg} is an evaluation benchmark without a training split, and two recent corpora add training data: AgriGround \citep{boudiaf2026agriscope}, annotated automatically, and AgroOmni \citep{zhang2026agroomni}, whose question--answer pairs include localization. Table~\ref{tab:compare} compares these resources with \agro{}.

\begin{table}[t]
\centering
\footnotesize
\setlength{\tabcolsep}{3pt}
\resizebox{\linewidth}{!}{\begin{tabular}{llcclc}
\toprule
Resource & Size & Train & Lang. & Target absent & Agreement \\
\midrule
PD40 \citep{zhang2026pd40} & 80K images & yes & -- & healthy classes & $\kappa$ 0.87 \\
AgroVG \citep{li2026agrovg} & 10K pairs & -- & yes & other target family & -- \\
gRef-CW \citep{haghighat2026gvg} & 8K images & yes & yes & replaced or swapped expressions & -- \\
AgriGround \citep{boudiaf2026agriscope} & 504K images & yes & yes & absence questions & -- \\
AgroOmni \citep{zhang2026agroomni} & 107K images & yes & yes & -- & -- \\
\agro{} (ours) & 172K images & yes & yes & healthy image, wrong name & F1 0.79 \\
\bottomrule
\end{tabular}}
\caption{\textbf{Agricultural localization resources.} Lang.: the target is selected by language. Agreement: reported statistic between annotators (PD40: Cohen's $\kappa$ on categories; ours: instance F1). Our size counts training image files (104K unique).}
\label{tab:compare}
\end{table}

\textbf{What grounding fine-tuning changes.} Fine-tuning for grounding can cost other abilities \citep{wu2025flmm,wu2024sesame}, and losses tied to the answer format or to the task can be repaired from the data side \citep{chen2025sefe,li2026forgetting}. Models that choose among listed options are sensitive to option tokens and positions \citep{zheng2024mcq,zong2024permutations,xue2024pif}. These observations motivate our controlled comparison of grounding-only and candidate-based instruction formats while holding the underlying images and spatial supervision fixed.

\textbf{Refusal in grounding.} Generalized referring adds expressions without a target \citep{liu2023gres,he2023grec}. GroundingME \citep{li2025groundingme} finds that most of the 25 multimodal LLMs it evaluates reject none of its target-absent queries, and that negatives in training raise rejection in domain far more than out of domain. HalluSegBench \citep{li2026hallusegbench} separates label-driven from image-driven errors by swapping one or the other, and RC-GRPO \citep{yang2026rcgrpo} trains refusal with general-domain RL and penalizes refusing a present target. In agriculture the two cases are a healthy image and a wrong target name on a positive image, and we show on the same models that they are different abilities.
 
\textbf{RL for grounding.} RL with verifiable IoU-style rewards refines the boxes that VLMs emit as text \citep{liu2025visualrft,shen2025vlmr1,yu2025perceptionr1,zhan2026egm,peh2026hgrpo}. In agriculture, WeedExpert-R1 \citep{yang2026weedexpert} and AgroOmni apply GRPO \citep{shao2024deepseekmath} without a target-absent condition. \citet{lu2026whyrl} find supervised fine-tuning ahead in distribution and RL ahead out of distribution. We isolate what the RL stage adds with a supervised control that is matched in prompts, presentations, and image size.

\section{The \agro{} Corpus and Benchmark}
\label{sec:corpus}
\begin{figure*}[t]
\centering
\includegraphics[width=\linewidth]{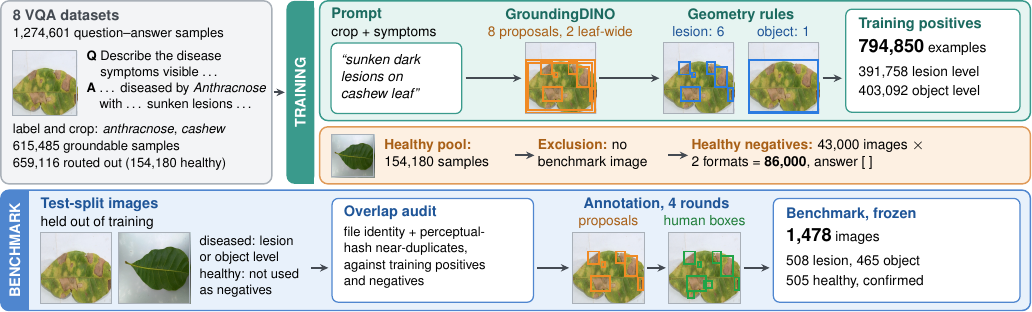}
\caption{\textbf{\agro{} Data Construction.} Training: rules convert detector proposals (orange) into lesion- or object-level examples (blue). Benchmark: human annotator corrects proposals into reference boxes (green) on images held out from training.} \vspace{-0.36cm}
\label{fig:data}
\end{figure*}
\subsection{Task Definition}
\label{sec:tasks}
Given an image and a text instruction, the model outputs a list of bounding boxes, each labeled with a target name, or an empty list. Two instruction formats share this output (Figures~\ref{fig:teaser} and~\ref{fig:method}). A \emph{known-target} instruction names the target and asks where it is. A \emph{candidate} instruction lists four targets, the true one and three distractors drawn preferentially from the same crop, in random order, and asks which one is present and where. Each instruction also fixes a granularity: individual lesions (\emph{lesion level}) or the affected leaf, fruit, or plant (\emph{object level}) \citep{dong2022annotation}. A healthy image is queried in either format with targets associated with its crop, and the correct output is an empty list. Targets are mainly diseases but also include pests, weeds, and other visually localizable agricultural entities. We use \emph{target} when referring to the task generically.

\subsection{Automatic Annotation of Training Data}
\label{sec:annotation}

\agro{} takes its training supervision from eight agricultural VQA datasets with 1,274,601 question--answer samples: CDDM, LeafNet, LeafBench, MIRAGE, AgroMind, AgroCoT, AgroBench, and AgMMU. The pipeline extracts a target label from each answer, generates boxes for it, and assigns each sample to one or both granularities (Figure~\ref{fig:data}).
 \textbf{Label extraction.} Source-specific handlers extract the target label and the host crop from each answer. Answers naming a visually localizable target are kept as \emph{groundable}: 615,485 samples, 97.6\% of which name a disease. Other answers are excluded, and the 154,180 healthy samples go to a separate pool for negatives. Labels are normalized to a taxonomy of 6,038 diseases, pests, and weeds, each with a description of its visual symptoms.

\textbf{Box generation.} The crop, the normalized label, and the symptom description form a prompt for GroundingDINO~\citep{liu2024groundingdino}. An audit found detector confidence only weakly related to box quality, so its boxes are then filtered by geometry and not by score \citep{mumuni2024samannotation}: degenerate and duplicate boxes and boxes outside 0.1\% to 95\% of the image area are removed. The thresholds were fixed before any evaluation data were collected. The kept boxes also decide the granularity. A lesion-level example keeps the small boxes, after boxes larger than half of the image are removed and non-maximum suppression is applied. An object-level example keeps one box around the affected leaf, fruit, or plant.

\textbf{Resulting corpus.} Of the 614,765 readable groundable samples, 579,260 (94.2\%) receive at least one valid box, which gives 794,850 examples: 391,758 at lesion level and 403,092 at object level, split 80/10/10 by source sample. The 171,580 training image files contain 104,150 perceptually unique images. In a stratified audit of 100 images, all boxes were correct in 62, and most remaining errors concerned granularity or missed instances.

\subsection{Human-Verified Benchmark}
\label{sec:eval}

\textbf{Construction.} The training boxes are automatic, so all benchmark results are computed against human boxes on images that training never saw. Benchmark images come from the held-out test split and must pass two rules: no file shared with the training split, and no perceptual near-duplicate of any training image, positive or negative (perceptual hash, flagged pairs verified at pixel level; Appendix~\ref{supp:benchmark}). One trained annotator labeled every image. The annotator saw the target name and the detector's boxes as editable proposals, drew a tight box around every instance, at lesion level when individual lesions are discernible and at object level otherwise, and removed images whose target could not be verified. Healthy images were confirmed visually.

The benchmark contains \textbf{1,478 images: 508 lesion-level, 465 object-level, and 505 healthy}, with 3,461 boxes (5.49 per lesion-level image) and 480 distinct target names. We call the 180 lesion-level images with five or more boxes \emph{dense}. Removing copies of training images shifts the benchmark toward field data: 463 of the 973 positive images (47.6\%) are in-the-wild photographs from MIRAGE and AgroBench, which supply 2.2\% of the training samples, and 62.2\% of the positive images show diseases, against 97.6\% in training.

\textbf{Agreement study.} To measure the reliability of these boxes, a second trained annotator, who never saw the first annotator's boxes, annotated a stratified subset of 500 benchmark images from all eight sources. Half of each stratum was annotated with the detector proposals shown, as in the original annotation, and half with no proposals. On the 400 positive images, instance F1 at IoU 0.5 between the two annotators is \textbf{0.789} (95\% CI [0.761, 0.816]), with 0.797 on the half with proposals and 0.781 on the half without. These are second-pass values: after a first pass (F1 0.755), the rule on lesions versus the whole leaf was made explicit and the half without proposals was annotated again. The annotators agree on whether a target is present (Cohen's $\kappa$ 0.981) and on lesion level versus object level ($\kappa$ 0.747). The models scored on this subset differ by at most 0.03 F1 between the two annotators, and ten of eleven paired F1 comparisons between models keep their sign against the second. Against the second annotator, the fine-tuned models lead GroundingDINO with raw prompts by 0.058 to 0.117 F1 on the half annotated without proposals (Appendix~\ref{supp:agreement}). All reported scores use the first annotator's boxes.

\textbf{Expert check.} A plant pathologist (M.Sc.\ in plant pathology, more than eight years as a plant disease and pest diagnostician) who took no part in the annotation judged a stratified random sample of 300 benchmark images, each with one name, without being told which names were the benchmark's own. The expert judged the benchmark's name visible on 81.8\% of 159 positive images [75.0, 87.0], the wrong name not visible on 81.8\% of 99 [73.1, 88.2], and the queried name not visible on 90.5\% of 42 healthy images [77.9, 96.2] (Wilson intervals). Re-scored on the images that the expert confirmed, the resolved differences in F1, recognition, and wrong-name refusal between the training configurations of Tables~\ref{tab:ablation} and~\ref{tab:refusal} keep their direction (Appendix~\ref{supp:expert}).

\textbf{Selection.} The reference boxes were fixed before any model was compared and were never revised toward model outputs, and no benchmark image was used for training or validation. A later audit removed two positive images that were copies of healthy training images (1,480 to 1,478; Appendix~\ref{supp:benchmark}). Three analysis choices were informed by benchmark inspection: the 12\% healthy-negative mixture, the step-170 RL checkpoint, and the seed-42 run used for paired image-level analyses and as the initialization for the RL experiments. Headline results for the main supervised model, RL, and its matched control are reported as mean $\pm$ standard deviation over three seeds.

\subsection{Evaluation Protocols and Metrics}
\label{sec:metrics}

\textbf{Metrics.} The \emph{known-target protocol} scores localization. Instances are matched greedily one-to-one at IoU 0.5 and pooled over images, which gives set precision, recall, and F1 (rules in Appendix~\ref{supp:protocols}), and region IoU is the IoU between the unions of the predicted and the reference boxes. The \emph{candidate protocol} scores recognition, the share of the 973 positive images on which the model names the correct target among four candidates (chance 25\%). Joint accuracy requires the correct name \emph{and} region IoU $\ge 0.5$.

\textbf{Refusal.} A refusal is an empty list. We measure it on three kinds of query: (i) the 505 healthy images, (ii) the 973 positive images queried with their true target, where a refusal is an error (\emph{false refusal}), and (iii) the same images queried in the same wording with the name of another target (\emph{wrong name}). The wrong name is another benchmark target of the image's crop where one exists (496 images, \emph{same-crop}), and otherwise a label drawn from all benchmark labels (477 images). To compare a plausible with an implausible wrong name on the same images, every image is also queried with a \emph{random} label that is not a target of its crop. \emph{Pair accuracy} is the share of positive images on which the model localizes the true target (region IoU $\ge 0.5$) and refuses the wrong one. To compare models without a decision rule, we report the AUROC of the \emph{empty-answer probability}, the probability that the model assigns to the empty list as its complete answer, for wrong against true names on the same images.

\textbf{Statistics.} Unless stated otherwise, intervals are 95\% percentile bootstraps over images (1,000 resamples, fixed seed). They reflect the sampling of benchmark images and not the variance between training runs. Differences between models are paired on the same images and computed before rounding. Seed variation is the mean and standard deviation of three training runs (main model, RL, and its control).

\section{Model and Training Configurations}
\label{sec:method}
\begin{figure*}[t]
\centering
\includegraphics[width=\linewidth]{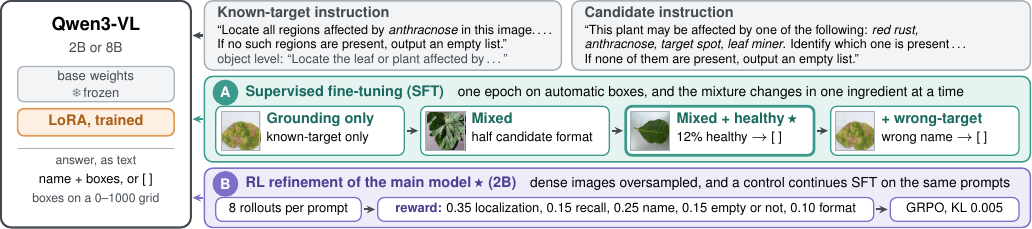}
\caption{\textbf{Controlled training configurations.} (A) Supervised mixtures vary one ingredient at a time: instruction format, healthy negatives, or wrong-target negatives. (B) RL starts from the main mixed+healthy model and is compared with continued SFT on matched prompts and image resolution.} %
\vspace{-0.35cm}
\label{fig:method}
\end{figure*}
\textbf{Model and output.} We fine-tune Qwen3-VL-Instruct \citep{qwen3vl2025} with LoRA \citep{hu2022lora} and the next-token loss (Figure~\ref{fig:method}). Every answer is text: a JSON list of boxes in the model's native 0--1000 coordinates, each labeled with the lowercased target name, or an empty list.

\textbf{Instruction format.} A known-target instruction already names the target, and the answer label repeats it, so a model can learn to localize without deciding what the image shows. The \emph{grounding-only} configuration uses this format alone. The \emph{mixed} configuration renders a random half of the positive examples in the candidate format, with the same images and boxes, so the new format adds only the choice of target.

\textbf{Negatives.} Positive examples never show an empty answer. The \emph{mixed + healthy} configuration adds images from the healthy pool, each queried in both formats with targets associated with its crop and answered with an empty list. A \emph{wrong-target} negative is a positive training image queried with another target, in 98\% of the cases one of its own crop, and also answered with an empty list.

\textbf{RL with a grounded reward.} Supervised fine-tuning gives no explicit credit for finding every instance. From the mixed + healthy model, we therefore apply GRPO \citep{shao2024deepseekmath} with the GSPO loss variant \citep{zheng2025gspo}, using the released code of \citet{zhan2026egm}. Sampled answers are scored against the automatic boxes with a reward that requires a valid format and credits localization (instance F1 and region IoU), instance recall, the correct candidate, and the right choice between an empty and a non-empty answer (weights in Figure~\ref{fig:method}; Appendix~\ref{supp:training}). The 18,750 rollout prompts come from the training split and oversample images with many boxes: half of the positive prompts have four or more boxes, against 3.2\% of the supervised positive examples with five or more.

\section{Experiments}
\label{sec:experiments}

\subsection{Experimental Setup}

\textbf{Training setup.} We fine-tune Qwen3-VL-2B and 8B. The model, the LoRA configuration, and the schedule are the same in every configuration, and no hyperparameter was searched on our data, so configurations differ only in their training mixture. Supervised runs take one epoch over the training split (635,977 examples before negatives) with images capped at 50,176 pixels, and evaluation applies no cap. RL and its control use images of up to 501,760 pixels.

\textbf{Negatives.} Healthy negatives use 21,500 or 43,000 healthy images with two negatives each, which is 6\% or 12\% of the training examples. These two candidate shares were fixed in advance. Wrong-target negatives are added to the 12\% mixture at 2.9\%, 5.6\%, or 10.6\% of the training examples, or replace the healthy negatives.

\textbf{Baselines and replications.} We compare with GroundingDINO in three configurations: raw names, the taxonomy prompts of the pipeline, and those prompts with the pipeline's geometry filters, which is the configuration that labeled the corpus (the \emph{teacher}). Further baselines are two zero-shot 7B reasoning-grounding systems, the zero-shot 2B and 8B bases, and, queried through an API without training, Qwen3-VL-235B and five closed models (Appendix~\ref{supp:baselines}). To test dependence on the model family, we also fine-tune GroundingDINO and Gemma~4 E4B~\citep{gemma4_2026} on a 20\% subsample of the corpus (Appendix~\ref{supp:baselines}).

\subsection{Controlled Findings}

\textbf{The fine-tuned model exceeds its teacher on lesion-level images.} The three GroundingDINO configurations land within 0.008 F1 of each other (Table~\ref{tab:main}). The main model (seed-42 run) exceeds each of them (paired +0.040 to +0.048, every interval above zero), and the RL model by +0.077 to +0.085. The margin comes from lesion-level images (F1 0.392 against 0.282 for the teacher), where the teacher's labels are incomplete (recall 0.194 at precision 0.517). On object-level images the teacher leads (0.701 against 0.613), and at IoU 0.75 GroundingDINO with taxonomy prompts leads (0.375 against 0.288, and 0.341 after RL). Fine-tuning helps where instances are many: on the 180 dense images F1 rises from 0.076 zero-shot to 0.316, and to 0.367 after RL, while it stays at 0.60 to 0.62 on the other images.
 
\textbf{Automatic boxes against expert labels.} Trained on 9,120 examples, the grounding-only recipe reaches F1 0.375 with PlantSeg's expert masks and 0.371 with a random subset of our boxes, whose sources the benchmark shares (paired $-$0.004 [$-$0.023, +0.015]; Appendix~\ref{supp:scale}). The full corpus (F1 0.425) adds 0.054 [0.036, 0.072] to the latter.
 
\textbf{A fine-tuned detector matches the VLM in F1, but not in the rest of the task.} GroundingDINO fine-tuned on a 20\% subsample of the same boxes reaches F1 0.466, against 0.445 for the mixed + healthy VLM trained on the same subsample (+0.021 [$-$0.002, +0.044] for the detector), and is level with the RL model (+0.011 [$-$0.010, +0.034] for RL). But it takes no granularity instruction, returns no box on 12.8\% of the positive images, and reaches 48.0\% recognition and 26.7\% joint accuracy when adapted to the candidate protocol, against 66.3\% and 41.4\% for that VLM.

\begin{figure}[t]
\centering
\includegraphics[width=\linewidth]{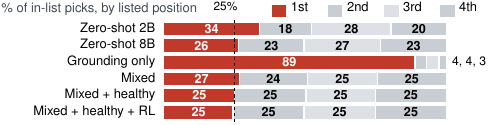}
\caption{\textbf{Position of the selected candidate among in-list predictions.} Grounding-only training picks the first listed one, whereas mixed candidate training removes it.} \vspace{-0.3cm}
\label{fig:positions}
\end{figure}

\textbf{Grounding-only training collapses recognition onto the first candidate.} The zero-shot 2B model names the right candidate on 51.7\% of the positive images (8B: 64.3\%). The grounding-only model, fine-tuned from the same base, scores \textbf{29.2\%}, barely above the 25\% chance level (Table~\ref{tab:ablation}). The position of its picks shows why: 88.8\% of its in-list answers are the first listed candidate (Figure~\ref{fig:positions}), against 33.8\% for the zero-shot model, and 77 answers name a disease that is not in the list. The bias is therefore induced by training, although the training task never lists candidates: the answer label always repeats the name in the instruction, which is consistent with a model that takes the name from the prompt and not from the image. Localization is intact: with the disease given, the model reaches F1 0.425 against 0.435 for the mixed model (paired +0.010 [$-$0.002, +0.022]). Recognition also falls when the four names are queried one at a time and the least refused one is taken: 43.0\% [40.0, 45.9], against 57.1\% zero-shot and 61.4\% for the mixed model (Appendix~\ref{supp:recognition}). 
A second model family shows the same dependence on the format. Gemma~4 E4B, trained with the same data recipe on a 20\% subsample, is right on 11.9\% of the images after grounding-only training, where 71.9\% of its answers list all four candidates, and on 67.5\% with mixed instructions. So do external images: on PlantDoc \citep{singh2020plantdoc} the grounding-only model scores 27.4\%, against 52.7\% for the mixed model and 44.1\% zero-shot.

\begin{table*}[t]
\centering\small
\setlength{\tabcolsep}{4pt}
\renewcommand{\arraystretch}{0.92}\resizebox{\linewidth}{!}{\begin{tabular}{lcccccccc}
\toprule
 & \multicolumn{2}{c}{Localization} & \multicolumn{2}{c}{Recognition} & \multicolumn{3}{c}{Refusal} & External \\
\cmidrule(lr){2-3}\cmidrule(lr){4-5}\cmidrule(lr){6-8}\cmidrule(lr){9-9}
Model & F1 [95\% CI] $\uparrow$ & Lesion F1 $\uparrow$ & Acc.\ (\%) $\uparrow$ & Joint (\%) $\uparrow$ & Healthy (\%) $\uparrow$ & False (\%) $\downarrow$ & Wrong (\%) $\uparrow$ & PlantSeg F1 $\uparrow$ \\
\midrule
GroundingDINO~\citep{liu2024groundingdino}\\
\quad Raw prompts & 0.400 [0.369, 0.436] & 0.323 & 27.5$^\dagger$ & 17.5$^\dagger$ & 0.0 & 0.1 & -- & 0.182 \\
\quad Taxonomy prompts & 0.395 [0.375, 0.418] & 0.358 & -- & -- & 0.0 & 0.0 & -- & 0.183 \\
\quad Pipeline configuration (teacher) & 0.392 [0.362, 0.427] & 0.282 & -- & -- & 20.2 & 0.5 & -- & 0.161 \\
\quad Fine-tuned on our boxes (20\%) & 0.466 [0.436, 0.498] & 0.423 & 48.0$^\dagger$ & 26.7$^\dagger$ & 84.6 & 12.8 & -- & -- \\
VisionReasoner-7B \citep{liu2026visionreasoner} & 0.369 [0.338, 0.402] & 0.302 & 40.7$^\dagger$ & 26.8$^\dagger$ & 0.2 & 0.5 & -- & -- \\
Seg-Zero-7B \citep{liu2025segzero} & 0.247 [0.219, 0.278] & 0.124 & -- & -- & 0.0 & 0.0 & -- & -- \\
Qwen3-VL-2B, zero-shot & 0.319 [0.290, 0.354] & 0.205 & 51.7 & 34.3 & 1.2 & 0.0 & 0.0 & 0.164 \\
Qwen3-VL-8B, zero-shot & 0.379 [0.342, 0.420] & 0.294 & 64.3 & 44.5 & 30.9 & 2.5 & 15.5 & 0.215 \\
Qwen3-VL-235B, zero-shot$^\S$ & 0.411 [0.376, 0.449] & 0.341 & 73.6 & \textbf{52.7} & 56.4 & 1.4 & 25.8 & -- \\
\midrule
Ours 2B, mixed + healthy (mean $\pm$ std, three runs) & 0.436 $\pm$ 0.004 & 0.388 $\pm$ 0.003 & 73.7 $\pm$ 1.1 & 43.5 $\pm$ 1.2 & 94.1 $\pm$ 0.9 & 4.9 $\pm$ 1.2 & 28.6 $\pm$ 1.5 & -- \\
\quad seed-42 run & 0.440 [0.412, 0.474] & 0.392 & 74.7 & 44.8 & 95.0 & 4.5 & \textbf{30.2} & 0.231 \\
Ours 2B, + RL (mean $\pm$ std, three runs) & 0.469 $\pm$ 0.012 & 0.432 $\pm$ 0.013 & 71.0 $\pm$ 1.5 & 43.6 $\pm$ 1.4 & 90.8 $\pm$ 2.6 & 2.0 $\pm$ 1.1 & 15.3 $\pm$ 6.3 & -- \\
\quad first RL run & \textbf{0.477} [0.447, 0.509] & \textbf{0.439} & 69.3 & 43.1 & 87.7 & 0.9 & 8.1 & \textbf{0.282} \\
Ours 8B, mixed + healthy & 0.429 [0.399, 0.460] & 0.369 & \textbf{78.2} & 48.6 & \textbf{96.0} & 2.8 & 27.9 & 0.237 \\
\midrule
Gemini 4 Argon$^\S$ & 0.499 [0.466, 0.533] & 0.452 & 80.8 & 60.5 & 92.5 & 10.2 & 77.4 & -- \\
GPT-6 Astra$^\S$ & 0.520 [0.481, 0.562] & 0.473 & 78.8 & 56.0 & 43.4 & 1.2 & 26.5 & -- \\
GPT-6.1 Sol$^\S$ & 0.531 [0.493, 0.571] & 0.480 & 79.5 & 57.7 & 39.0 & 0.9 & 23.5 & -- \\
Claude Opus 5.5$^\S$ & 0.561 [0.531, 0.593] & 0.535 & 80.2 & 64.9 & 92.7 & 7.2 & 75.1 & -- \\
Claude Fable 5.2$^\S$ & 0.597 [0.570, 0.625] & 0.578 & 84.8 & 70.7 & 89.1 & 4.6 & 67.8 & -- \\

\bottomrule
\end{tabular}}
\caption{\textbf{Main results.} Localization and refusal: known-target protocol. Recognition: candidate protocol. Healthy, False, Wrong: refusals on healthy images, of the true target, and of a wrong name. Bold: best open model. $^\dagger$Adapted protocol (Appendix~\ref{supp:baselines}). $^\S$Zero-shot through an API in November 2026 (third block: closed models).}
\label{tab:main}
\end{table*}

\begin{table*}[t]
\centering\small
\setlength{\tabcolsep}{4pt}
\renewcommand{\arraystretch}{0.92}\resizebox{0.8\linewidth}{!}{\begin{tabular}{lccccccc}
\toprule
Configuration & F1 $\uparrow$ & Lesion F1 $\uparrow$ & Recog.\ (\%) $\uparrow$ & Pos.-1 (\%) & Joint (\%) $\uparrow$ & Healthy ref.\ (\%) $\uparrow$ & False ref.\ (\%) $\downarrow$ \\
\midrule
Zero-shot Qwen3-VL-2B & 0.319 & 0.205 & 51.7 & 33.8 & 34.3 & 1.2 & 0.0 \\
Grounding only & 0.425 & 0.366 & 29.2 & 88.8 & 19.2 & 0.0 & 0.0 \\
Mixed & 0.435 & 0.377 & 72.7 & 26.7 & 43.3 & 0.2 & 0.0 \\
Mixed + healthy (6\%) & 0.433 & 0.383 & 73.1 & 25.8 & 44.3 & 93.3 & 5.1 \\
Mixed + healthy (12\%) & 0.440 & 0.392 & 74.7 & 25.1 & 44.8 & 95.0 & 4.5 \\
Mixed + healthy + RL & 0.477 & 0.439 & 69.3 & 24.6 & 43.1 & 87.7 & 0.9 \\
Mixed + healthy + continued SFT (control) & 0.458 & 0.410 & 73.7 & 24.3 & 45.2 & 96.6 & 5.1 \\
\midrule
Final RL checkpoint (step 292) & 0.467 & 0.440 & 67.2 & 24.7 & 39.0 & 89.3 & 6.4 \\
\bottomrule
\end{tabular}}
\caption{\textbf{Ablation on the 2B model.} Protocols as in Table~\ref{tab:main}. RL and continued-SFT entries correspond to the first matched runs used for controlled ablations; the three-run comparison is reported in Table 2 and Section 5.2. Pos.-1: in-list answers on the first candidate.}
\label{tab:ablation}
\end{table*}

\textbf{Changing only the instruction format restores recognition among candidates.} In the mixed configuration half of the training examples are rewritten in the candidate format, while answers, split, and hyperparameters are unchanged. Recognition rises from 29.2\% to \textbf{72.7\%} [69.6, 75.4] and joint accuracy from 19.2\% to 43.3\% (Table~\ref{tab:ablation}). The answers no longer follow the position: 26.7\% of in-list picks are the first listed candidate, and off-list answers drop from 77 to 7. Two controls qualify what the model reads. With the image tiles shuffled, recognition stays at 69.8\%, so the global layout is not needed. With the image blanked, the mixed model is still right on 34.0\% [31.1, 37.2] of the queries, so up to nine points above chance can come from the candidate lists alone. The 8B model recognizes more than the 2B (78.2\% against 74.7\%, paired +3.5 points [+0.6, +6.1]) at a similar F1 ($-$0.011 [$-$0.027, +0.005]). Without a candidate list the 2B models do not name the disease (lenient match at most 4.6\%), so we evaluate recognition with candidates.

\textbf{The model refuses healthy images, but wrong names only in part.} With 12\% healthy negatives the model returns an empty list on \textbf{94.1\%} of the healthy benchmark images (mean of three training runs, $\pm$0.9; 95.0\% [93.0, 96.8] for seed 42) and on 4.9\% ($\pm$1.2; 4.5\%) of the positive images queried with their true target (Table~\ref{tab:main}). For seed 42, instance F1 does not fall (0.440 against 0.435 without negatives), but region IoU does (0.616 to 0.578, paired $-$0.039 [$-$0.053, $-$0.027]). Gemma~4 E4B refuses 0.0\% of healthy images without negatives and 94.7\% with them, at a cost of 0.025 F1 [0.012, 0.039]. The 12\% model is our main model. Tables~\ref{tab:ablation}--\ref{tab:external} show its seed-42 run, the starting point for RL and the reference of the paired comparisons.

Refusing a wrong name is a different ability. On the wrong-name queries of Section~\ref{sec:metrics}, the main model returns an empty list on only \textbf{28.6\%} ($\pm$1.5; 30.2\% [27.5, 33.1] for seed 42), and its pair accuracy is 18.1\% ($\pm$1.3; 19.5\%). The empty-answer probability shows the gap without a decision rule: its AUROC is about 0.99 for healthy against positive images, but 0.812 for wrong against true names ($\pm$0.010; 0.823 [0.807, 0.839]). A plausible wrong name is the hard case. On the 496 images with a same-crop alternative, the seed-42 run refuses the same-crop name on 25.4\% and a random name on 36.7\% (AUROC 0.804 against 0.893, paired +0.089 [+0.069, +0.111]). In a wording that differs from that of the true-name query, it refuses 15.0\% [12.8, 17.4] of the wrong names, so we ask both names in one wording.

\textbf{Closed and larger models leave headroom in F1 and split on refusal.} Without training, five closed models reach F1 0.499--0.597 (Table~\ref{tab:main}), against at most 0.477 for open models and 0.789 between two annotators (Section~\ref{sec:eval}). Refusal splits them: three refuse 89.1--92.7\% of healthy images and 67.8--77.4\% of wrong names, two only 39.0--43.4\% and 23.5--26.5\%. All five and the 235B open model refuse the wrong name less often for the 496 same-crop names (7.1--64.9\%) than for the other 477 (40.7--92.0\%).

\textbf{Wrong-name negatives raise this refusal at a cost, and mainly for what they show.} At 2.9\% to 10.6\%, they raise wrong-name refusal to 52--62\% and pair accuracy to 30--32\% (Table~\ref{tab:refusal}), but the true name is then refused on 10--16\% of the images, F1 falls by 0.028--0.048, and recognition by 2.7--7.2 points. On the 496 images, the AUROC of the three doses rises by 0.049--0.061 for same-crop names and changes by $-$0.011 to +0.021 for random names. For same-crop names it rises from 0.851 to 0.906--0.929 on the six sources that supply 98\% of the training samples, but not on the in-the-wild MIRAGE and AgroBench images (0.649 to 0.625--0.640). A second seed of the 5.6\% dose agrees.

Two alternatives need no retraining. A cross-fitted threshold on the main model's empty-answer probability reaches pair accuracy 29.4\% [26.7, 32.5] at F1 0.431 (26.6\% and 20.6\% for the other two seeds), against 29.9--31.7\% at F1 0.392--0.412 after retraining. A name check reaches pair accuracy 37.7\% but refuses 26.2\% of the true names (Table~\ref{tab:refusal}).

\begin{table}[t]
\centering
\footnotesize
\setlength{\tabcolsep}{3pt}
\resizebox{\linewidth}{!}{\begin{tabular}{lcccccccc}
\toprule
 & \multicolumn{3}{c}{Empty answers (\%)} & & \multicolumn{3}{c}{AUROC} & \\
\cmidrule(lr){2-4}\cmidrule(lr){6-8}
Training negatives & Healthy & True & Wrong & Pair & All & Same & Rand. & F1 \\
\midrule
None (mixed) & 0.2 & 0.0 & 0.0 & 0.0 & 0.666 & 0.647 & 0.697 & 0.435 \\
Healthy 12\% (main) & 95.0 & 4.5 & 30.2 & 19.5 & 0.823 & 0.804 & 0.893 & 0.440 \\
\quad + wrong-target 2.9\% & 97.0 & 10.4 & 52.1 & 29.9 & 0.854 & 0.865 & 0.914 & 0.412 \\
\quad + wrong-target 5.6\% & 96.8 & 13.2 & 59.7 & 31.7 & 0.844 & 0.853 & 0.899 & 0.409 \\
\quad + wrong-target 10.6\% & 95.8 & 15.7 & 62.0 & 30.3 & 0.826 & 0.865 & 0.882 & 0.392 \\
Wrong-target 11.9\% only & 84.8 & 13.8 & 61.9 & 31.6 & 0.846 & 0.872 & 0.910 & 0.395 \\
\midrule
Main + threshold$^\ast$ & 97.6 & 9.9 & 53.9 & 29.4 & -- & -- & -- & 0.431 \\
Main + name check$^\dagger$ & -- & 26.2 & 77.9 & 37.7 & -- & -- & -- & 0.373 \\
\midrule
Main + RL & 87.7 & 0.9 & 8.1 & 6.7 & 0.720 & 0.707 & 0.794 & 0.477 \\
Main + continued SFT & 96.6 & 5.1 & 36.5 & 22.2 & 0.828 & 0.808 & 0.887 & 0.458 \\
\bottomrule
\end{tabular}}
\caption{\textbf{Two kinds of refusal.} Empty answers on healthy images, and on positive images queried with the true or a wrong name. Main: seed-42 run. Pair: pair accuracy (\%). AUROC of the empty-answer probability for wrong against true names: on all 973 positive images (All) and, on the 496 with a same-crop alternative, for a same-crop (Same) and a random (Rand.) name. $^\ast$Cross-fitted threshold on the empty-answer probability. $^\dagger$An answer is kept only if the model picks the queried name among four candidates; for a wrong name, the list excludes the true target.}
\label{tab:refusal}
\end{table}

\begin{figure}[t]
\centering
\includegraphics[width=\linewidth]{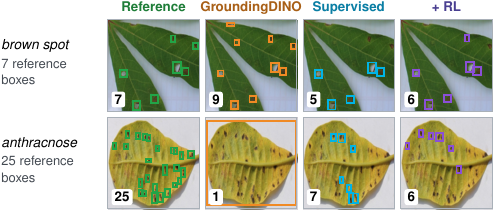}
\caption{\textbf{Known-target grounding on two dense benchmark images.} Columns compare the human reference, GroundingDINO with raw prompts, supervised fine-tuning and RL. Numbers denote box counts.} \vspace{-0.4cm}
\label{fig:qual}
\end{figure}

\textbf{RL improves instance coverage beyond a matched supervised control, and costs recognition and refusal.} The supervised model finds too few instances (Figure~\ref{fig:qual}): it returns 2.2 boxes per lesion-level image, against 5.49 in the reference. From the main model, RL raises F1 from 0.440 to \textbf{0.477} (paired +0.037 [+0.020, +0.053]), and lesion-level F1 from 0.392 to 0.439 (Table~\ref{tab:ablation}). Part of this gain is additional training. A matched control continues supervised training from the same model on the same 18,750 rollout prompts, as often as RL and at the same image size. The control reaches F1 0.458 (+0.017 [+0.003, +0.031]) and keeps recognition ($-$1.0 points [$-$2.4, +0.4]). Over three runs each, RL adds +0.014 F1 [+0.003, +0.025] to the control (0.469 $\pm$ 0.012 against 0.456 $\pm$ 0.005; +0.028 [+0.013, +0.045] on dense images), and it loses 3.4 recognition points [$-$4.9, $-$1.8] and 6.0 points of healthy-image refusal [$-$8.0, $-$4.3], while joint accuracy is not resolved ($-$1.5 [$-$3.4, +0.4]). Every RL run is below every control run in recognition and healthy-image refusal, but one of the three is level with the control in F1 (0.455), and at IoU 0.75 the two are level (0.334 and 0.343). Tables~\ref{tab:ablation}--\ref{tab:external} and all other comparisons use the first runs. At its final step (Table~\ref{tab:ablation}), the first RL run is not resolved against the control in F1 (+0.009 [$-$0.006, +0.024]) and is 6.5 recognition points below it. With uniform rollout sampling RL gives no gain, so its gain appears to depend on oversampling images with many boxes, and a reward that weights recognition more does not remove the cost. No RL configuration exceeds its supervised starting point in joint accuracy (Appendix~\ref{supp:rl}).
RL also weakens the refusal of wrong names (Table~\ref{tab:refusal}). The seed-42 run refuses 30.2\% of them, the three RL runs 8.1--19.6\%, and the control runs 31.9--36.5\%. The AUROC for wrong against true names falls from 0.823 to 0.720--0.780 (first run: paired $-$0.103 [$-$0.116, $-$0.090]) and lies between 0.686 and 0.756 in the other RL configurations (Appendix~\ref{supp:rl}), while the control keeps it (0.824--0.828). The reward gives credit for a non-empty answer on every positive rollout image and never shows a wrong name, which is consistent with this loss.
\subsection{Held-Out Sources and External Benchmarks}
\label{sec:external}

\textbf{Held-out sources.} To measure transfer beyond the training sources, we retrain the main model seven times, each time without one source, and score the benchmark images of that source (1,473 in all) with the model that never saw it. The comparison is the seed-42 run trained on all sources, on the same images. Localization and the refusal of healthy images carry over: F1 is 0.424 against 0.440 (paired $-$0.016 [$-$0.031, $-$0.001]; zero-shot 0.319), about seven eighths of the gain over zero-shot, and healthy images are refused on 91.4\% against 95.2\%. Recognition keeps about one fifth of its gain: 56.1\% against 74.9\% (paired $-$18.7 [$-$21.6, $-$15.7]; zero-shot 51.6\%), and joint accuracy is 35.8\% against 44.9\% ($-$9.1 [$-$11.8, $-$6.2]; zero-shot 34.2\%). The true name is refused on 12.2\% of the images against 4.4\% and a wrong name on 39.1\% against 30.1\%, but the two names are separated about as well as zero-shot: the AUROC is 0.725 against 0.824 (paired $-$0.098 [$-$0.112, $-$0.086]; zero-shot 0.706). The recognition, joint, false-refusal, and AUROC numbers of Tables~\ref{tab:main}--\ref{tab:refusal} are therefore in-source numbers.

\textbf{PlantSeg.} On this expert-reviewed set \citep{wei2024plantseg} (2,165 of its test images; masks converted to boxes) instance F1 transfers: the RL model reaches 0.282 against 0.161--0.183 for GroundingDINO, 0.164 and 0.215 for the zero-shot models (Table~\ref{tab:main}), and 0.251 for a 2B model trained on PlantSeg's own masks (paired +0.031 [+0.018, +0.042]; supervised 0.231), while the region IoU of our models (0.367--0.463) stays below that of the zero-shot models (0.497 and 0.547).
 
\textbf{PlantDoc.} On these web photographs (986 images after removing near-duplicates of training files; Table~\ref{tab:external}) the fine-tuned models localize worse than their zero-shot bases (2B: paired $-$0.099 [$-$0.135, $-$0.066]; 8B: $-$0.121 [$-$0.158, $-$0.086]). They answer with one box per image, as the object-level training examples do, and lose on images with several leaves. The refusal of healthy images largely transfers (68.8\% against 0.5\% zero-shot, and 41.8\% after RL). Candidate recognition rises at 2B only (+7.9 points [1.6, 14.8]), as does joint accuracy (paired +9.5 [3.7, 15.5]).
 
\textbf{AgroVG.} On its task T1 \citep{li2026agrovg} (1,288 test queries from six target families) the main model is at most level with its base (paired +0.008 [$-$0.012, +0.027]; other seeds $-$0.009, $-$0.045) and the 8B model is 0.123 below [$-$0.149, $-$0.099]. The three 2B runs gain on present targets (+0.037 to +0.054) and lose on absent ones, where accuracy follows the training mixture: 0.673 zero-shot, 0.010 with grounding only, 0.156--0.523 with healthy negatives, 0.633--0.925 with wrong-name negatives, and 0.131--0.296 after RL (control: 0.417--0.442). The 10.6\% wrong-name model gains on both (overall +0.044 [+0.025, +0.061]). Two of the other three return no box on 47--52\% of the present targets (Appendix~\ref{supp:external}).

\textbf{General-domain grounding.} On GroundingME \citep{li2025groundingme} (1,005 samples) accuracy falls at 2B (paired $-$12.5 points [$-$14.9, $-$10.1]) and at 8B ($-$16.4 [$-$19.0, $-$13.9]).

\begin{table}[t]
\centering
\footnotesize
\setlength{\tabcolsep}{3pt}
\resizebox{\linewidth}{!}{\begin{tabular}{lccccc}
\toprule
 & \multicolumn{3}{c}{2B} & \multicolumn{2}{c}{8B} \\
\cmidrule(lr){2-4}\cmidrule(lr){5-6}
Evaluation & Zero-shot & Ours & + RL & Zero-shot & Ours \\
\midrule
PlantDoc, F1 & 0.512 & 0.413 & 0.449 & 0.570 & 0.450 \\
PlantDoc, healthy refused (\%) & 0.5 & 68.8 & 41.8 & 51.5 & 71.4 \\
PlantDoc, recognition (\%) & 44.1 & 52.0 & 48.0 & 57.5 & 50.1 \\
PlantDoc, joint (\%) & 28.8 & 38.3 & 36.9 & 46.9 & 37.6 \\
AgroVG T1, All@0.5 & 0.248 & 0.256 & 0.226 & 0.329 & 0.205 \\
AgroVG T1, target absent & 0.673 & 0.523 & 0.131 & 0.889 & 0.141 \\
GroundingME, Acc@0.5 (\%) & 21.2 & 8.7 & 8.4 & 31.0 & 14.6 \\
\bottomrule
\end{tabular}}
\caption{\textbf{External benchmarks.} Ours: main model. PlantDoc: 431 diseased, 555 healthy; object-level wording. AgroVG T1: official protocol; target absent: accuracy on the 199 such queries.}
\label{tab:external}
\end{table}

\section{Limitations and Conclusion}
\textbf{Limitations.} The training boxes are automatic and incomplete at lesion level, and the RL reward is computed on them. The reference boxes come from one annotator who saw the detector's proposals, and the second annotator covers 500 images. Target names are inherited from the sources and wrong names are drawn automatically. One expert judged both on 300 images. Recognition is measured among four candidates. The main model, RL, and its control have three runs each. The second-family replications and the held-out folds are single runs. Each API model was queried once. Seven of the eight sources contain public evaluation sets, so models trained on \agro{} are contaminated for them.
\textbf{Conclusion.} \agro{} pairs an automatic grounding corpus with a human-verified benchmark to measure what fine-tuning on such a corpus changes. Trained on it, a 2B model reaches the localization F1 of a zero-shot 235B open model. The instruction format decides recognition, healthy negatives teach little refusal of wrong names, and RL mainly moves the boxes, at a cost in recognition and refusal. The training mixture and a benchmark with both kinds of refusal deserve as much attention as the learning rule.

\clearpage\subsubsection*{LLM Usage Statement}
We used Claude (Anthropic) for feedback on methodology and experimental design, for implementing data-processing, evaluation, and figure code, for polishing text across the paper, for finding and summarizing related literature, and for brainstorming. The research questions, the dataset design, and the conclusions are the authors' own, and the authors ran all experiments. All AI-assisted help and text were reviewed by the authors, every reported number was verified against the prediction files and seeded scripts underlying the results, and all statistics were checked.

{
    \small
    \bibliographystyle{ieeenat_fullname}
    \bibliography{refs}
}

\clearpage
\twocolumn[\centering\vspace*{24pt}{\Large\textbf{Appendix}}\par\vspace{33.778pt}]
\appendix
\setcounter{table}{0}\setcounter{figure}{0}\setcounter{algorithm}{0}
\def\theHtable{S\arabic{table}}\def\theHfigure{S\arabic{figure}}\def\theHalgorithm{S\arabic{algorithm}}
\microtypesetup{activate=false}
\setlength{\abovecaptionskip}{10pt}
\hyphenation{Grounding-DINO Grounding-ME AUROC}
\renewcommand{\thetable}{S\arabic{table}}
\renewcommand{\thefigure}{S\arabic{figure}}
\setcounter{topnumber}{4}\setcounter{dbltopnumber}{4}\setcounter{totalnumber}{6}
\renewcommand{\topfraction}{0.9}\renewcommand{\dbltopfraction}{0.9}
\renewcommand{\textfraction}{0.08}\renewcommand{\floatpagefraction}{0.8}\renewcommand{\dblfloatpagefraction}{0.8}
\makeatletter
\renewenvironment{table}[1][]{\@float{table}[!tbp]}{\end@float}
\renewenvironment{table*}[1][]{\@dblfloat{table}[!tbp]}{\end@dblfloat}
\renewenvironment{figure}[1][]{\@float{figure}[!tbp]}{\end@float}
\renewenvironment{figure*}[1][]{\@dblfloat{figure}[!tbp]}{\end@dblfloat}
\setlength{\@fptop}{0pt}\setlength{\@fpsep}{14pt}\setlength{\@fpbot}{0pt plus 1fil}
\setlength{\@dblfptop}{0pt}\setlength{\@dblfpsep}{14pt}\setlength{\@dblfpbot}{0pt plus 1fil}
\makeatother

\noindent\textbf{Contents.} Each section gives the details of a part of the paper.

\smallskip
\noindent{\footnotesize\setlength{\tabcolsep}{5pt}%
\begin{tabular}{@{}llll@{}}
\toprule
 & Section & In the paper & Page \\
\midrule
\ref{supp:corpus} & Corpus Details & Section~3.2 & \pageref{supp:corpus} \\
\ref{supp:benchmark} & Benchmark Details & Section~3.3 & \pageref{supp:benchmark} \\
\ref{supp:agreement} & Agreement Study & Section~3.3 & \pageref{supp:agreement} \\
\ref{supp:expert} & Expert Check & Section~3.3 & \pageref{supp:expert} \\
\ref{supp:protocols} & Evaluation Protocols and Metrics & Section~3.4 & \pageref{supp:protocols} \\
\ref{supp:training} & Training Details & Sections~4 and~5.1 & \pageref{supp:training} \\
\ref{supp:baselines} & Baselines and Replications & Section~5.1 & \pageref{supp:baselines} \\
\ref{supp:results} & Detailed Results & Section~5.2 & \pageref{supp:results} \\
\ref{supp:scale} & Automatic Boxes and Expert Labels & Section~5.2 & \pageref{supp:scale} \\
\ref{supp:recognition} & Recognition & Section~5.2 & \pageref{supp:recognition} \\
\ref{supp:refusal} & Refusal & Section~5.2 & \pageref{supp:refusal} \\
\ref{supp:rl} & RL Results & Section~5.2 & \pageref{supp:rl} \\
\ref{supp:heldout} & Held-Out Sources & Section~5.3 & \pageref{supp:heldout} \\
\ref{supp:external} & External Benchmarks & Section~5.3 & \pageref{supp:external} \\
\bottomrule
\end{tabular}}

\medskip
\makeatletter
\providecommand{\suppref}[1]{\@ifundefined{r@#1}{a later section}{Section~\ref{#1}}}
\providecommand{\supptab}[1]{\@ifundefined{r@#1}{a later table}{Table~\ref{#1}}}
\makeatother

\section{Corpus Details}
\label{supp:corpus}

This section gives the details of the automatic annotation that Section~3.2 of the paper describes.

\begin{table*}[t]
\centering\small
\setlength{\tabcolsep}{4pt}
\begin{tabular}{lrrl}
\toprule
Dataset & Samples ingested & Groundable & Primary domain \\
\midrule
CDDM \citep{liu2024cddm} & 1,060,294 & 513,650 & Crop disease QA \\
LeafNet \citep{nguyen2026leafnet} & 121,337 & 81,747 & Leaf disease \\
MIRAGE \citep{dongre2025mirage} & 40,889 & 10,560 & Expert consultations \\
AgroMind \citep{li2025agromind} & 28,482 & 1,664 & Agricultural remote sensing \\
LeafBench \citep{nguyen2026leafnet} & 13,950 & 4,140 & Leaf disease VQA \\
AgroCoT \citep{wen2025agrocot} & 4,535 & 318 & Chain-of-thought reasoning \\
AgroBench \citep{shinoda2025agrobench} & 4,342 & 2,810 & Disease, pest, and weed identification \\
AgMMU \citep{gauba2025agmmu} & 772 & 596 & Grower--expert QA \\
\midrule
Total & 1,274,601 & 615,485 & \\
\bottomrule
\end{tabular}
\caption{\textbf{Source datasets and the outcome of label extraction.} A sample is one question--answer pair. The groundable samples and the 659,116 excluded samples sum to the ingested total. The counts are those of the releases that we ingested and can differ from the numbers in the source papers. For LeafNet we ingested the public training split, and for AgroCoT the first release, which has 4,535 samples where later versions report 4,759.}
\label{tab:s_sources}
\end{table*}

\textbf{Sources.} \agro{} draws on eight agricultural VQA datasets (Table~\ref{tab:s_sources}). A \emph{sample} is one question--answer pair. The sources ask several questions about the same photograph and share photographs with one another, so we also report numbers of unique image files and, for the training split, of perceptually unique images. Where a source carries box annotations of its own, as AgroBench does for its weed-identification images, we do not use them as supervision. Every training box comes from the pipeline below.

\textbf{Label extraction.} Many VQA answers name nothing that can be localized in the image. Source-specific handlers assign each sample that they can parse a kind (disease, pest, weed, symptom description, stress condition, object, action, cause, or healthy) and extract a clean target label and the host crop. They also repair source-specific irregularities such as answer templates and abbreviations. A sample whose answer names a target that can be localized is \emph{groundable}. The other samples are excluded. These are answers that name a management action or a cause, the plant-identification questions of CDDM (4,115 samples), and the healthy samples (154,180). Also excluded are aerial field-level labels, labels that are not agricultural, too generic, or only the name of a crop, and samples from which no target can be extracted. Species answers of the other sources that name a visible organism other than a bare crop name are kept as samples of the kind ``object''. The excluded samples are kept with the reason for their exclusion (659,116 samples, 229,938 unique image files). The healthy samples later supply the healthy negatives of training and the healthy images of the benchmark. Table~\ref{tab:s_kinds} lists the 615,485 groundable samples by kind.

\begin{table}[t]
\centering\small
\setlength{\tabcolsep}{8pt}
\begin{tabular}{lrr}
\toprule
Kind & Samples & Share \\
\midrule
Disease & 600,905 & 97.6\% \\
Object & 7,278 & 1.2\% \\
Pest & 4,103 & 0.7\% \\
Symptom description & 2,543 & 0.4\% \\
Weed & 609 & 0.1\% \\
Stress condition & 47 & $<$0.1\% \\
\midrule
Total & 615,485 & 100\% \\
\bottomrule
\end{tabular}
\caption{\textbf{Groundable samples by kind of target.} A symptom description is an answer that describes a visible symptom without naming a disease. An object is an identifiable agricultural object or organism.}
\label{tab:s_kinds}
\end{table}

\textbf{Taxonomy and detector prompts.} The detector is not queried with the bare label alone, which names the target without the host crop or the visual symptom. Labels are normalized to a taxonomy of 6,038 labels. The detector prompt of a sample is built from its crop and from the taxonomy entry of its normalized label. For the sample in Figure~2 of the paper, labeled anthracnose on cashew, the prompt is ``sunken dark lesions on cashew leaf''. Three rules complete the prompts. Synonyms and variant spellings of a label are mapped to one taxonomy entry. The default host of a taxonomy entry is replaced by the crop that the sample names (over 82,000 CDDM and LeafNet samples). Diseases whose symptoms are not on leaves, such as root rot, crown gall, and stem canker, have prompts of their own. All labels are lowercased wherever they appear, in training instructions and answers, in candidate lists, and in evaluation prompts, so that no casing pattern distinguishes one label from another.

\textbf{Box generation.} Each readable groundable sample is annotated by GroundingDINO~\citep{liu2024groundingdino} (Swin-T backbone) with its prompt. The detector runs with a box threshold of 0.25 and a text threshold of 0.20, a recall-oriented setting that was fixed before any evaluation data were collected. Because the audit below found detector confidence only weakly related to box quality, the boxes are then filtered by geometry~\citep{mumuni2024samannotation} and not by a further threshold on the score. Degenerate boxes, exact duplicates, and boxes outside 0.1\% to 95\% of the image area are removed. For targets of the kinds object and stress condition, only the box with the highest detector confidence is kept. For the other kinds, all remaining boxes are kept. Detector confidence thus chooses a box where only one is kept, but no box is removed for falling below a score.

\textbf{Granularity.} The level of an example follows from the geometry of its boxes. The remaining boxes of a sample are divided by area. Boxes that cover at most 50\% of the image form the lesion-level group, and larger boxes form the object-level group. Class-agnostic non-maximum suppression at IoU 0.5 is applied within each group, so a large box cannot suppress a small one. The lesion-level group gives a lesion-level example with all of its boxes that remain after suppression. The object-level group gives an object-level example with one box, the box of the group with the highest detector confidence. A sample with boxes in both groups contributes one example at each level, each with the instruction of its level. The level does not depend on the kind of target: a sample of the kind object whose box covers at most half of the image gives a lesion-level example. Diffuse diseases produce boxes around the whole leaf. Such a box becomes an object-level example when it covers more than half of the image.

\textbf{Accounting.} Of the 1,274,601 ingested samples, 615,485 are groundable (209,740 unique image files). Of these, 614,765 are readable, and 579,260 of the readable samples (94.2\%) receive at least one valid box (188,832 unique image files). Granularity assignment gives 794,850 examples, 391,758 at lesion level and 403,092 at object level. They are split 80/10/10 by source sample into 635,977 training, 79,464 validation, and 79,409 test examples, which contain 171,580, 49,395, and 49,275 unique image files. The split is stratified by source and keeps the examples of one sample in the same split. Because the split is made by sample and the sources reuse photographs, the three splits share image files. Benchmark images must therefore share no file with the training split and be no perceptual near-duplicate of any training image, positive or negative (Section~3.3 of the paper). The 171,580 training image files contain 104,150 perceptually unique images (64-bit perceptual hash).

\textbf{Audit of the automatic boxes.} We audited the automatic boxes of 100 images. The images were sampled from the annotated corpus over the eight largest strata of source and level (seed 7). Six strata contribute 13 images each. These are CDDM and LeafNet at lesion level and at object level, MIRAGE at lesion level, and LeafBench at object level. AgroBench at lesion level and LeafBench at lesion level contribute 11 images each. Each image was judged with its boxes drawn on it, under the conventions of the benchmark annotation. All boxes were correct in 62 images. In 9 images, 7 of them object-level examples, a box covered the whole leaf where the conventions call for boxes around the lesions. In 10 images some instances of the target had no box. In 10 images a box around the whole leaf appeared next to the boxes of its lesions. In 9 images the boxes were on something other than the labeled target. No image fell into more than one of these categories. Detector confidence is only weakly related to box quality. The mean box confidence is 0.425 on the 62 images with all boxes correct and 0.372 on the 38 others. Its rank correlation with the verdict is weak (Spearman $\rho = 0.17$, $p = 0.08$). For the maximum box confidence of an image, $\rho = 0.05$ ($p = 0.65$). This supports applying no confidence threshold to the boxes beyond the thresholds of the detector itself.

\section{Benchmark Details}
\label{supp:benchmark}

This section gives the details of the benchmark of Section~3.3 of the paper. It covers the near-duplicate check, the construction history, the audit that removed two images, the composition, the annotation conventions, the detector proposals, and the annotators.

\textbf{Near-duplicate check.} Every image file is hashed with a 64-bit perceptual hash (the 8$\times$8 lowest frequencies of a discrete cosine transform). Two images are flagged as near-duplicates if the Hamming distance between their hashes is at most 6. In the audit described below, a flagged pair is then verified at pixel level by the correlation between 96$\times$96 grayscale thumbnails of the two images. A correlation above 0.97 confirms a copy. A pair between 0.90 and 0.97 is treated as ambiguous, and its benchmark image is excluded. A pair at 0.90 or below is a hash collision, and its benchmark image is kept. Perceptual hashing catches copies, resized images, and re-encoded images.

\textbf{Construction history.} The benchmark was annotated in four rounds. The positive candidates came from the test split of the corpus, which is split by source sample, and the healthy candidates from the healthy pool of \suppref{supp:corpus}. The images of the first three rounds were checked against the training split, and their healthy images also against the healthy images used as training negatives. The candidates of the fourth round were checked before annotation, by file and by perceptual hash, against the training split, the healthy training negatives, and the images already in the benchmark. Images found to be near-duplicates and images repeated within the benchmark were removed. This gave 1,480 images, with no near-duplicate pair within the set.

\textbf{The audit that removed two images.} For the images of the first three rounds, the checks above had compared every image with the training split, but only the healthy images with the healthy training negatives. A later audit compared every benchmark image with every training image, positive or negative. It found two positive images from the first three rounds that are copies of healthy training negatives. Both are object-level images with one box. Removing them gives the 1,478 images of the paper. These 1,478 images pass the audit. None shares a file with a training image, and none has a verified copy or an ambiguous pair among the training images. No two of them are near-duplicates of each other. Unless stated otherwise, the scores of the paper and of this appendix are computed on these 1,478 images. Three experiments were run before the audit: the detector fine-tuned on the automatic boxes, the second model family, and the study of image resolution. Their scores were recomputed on the 1,478 images, which changed no F1 value by more than 0.001 and no percentage by more than 0.2 points. Their intervals and paired differences, and the joint accuracy of the fine-tuned detector, were computed on the 1,480 images.

\begin{table}[t]
\centering\small
\setlength{\tabcolsep}{5pt}
\begin{tabular}{lrrr}
\toprule
Source & Images & Positive & Healthy \\
\midrule
LeafNet & 512 & 342 & 170 \\
CDDM & 299 & 79 & 220 \\
MIRAGE & 289 & 289 & 0 \\
AgroBench & 174 & 174 & 0 \\
LeafBench & 122 & 20 & 102 \\
AgroMind & 42 & 32 & 10 \\
AgMMU & 35 & 35 & 0 \\
AgroCoT & 5 & 2 & 3 \\
\midrule
Total & 1,478 & 973 & 505 \\
\bottomrule
\end{tabular}
\caption{\textbf{Benchmark images by source.} A positive image shows its target. A healthy image shows no target.}
\label{tab:s_bench_sources}
\end{table}

\textbf{Composition.} Table~\ref{tab:s_bench_sources} lists the 1,478 benchmark images by source. Of the 973 positive images, 508 are lesion-level and 465 are object-level. By kind of target, they show diseases (605), pests (188), objects or species (163), symptom descriptions (11), weeds (3), and stress conditions (3). The 3,461 boxes are 2,790 lesion-level boxes and 671 object-level boxes. The 180 dense images have 12.2 boxes on average.

\textbf{Annotation conventions.} The target name and a reference sheet with the symptom description of the taxonomy were visible during annotation. Boxes are tight to the visible symptom, and all instances are annotated. A dense cluster of merged lesions may be covered by one box over the contiguous affected region. An image whose target is not verifiably visible is removed, and ambiguous cases are flagged and excluded. Healthy images were confirmed visually.

\textbf{Detector proposals.} Every round showed the positive images with detector boxes as editable proposals. A positive candidate is a lesion-level or object-level example of the test split, and its proposals are the automatic boxes of that example (\suppref{supp:corpus}). The record of each round keeps the proposals, so their effect on the final boxes can be measured (Table~\ref{tab:s_proposals}). These statistics were computed before the audit above, on 975 positive images with 3,463 boxes. The annotation tool shifts coordinates by a pixel or two on export, so a proposal counts as kept if it overlaps a final box at IoU $\ge 0.9$. The annotator kept 66.7\% of the proposals, adjusted 2.6\% (IoU 0.5 to 0.9), and deleted 30.7\%. Of the final boxes, 63.2\% have no proposal within IoU 0.5 and were drawn by the annotator, and 36.8\% match a kept or adjusted proposal.

\begin{table}[t]
\centering\small
\setlength{\tabcolsep}{5pt}
\begin{tabular}{lrrrrr}
\toprule
Round & 1 & 2 & 3 & 4 & All \\
\midrule
Images & 213 & 106 & 218 & 438 & 975 \\
Proposals & 356 & 411 & 448 & 623 & 1,838 \\
Kept & 51.4\% & 68.6\% & 72.8\% & 69.8\% & 66.7\% \\
Adjusted & 6.5\% & 0.5\% & 2.5\% & 1.9\% & 2.6\% \\
Deleted & 42.1\% & 30.9\% & 24.8\% & 28.3\% & 30.7\% \\
Final boxes & 791 & 674 & 791 & 1,207 & 3,463 \\
Drawn & 74.0\% & 57.9\% & 57.4\% & 63.0\% & 63.2\% \\
IoU & 0.963 & 0.997 & 0.991 & 0.991 & 0.988 \\
\bottomrule
\end{tabular}
\caption{\textbf{Detector proposals by annotation round.} Positive images only, as they stood before the audit removed two images. Kept: proposals that overlap a final box at IoU $\ge 0.9$. Adjusted: IoU 0.5 to 0.9. Drawn: final boxes with no proposal within IoU 0.5. IoU: mean IoU over the matched pairs of a proposal and a final box. Percentages may not sum to 100 because of rounding.}
\label{tab:s_proposals}
\end{table}

\textbf{Annotators.} The annotator of the benchmark and the second annotator of the agreement study were both trained on these conventions. The expert check of \suppref{supp:expert} was done by a plant pathologist who took no part in the annotation.

\section{Agreement Study}
\label{supp:agreement}

This section gives the details of the agreement study of Section~3.3 of the paper.

\textbf{Design.} The study covers a stratified random subset of 500 benchmark images (seed 55). It has 100 dense lesion-level images, 150 other lesion-level images, 150 object-level images, and 100 healthy images. Of the object-level images, 65 have a target other than a disease. Every source is represented. Within each stratum, half of the images were assigned at random to annotation with proposals and half to annotation without proposals, which gives 250 images each. With proposals, the second annotator saw the detector proposals of the original annotation as editable boxes (407 boxes on the 200 positive images of this half). Without proposals, no box was shown. The second annotator worked under the conventions of Section~\ref{supp:benchmark} and never saw the boxes of the first annotator. The reference sheet of each image gave the target name, the crop, the symptom description of the taxonomy, and whether the image is a healthy image. The healthy images were thus presented for confirmation, as in the original annotation. The first annotator had annotated every positive image with proposals, so for the first annotator the two halves were annotated in the same way.

\textbf{Scoring.} The second annotator drew 1,590 boxes, 771 with proposals and 819 without. Two healthy images received one box each, and one positive image received none. These three images count as disagreements. Agreement on boxes is computed on the 400 positive images with the matching of the paper. It is greedy one-to-one matching in descending IoU order at IoU 0.5, pooled over images, with the first annotator as reference. Taking the second annotator as reference exchanges precision and recall and leaves F1 unchanged. Intervals are 95\% bootstrap intervals over images (1,000 resamples).

\begin{table*}[t]
\centering\small
\setlength{\tabcolsep}{8pt}
\begin{tabular}{lrlccc}
\toprule
Images & $n$ & F1 [95\% CI] & Precision & Recall & Box ratio \\
\midrule
All positive images & 400 & 0.789 [0.761, 0.816] & 0.811 & 0.768 & 0.95 \\
\quad With proposals & 200 & 0.797 [0.752, 0.836] & 0.829 & 0.767 & 0.93 \\
\quad Without proposals & 200 & 0.781 [0.737, 0.820] & 0.794 & 0.769 & 0.97 \\
\midrule
Dense lesion-level & 100 & 0.794 [0.756, 0.826] & 0.830 & 0.761 & 0.92 \\
Other lesion-level & 150 & 0.762 [0.698, 0.824] & 0.773 & 0.751 & 0.97 \\
Object-level & 150 & 0.796 [0.714, 0.879] & 0.767 & 0.828 & 1.08 \\
\bottomrule
\end{tabular}
\caption{\textbf{Agreement between the two annotators on the 400 positive images of the subset.} F1, precision, and recall use greedy one-to-one matching at IoU 0.5 with the first annotator as reference, pooled over images. Box ratio is the number of boxes of the second annotator divided by that of the first. With and without proposals describe what the second annotator saw. Strata follow the levels of the benchmark.}
\label{tab:s_agree}
\end{table*}

\begin{figure*}[t]
\centering
\includegraphics[width=\linewidth]{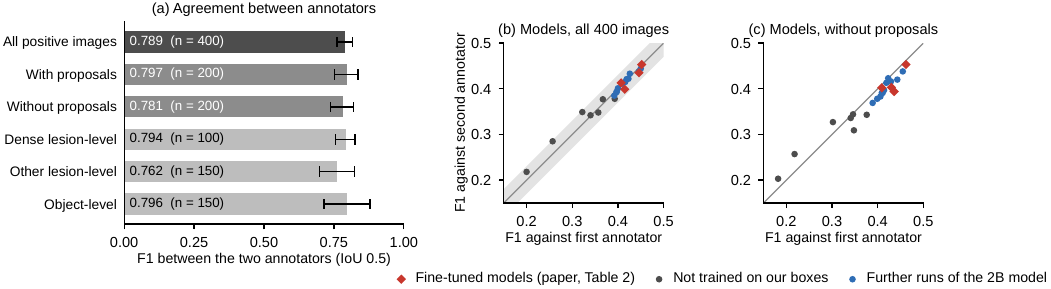}
\caption{\textbf{Agreement study.} (a) Instance F1 between the two annotators at IoU 0.5 with 95\% bootstrap intervals, on all 400 positive images, on the halves with and without proposals for the second annotator, and by stratum. (b, c) F1 of the 22 models of Table~\ref{tab:s_agree_models} against the first and against the second annotator. Panel (b) uses all 400 positive images, and panel (c) the 200 positive images that the second annotator annotated without proposals. Fine-tuned models (paper, Table~2): the four models of Table~2 of the paper that were fine-tuned on our boxes. Not trained on our boxes: the seven other models of the first block of Table~\ref{tab:s_agree_models}. Further runs of the 2B model: the second block of that table. The gray band in panel (b) marks a difference of at most 0.03.}
\label{fig:s_agreement}
\end{figure*}

\textbf{Agreement on boxes.} Table~\ref{tab:s_agree} gives the results, and Figure~\ref{fig:s_agreement}a shows them with their intervals. On the 400 positive images, F1 is 0.789 [0.761, 0.816], with 1,288 matched pairs among the 1,677 boxes of the first annotator and the 1,588 boxes of the second. F1 is close with and without proposals (0.797 and 0.781). By stratum, F1 is 0.794 on the dense lesion-level images, 0.762 on the other lesion-level images, and 0.796 on the object-level images.

\textbf{Agreement on decisions.} On whether a target is present, the annotators agree on 497 of the 500 images (Cohen's $\kappa$ 0.981, raw agreement 99.4\%). The three other images are the two healthy images that the second annotator boxed and the positive image without a box. For lesion level versus object level, one geometric rule classifies both sets of boxes, so that the comparison is symmetric. An image is at object level when its largest box covers more than half of the image, and at lesion level otherwise. On the 399 positive images that both annotators boxed, this rule gives $\kappa$ 0.747 (raw agreement 90.0\%).

\begin{table*}[t]
\centering\small
\setlength{\tabcolsep}{6pt}
\begin{tabular}{lccccccc}
\toprule
 & \multicolumn{3}{c}{All (400 images)} & \multicolumn{2}{c}{Lesion-level (250)} & \multicolumn{2}{c}{No proposals (200)} \\
\cmidrule(lr){2-4}\cmidrule(lr){5-6}\cmidrule(lr){7-8}
Model & First & Second & Difference & First & Second & First & Second \\
\midrule
\multicolumn{8}{l}{\emph{Models of Table 2 of the paper}} \\
GroundingDINO, raw prompts & 0.367 & 0.377 & +0.010 & 0.310 & 0.326 & 0.341 & 0.336 \\
GroundingDINO, taxonomy prompts & 0.393 & 0.378 & $-$0.015 & 0.362 & 0.357 & 0.376 & 0.343 \\
GroundingDINO, pipeline configuration (teacher) & 0.357 & 0.348 & $-$0.009 & 0.282 & 0.277 & 0.348 & 0.309 \\
GroundingDINO, fine-tuned on our boxes (20\%) & 0.446 & 0.435 & $-$0.011 & 0.413 & 0.409 & 0.430 & 0.403 \\
VisionReasoner-7B & 0.340 & 0.342 & +0.002 & 0.294 & 0.301 & 0.346 & 0.344 \\
Seg-Zero-7B & 0.200 & 0.218 & +0.018 & 0.118 & 0.139 & 0.182 & 0.203 \\
Qwen3-VL-2B, zero-shot & 0.257 & 0.285 & +0.028 & 0.174 & 0.210 & 0.218 & 0.257 \\
Qwen3-VL-8B, zero-shot & 0.322 & 0.349 & +0.027 & 0.261 & 0.296 & 0.302 & 0.327 \\
Ours 2B, mixed + healthy, seed-42 run (main model) & 0.415 & 0.399 & $-$0.016 & 0.381 & 0.365 & 0.436 & 0.394 \\
Ours 2B, + RL, first RL run & 0.452 & 0.453 & +0.001 & 0.425 & 0.429 & 0.462 & 0.453 \\
Ours 8B, mixed + healthy & 0.407 & 0.413 & +0.006 & 0.370 & 0.375 & 0.409 & 0.402 \\
\midrule
\multicolumn{8}{l}{\emph{Further runs and configurations of the 2B model}} \\
Mixed + healthy (12\%), seed 1 & 0.397 & 0.393 & $-$0.004 & 0.362 & 0.358 & 0.409 & 0.391 \\
Mixed + healthy (12\%), seed 2 & 0.409 & 0.411 & +0.002 & 0.379 & 0.382 & 0.411 & 0.391 \\
Grounding only & 0.392 & 0.385 & $-$0.007 & 0.349 & 0.343 & 0.389 & 0.369 \\
Mixed & 0.397 & 0.393 & $-$0.004 & 0.359 & 0.356 & 0.399 & 0.378 \\
Mixed + healthy (6\%) & 0.400 & 0.401 & +0.001 & 0.367 & 0.369 & 0.406 & 0.383 \\
Mixed + healthy + continued SFT (control), first run & 0.426 & 0.433 & +0.007 & 0.393 & 0.401 & 0.423 & 0.423 \\
Mixed + healthy + continued SFT (control), seed 1 & 0.423 & 0.422 & $-$0.001 & 0.389 & 0.387 & 0.429 & 0.416 \\
Mixed + healthy + continued SFT (control), seed 2 & 0.419 & 0.421 & +0.002 & 0.385 & 0.385 & 0.419 & 0.413 \\
Final RL checkpoint (step 292) & 0.446 & 0.438 & $-$0.008 & 0.428 & 0.422 & 0.443 & 0.420 \\
RL, recognition-weighted reward & 0.450 & 0.444 & $-$0.006 & 0.422 & 0.418 & 0.455 & 0.438 \\
RL, uniform rollout sampling & 0.416 & 0.415 & $-$0.001 & 0.395 & 0.397 & 0.414 & 0.397 \\
\bottomrule
\end{tabular}
\caption{\textbf{Models against each annotator.} Known-target protocol, instance F1 at IoU 0.5 on the positive images of the agreement subset, with the boxes of the first or of the second annotator as reference (First, Second). All: the 400 positive images. Lesion-level: the 250 lesion-level images. No proposals: the 200 positive images that the second annotator annotated without proposals. Difference: second minus first. Seed 1 and seed 2 are the two further training runs of a configuration. The models of Table~2 of the paper that are queried through an API are not included.}
\label{tab:s_agree_models}
\end{table*}

\textbf{Models against each annotator.} Table~\ref{tab:s_agree_models} scores the models on the 400 positive images of the subset against each annotator. The values differ from those of Table~2 of the paper, which uses all 973 positive images. No model differs by more than 0.028 F1 between the two references on these 400 images (Figure~\ref{fig:s_agreement}b). On the 200 positive images that the second annotator annotated without proposals, GroundingDINO with raw prompts reaches 0.336 against the second annotator (Figure~\ref{fig:s_agreement}c). The four models of Table~2 of the paper that were fine-tuned on our boxes reach 0.394 (main model, seed-42 run), 0.402 (Ours 8B), 0.403 (fine-tuned detector), and 0.453 (first RL run). These are the differences of 0.058 to 0.117 that the paper reports. For the main model, the paired difference on this half against the second annotator is +0.058 [+0.007, +0.114]. All benchmark scores of the paper use the boxes of the first annotator.

\begin{table*}[t]
\centering\small
\setlength{\tabcolsep}{10pt}
\begin{tabular}{lccc}
\toprule
Comparison & First annotator & Second annotator & Same sign \\
\midrule
Main model minus GroundingDINO, raw prompts & +0.048 & +0.022 & yes \\
Main model minus GroundingDINO, pipeline configuration (teacher) & +0.058 & +0.051 & yes \\
Main model minus GroundingDINO, fine-tuned on our boxes (20\%) & $-$0.031 & $-$0.037 & yes \\
Main model minus Qwen3-VL-2B, zero-shot & +0.158 & +0.113 & yes \\
Ours 8B minus Qwen3-VL-8B, zero-shot & +0.085 & +0.063 & yes \\
Ours 8B minus main model & $-$0.008 & +0.014 & no \\
Mixed minus grounding only & +0.005 & +0.007 & yes \\
Main model minus mixed & +0.018 & +0.006 & yes \\
RL minus main model & +0.037 & +0.054 & yes \\
RL minus continued SFT (control) & +0.026 & +0.020 & yes \\
Continued SFT (control) minus main model & +0.011 & +0.035 & yes \\
\bottomrule
\end{tabular}
\caption{\textbf{Paired comparisons under each annotator.} Difference in F1 between two models on the same 400 positive images of the agreement subset (known-target protocol, IoU 0.5). The reference is the first or the second annotator. Main model: Ours 2B, mixed + healthy, seed-42 run. RL and continued SFT (control): first runs.}
\label{tab:s_agree_paired}
\end{table*}

\textbf{Paired comparisons.} Table~\ref{tab:s_agree_paired} gives the eleven paired comparisons between models that the paper counts, under each reference. Ten of the eleven differences keep their sign. The exception is the 8B model against the main model, whose difference is below 0.015 under both references.

\clearpage\clearpage
\section{Expert Check}
\label{supp:expert}

This section gives the details of the expert check of Section~3.3 of the paper.

\textbf{Design.} The design and the main numbers to report were fixed before any answer was collected. These numbers are the judgments on each kind of item, with the wrong names of the same crop separately, and their Wilson intervals. The sample has 300 benchmark images, each used once, in three kinds of item. In 159 items a positive image is shown with the benchmark's own target name. In 99 items another positive image is shown with the wrong name of its wrong-name query. In 42 items a healthy image is shown with the name that the benchmark queries on it. Within the positive images and within the healthy images, the sample is proportional to the sources and random within each source (fixed seed). The expert is a plant pathologist (M.Sc.\ in plant pathology, more than eight years as a plant disease and pest diagnostician) who took no part in the annotation. The expert saw one image at a time, without boxes, with one name and with the crop where the benchmark records one. The expert was told that names that should be visible and names that should not were mixed, but not which was which, and answered ``visible'', ``not visible'', or ``cannot tell''. All 300 items are counted.

\begin{table}[t]
\centering\footnotesize
\setlength{\tabcolsep}{3.5pt}
\begin{tabular}{@{}lrrcrr@{}}
\toprule
 & & & Agrees (\%) & & Cannot \\
Item & $n$ & Agrees & [95\% CI] & Against & tell \\
\midrule
Own name & 159 & 130 & 81.8 [75.0, 87.0] & 25 & 4 \\
\quad lesion-level & 88 & 75 & 85.2 [76.3, 91.2] & 12 & 1 \\
\quad object-level & 71 & 55 & 77.5 [66.5, 85.6] & 13 & 3 \\
Wrong name & 99 & 81 & 81.8 [73.1, 88.2] & 18 & 0 \\
\quad same crop & 51 & 40 & 78.4 [65.4, 87.5] & 11 & 0 \\
\quad other & 48 & 41 & 85.4 [72.8, 92.8] & 7 & 0 \\
Healthy image & 42 & 38 & 90.5 [77.9, 96.2] & 4 & 0 \\
\midrule
All items & 300 & 249 & 83.0 [78.3, 86.8] & 47 & 4 \\
\bottomrule
\end{tabular}
\caption{\textbf{Expert check.} Own name: a positive image with the benchmark's own target name. Wrong name: a positive image with the wrong name of its wrong-name query. Healthy image: a healthy image with the name that the benchmark queries on it. Agrees: the expert judged an own name visible, or a wrong name or the name on a healthy image not visible. Against: the opposite judgment. Lesion-level and object-level: level of the image in the benchmark. Same crop: the wrong name is another benchmark target of the crop of the image. Other: no such target exists, and the wrong name is drawn from all benchmark labels. Percentages are of all $n$ items, with Wilson 95\% intervals.}
\label{tab:s_expert}
\end{table}

\begin{figure}[t]
\centering
\includegraphics[width=\linewidth]{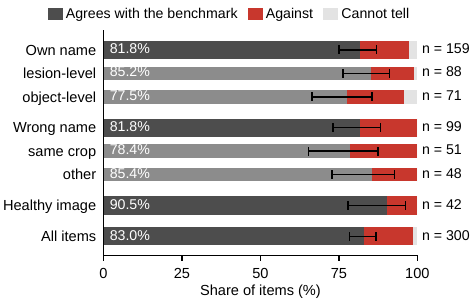}
\caption{\textbf{Judgments of the expert by kind of item.} Share of items on which the expert agrees with the benchmark, decides against it, or cannot tell. Black lines: Wilson 95\% intervals of the share that agrees. Lighter bars: subsets of the darker bar above them. Numbers as in Table~\ref{tab:s_expert}.}
\label{fig:s_expert}
\end{figure}

\textbf{Results.} Table~\ref{tab:s_expert} and Figure~\ref{fig:s_expert} give the judgments. The expert judged the benchmark's own name visible on 130 of the 159 positive images (81.8\% [75.0, 87.0]), not visible on 25, and could not tell on 4. The wrong name was judged not visible on 81 of the 99 images (81.8\% [73.1, 88.2]) and visible on 18. On the healthy images, the queried name was judged not visible on 38 of 42 (90.5\% [77.9, 96.2]). Over all 300 items, the expert agreed with the benchmark on 249 (83.0\% [78.3, 86.8]). For wrong names, the agreement is 78.4\% on the 51 names of the same crop and 85.4\% on the 48 other names.

\begin{table*}[t]
\centering\small
\setlength{\tabcolsep}{4pt}
\begin{tabular}{lcccccc}
\toprule
 & \multicolumn{2}{c}{F1} & \multicolumn{2}{c}{Recognition (\%)} & \multicolumn{2}{c}{Wrong-name refusal (\%)} \\
\cmidrule(lr){2-3}\cmidrule(lr){4-5}\cmidrule(lr){6-7}
Configuration & All (973) & Confirmed (130) & All (973) & Confirmed (130) & All (973) & Confirmed (81) \\
\midrule
Zero-shot Qwen3-VL-2B & 0.319 & 0.314 & 51.7 & 50.0 & 0.0 & 0.0 \\
Grounding only & 0.425 & 0.406 & 29.2 & 29.2 & 0.0 & 0.0 \\
Mixed & 0.435 & 0.398 & 72.7 & 75.4 & 0.0 & 0.0 \\
Mixed + healthy (6\%) & 0.433 & 0.414 & 73.1 & 74.6 & 25.7 & 22.2 \\
Mixed + healthy (12\%), main (seed-42 run) & 0.440 & 0.412 & 74.7 & 76.9 & 30.2 & 35.8 \\
\quad + wrong-target 2.9\% & 0.412 & 0.388 & 72.0 & 74.6 & 52.1 & 56.8 \\
\quad + wrong-target 5.6\% & 0.409 & 0.362 & 68.0 & 71.5 & 59.7 & 64.2 \\
\quad + wrong-target 10.6\% & 0.392 & 0.368 & 67.5 & 70.0 & 62.0 & 65.4 \\
Wrong-target 11.9\% only & 0.395 & 0.381 & 68.6 & 70.8 & 61.9 & 67.9 \\
Mixed + healthy + RL & 0.477 & 0.475 & 69.3 & 70.0 & 8.1 & 3.7 \\
Mixed + healthy + continued SFT (control) & 0.458 & 0.432 & 73.7 & 77.7 & 36.5 & 40.7 \\
Final RL checkpoint (step 292) & 0.467 & 0.455 & 67.2 & 69.2 & 22.1 & 19.8 \\
\bottomrule
\end{tabular}
\caption{\textbf{Scores on the images that the expert confirmed.} F1: known-target protocol, instance F1 at IoU 0.5 pooled over the positive images. Recognition: candidate protocol. Wrong-name refusal: share of positive images answered with an empty list when queried with a wrong name. All: the 973 positive images of the benchmark. Confirmed: for F1 and recognition, the 130 positive images whose own name the expert judged visible. For wrong-name refusal, the 81 positive images whose wrong name the expert judged not visible. RL and continued SFT (control): first runs.}
\label{tab:s_expert_rescored}
\end{table*}

\begin{table}[t]
\centering\footnotesize
\setlength{\tabcolsep}{3pt}
\begin{tabular}{@{}lcc@{}}
\toprule
 & Full benchmark & Confirmed \\
Difference & [95\% CI] & images \\
\midrule
\multicolumn{3}{l}{\emph{F1}} \\
Grounding only minus zero-shot & +0.106 [+0.084, +0.125] & +0.092 \\
Wrong-target 2.9\% minus main & $-$0.028 [$-$0.043, $-$0.013] & $-$0.024 \\
Wrong-target 5.6\% minus main & $-$0.032 [$-$0.047, $-$0.017] & $-$0.050 \\
Wrong-target 10.6\% minus main & $-$0.048 [$-$0.065, $-$0.032] & $-$0.044 \\
Wrong-target only minus main & $-$0.046 [$-$0.063, $-$0.030] & $-$0.031 \\
RL minus main & +0.037 [+0.020, +0.053] & +0.063 \\
Final RL minus main & +0.026 [+0.010, +0.043] & +0.043 \\
Control minus main & +0.017 [+0.003, +0.031] & +0.020 \\
RL minus control & +0.019 [+0.004, +0.034] & +0.043 \\
\midrule
\multicolumn{3}{l}{\emph{Recognition (points)}} \\
Grounding only minus zero-shot & $-$22.5 [$-$26.3, $-$18.4] & $-$20.8 \\
Mixed minus grounding only & +43.5 [+39.8, +47.2] & +46.2 \\
Wrong-target 2.9\% minus main & $-$2.7 [$-$4.9, $-$0.3] & $-$2.3 \\
Wrong-target 5.6\% minus main & $-$6.7 [$-$8.9, $-$4.6] & $-$5.4 \\
Wrong-target 10.6\% minus main & $-$7.2 [$-$9.6, $-$4.8] & $-$6.9 \\
Wrong-target only minus main & $-$6.2 [$-$8.8, $-$3.7] & $-$6.1 \\
RL minus main & $-$5.4 [$-$7.6, $-$3.3] & $-$6.9 \\
Final RL minus main & $-$7.5 [$-$10.1, $-$5.1] & $-$7.7 \\
RL minus control & $-$4.4 [$-$6.6, $-$2.2] & $-$7.7 \\
Final RL minus control & $-$6.5 [$-$9.1, $-$4.0] & $-$8.5 \\
\midrule
\multicolumn{3}{l}{\emph{Wrong-name refusal (points)}} \\
Zero-shot minus main & $-$30.2 [$-$33.1, $-$27.5] & $-$35.8 \\
Grounding only minus main & $-$30.2 [$-$33.1, $-$27.5] & $-$35.8 \\
Mixed minus main & $-$30.2 [$-$33.1, $-$27.5] & $-$35.8 \\
Healthy 6\% minus main & $-$4.5 [$-$6.8, $-$2.2] & $-$13.6 \\
Wrong-target 2.9\% minus main & +21.9 [+18.4, +24.9] & +21.0 \\
Wrong-target 5.6\% minus main & +29.5 [+26.2, +33.1] & +28.4 \\
Wrong-target 10.6\% minus main & +31.8 [+28.0, +35.3] & +29.6 \\
Wrong-target only minus main & +31.7 [+27.9, +35.0] & +32.1 \\
RL minus main & $-$22.1 [$-$24.7, $-$19.4] & $-$32.1 \\
Final RL minus main & $-$8.1 [$-$10.8, $-$5.4] & $-$16.0 \\
Control minus main & +6.3 [+4.5, +8.3] & +4.9 \\
\bottomrule
\end{tabular}
\caption{\textbf{Resolved differences among the computed pairs of configurations of Tables~3 and~4 of the paper, on the full benchmark and on the images that the expert confirmed.} The 30 of the 41 computed paired differences whose 95\% interval excludes zero on the full benchmark, and the same differences on the confirmed images, computed from the values of Table~\ref{tab:s_expert_rescored} (point estimates). Main: mixed + healthy (12\%), seed-42 run. Zero-shot: Qwen3-VL-2B. Healthy 6\%: mixed + healthy (6\%). Wrong-target 2.9\% to 10.6\%: the main mixture with wrong-target negatives. Wrong-target only: wrong-target 11.9\% only. RL: first RL run. Final RL: its final checkpoint (step 292). Control: continued SFT, first run.}
\label{tab:s_expert_diffs}
\end{table}

\textbf{Scores on the confirmed images.} For the benchmark's own names, the confirmed images are the 130 positive images on which the expert judged the name visible. For wrong names, they are the 81 positive images on which the expert judged the wrong name not visible. Table~\ref{tab:s_expert_rescored} gives F1, recognition, and wrong-name refusal on the full benchmark and on these images for twelve configurations of Tables~3 and~4 of the paper, the zero-shot model and eleven trained configurations. The two rows of Table~4 that need no retraining, main + threshold and main + name check, are not included. In F1 and in recognition, paired differences were computed for 15 pairs of these configurations. These are grounding only against zero-shot, mixed against grounding only, healthy 6\% against mixed, and the main model against mixed and against healthy 6\%. They are also each of the four wrong-target configurations, RL, final RL, and the control against the main model, and RL and final RL against the control and against each other. In wrong-name refusal, differences were computed between the main model and each of the 11 other configurations. Of these 41 differences, 30 are resolved on the full benchmark, which means that their 95\% interval excludes zero: 9 in F1, 10 in recognition, and 11 in wrong-name refusal (Table~\ref{tab:s_expert_diffs}). All 30 keep their direction on the confirmed images.

\clearpage\clearpage
\renewcommand{\thealgorithm}{S\arabic{algorithm}}
\section{Evaluation Protocols and Metrics}
\label{supp:protocols}

This section gives the queries, the matching rules, and the statistics of Section~3.4 of the paper.

\begin{table*}[t]
\centering\small
\setlength{\tabcolsep}{5pt}
\begin{tabular}{llp{0.66\textwidth}}
\toprule
Protocol & Level & Instruction \\
\midrule
Known target & Lesion & Locate all regions affected by \textit{name} in this image. Output bounding boxes in JSON format. If no such regions are present, output an empty list. \\
\addlinespace
Known target & Object & Locate the leaf or plant affected by \textit{name} in this image. Output the bounding box in JSON format. If not present, output an empty list. \\
\addlinespace
Candidate & Lesion & This plant may be affected by one of the following: \textit{name 1}, \textit{name 2}, \textit{name 3}, \textit{name 4}. Identify which one is present in the image and locate all affected regions. Output bounding boxes in JSON format with the identified name as the label. If none of them are present, output an empty list. \\
\addlinespace
Candidate & Object & This plant may be affected by one of the following: \textit{name 1}, \textit{name 2}, \textit{name 3}, \textit{name 4}. Identify which one is present in the image and locate the affected leaf or plant. Output the bounding box in JSON format with the identified name as the label. If none of them are present, output an empty list. \\
\bottomrule
\end{tabular}
\caption{\textbf{Evaluation instructions of the Qwen3-VL models that we run ourselves, zero-shot and fine-tuned.} Each query is one image followed by one instruction. Names are lowercased.}
\label{tab:s_prompts}
\end{table*}

\textbf{Queries.} Table~\ref{tab:s_prompts} gives the instructions of the Qwen3-VL models that we run ourselves, zero-shot and fine-tuned. The rules of this section on queries, decoding, and the reading of answers describe these models. The queries and the settings of the other baselines and of the models queried through an API are in \suppref{supp:baselines}. The level of an instruction is the level of the image in the benchmark, and a healthy image is queried in the lesion-level wording. A positive image is queried with its true target in the known-target protocol and with its four candidates in the candidate protocol. For the refusal measures it is also queried with a wrong name, in the known-target wording of its level, so that only the name differs from the true-name query. A healthy image is queried with the disease name that the benchmark stores for it, and its correct answer is the empty list. The answer is read as a JSON list whose entries carry a box \texttt{[x1, y1, x2, y2]} on the 0--1000 grid of the model and a label. If the answer is not valid JSON, boxes are read from lists of four numbers. Answers are decoded greedily with at most 256 new tokens, and no cap is applied to the image size.

\textbf{Candidate lists.} The four candidates of a positive image are its true target and three distractors. The distractors are drawn at random from the corpus targets of the same crop and the same kind. If these are fewer than three, the draw continues with the targets of the same source and kind, then with all targets of that kind, and last with the disease targets. A name is rejected if it equals another candidate, contains one, or is contained in one. The order of the four candidates is random with a fixed seed. The four candidates of a healthy image are its stored disease name and three diseases drawn in the same way.

\textbf{Wrong and random names.} The wrong name of a positive image is drawn at random, with a fixed seed, from the targets of the other positive benchmark images of the same crop (496 images, \emph{same-crop}). Where the benchmark has no other target for the crop, or records no crop for the image, the name is drawn from all benchmark targets (477 images). Names that equal the true name, contain it, or are contained in it are excluded. The \emph{random} name of an image is drawn, with a fixed seed, from the benchmark targets that are not a target of its crop and differ from its wrong name, under the same exclusion.

\begin{algorithm}[t]
\caption{Localization metrics of one model (known-target protocol)}
\label{alg:s_metrics}
\begin{algorithmic}[1]
\STATE \textbf{Input:} predicted boxes $P_i$ and reference boxes $G_i$ of every positive image $i$
\STATE $\mathit{TP}, \mathit{FP}, \mathit{FN} \leftarrow 0$
\FOR{each positive image $i$}
  \STATE list all pairs $(p, g) \in P_i \times G_i$ with $\mathrm{IoU}(p, g) \ge 0.5$ in descending order of IoU
  \FOR{each pair $(p, g)$ of the list}
    \IF{neither $p$ nor $g$ is matched}
      \STATE match $p$ with $g$
    \ENDIF
  \ENDFOR
  \STATE $\mathit{TP} \mathrel{+}=$ matched pairs
  \STATE $\mathit{FP} \mathrel{+}=$ unmatched boxes of $P_i$
  \STATE $\mathit{FN} \mathrel{+}=$ unmatched boxes of $G_i$
  \STATE $r_i \leftarrow$ IoU of the unions of $P_i$ and of $G_i$ \hfill (0 if $P_i = \emptyset$)
\ENDFOR
\STATE precision $= \mathit{TP} / (\mathit{TP} + \mathit{FP})$
\STATE recall $= \mathit{TP} / (\mathit{TP} + \mathit{FN})$
\STATE $\mathrm{F1} = 2 \cdot \mathrm{precision} \cdot \mathrm{recall} / (\mathrm{precision} + \mathrm{recall})$
\STATE region IoU $=$ mean of $r_i$ over the positive images
\end{algorithmic}
\end{algorithm}

\textbf{Matching and localization metrics.} Algorithm~\ref{alg:s_metrics} gives the rules behind F1 and region IoU. Matching is greedy and one-to-one in descending order of IoU, and true positives, false positives, and false negatives are pooled over the 973 positive images before precision, recall, and F1 are computed. Healthy images do not enter F1. They are scored by refusal. Lesion-level F1 and the F1 on dense images apply the same rules to those images only. Region IoU compares the union of the predicted boxes with the union of the reference boxes on a raster at a quarter of the image width and height. An empty answer has region IoU 0.

\textbf{Recognition and joint accuracy.} In the candidate protocol, the name of an answer is the most frequent label of its boxes. If the answer has no labeled box, it is the candidate that the raw answer mentions, provided it mentions exactly one. Recognition is the share of the 973 positive images on which this name equals the true target. An empty list, a name outside the list, and an answer that cannot be read count as errors. The adapted protocol of three baselines is described in \suppref{supp:baselines}. Joint accuracy also requires a region IoU of at least 0.5 between the boxes of the same answer and the reference boxes.

\textbf{Refusal.} A refusal is an answer without boxes. Healthy refusal and false refusal are counted on the answers of the known-target protocol to the 505 healthy images and to the 973 positive images queried with their true target. For the Qwen3-VL models that we run ourselves, the refusal of a wrong or a random name in the known-target wording is read from the token probabilities. We read the probability of every token of the empty list under the query, and the name counts as refused when the empty list is the answer of greedy decoding. For the zero-shot models, an empty list inside a code block counts as well. Pair accuracy is the share of positive images on which the true target is localized (region IoU of at least 0.5) and the wrong name is refused. The \emph{empty-answer probability} is the probability that the model gives to the empty list followed by the end of the answer. For zero-shot models, which also write JSON inside a code block, the probabilities of the two forms are added. The AUROC is computed over the 973 positive images, with the empty-answer probabilities of the wrong-name queries as positives and those of the true-name queries as negatives. The AUROC of healthy against positive images compares the empty-answer probabilities of the healthy images with those of the positive images queried with their true target. It is computed within one wording: against the lesion-level images in the lesion-level wording, and against the object-level images in the object-level wording, in which each healthy image is queried once more. The empty-answer probability needs a forward pass with the weights of the model. The models queried through an API therefore have no AUROC, and their wrong-name refusal is counted on generated answers.

\textbf{Statistics.} Intervals are 95\% percentile bootstrap intervals over images with 1,000 resamples and a fixed seed. A paired difference between two models is computed on the same resampled images, and the reported value is the difference on the full image set. For the three-run summaries we give the mean and the sample standard deviation of the three runs. Intervals on the expert judgments are Wilson intervals.

\begin{table}[t]
\centering\footnotesize
\setlength{\tabcolsep}{4pt}
\begin{tabular}{@{}lcccc@{}}
\toprule
 & \multicolumn{4}{c}{F1} \\
\cmidrule(l){2-5}
Model & All & Dense & Other & IoU 0.75 \\
\midrule
Qwen3-VL-2B, zero-shot & 0.319 & 0.076 & 0.601 & 0.238 \\
Ours 2B, mixed + healthy & 0.440 & 0.316 & 0.603 & 0.288 \\
Ours 2B, + RL & 0.477 & 0.367 & 0.619 & 0.341 \\
\bottomrule
\end{tabular}
\caption{\textbf{F1 on subsets of the images and at a stricter IoU.} Known-target protocol. All: the F1 of the paper on the 973 positive images. Dense: the 180 lesion-level images with five or more reference boxes. Other: the other 793 positive images. IoU 0.75: all positive images, matching at IoU 0.75. Ours 2B, mixed + healthy: the main model (seed-42 run). RL: first RL run. At IoU 0.75, GroundingDINO with taxonomy prompts reaches 0.375.}
\label{tab:s_variants}
\end{table}

\textbf{Dense images and stricter matching.} Table~\ref{tab:s_variants} gives F1 on the dense images, on the other positive images, and at IoU 0.75 for the zero-shot Qwen3-VL-2B, the main model, and RL. These are the values behind three statements of Section~5.2 of the paper.

\renewcommand{\thealgorithm}{S\arabic{algorithm}}
\section{Training Details}
\label{supp:training}

This section gives the training sets, the hyperparameters, the reward, and the RL setup of Sections~4 and~5.1 of the paper.

\textbf{Positive examples.} Every lesion-level or object-level example of the training split gives one supervised example in the known-target format. Its answer is a JSON list with one entry per box. An entry holds the box on the 0--1000 grid of the model and the lowercased target label. The \emph{grounding-only} configuration trains on these 635,977 examples.

\textbf{Mixed format.} For the \emph{mixed} configuration, each example is rewritten in the candidate format with probability 0.5 (fixed seed). The candidate list holds the true target and three distractors (two where no third is found) at a random position, drawn by the rule of the benchmark (\suppref{supp:protocols}). An example for which fewer than two distractors are found keeps its known-target instruction. Images, answers, and the split do not change. This gives 318,203 examples in the candidate format and 317,774 in the known-target format (Table~\ref{tab:s_train_sets}).

\textbf{Healthy negatives.} The healthy negatives are drawn from the images of the healthy samples that have a crop (\suppref{supp:corpus}). No benchmark image is among them. Each drawn image gives two negatives, and the answer of both is an empty list. One is in the candidate format with four diseases of the crop of the image. The other is in the known-target format with a fifth disease of that crop. Where a crop has too few diseases, diseases of the same source and then of all sources complete the draw. The 6\% mixture adds 43,000 negatives from 21,500 images, and the 12\% mixture adds 86,000 negatives from 43,000 images.

\begin{table*}[t]
\centering\small
\setlength{\tabcolsep}{6pt}
\begin{tabular}{lrrrrr}
\toprule
 & \multicolumn{2}{c}{Positive examples} & \multicolumn{2}{c}{Negatives} & \\
\cmidrule(lr){2-3}\cmidrule(lr){4-5}
Configuration & Known-target & Candidate & Healthy & Wrong-target & Total \\
\midrule
Grounding only & 635,977 & -- & -- & -- & 635,977 \\
Mixed & 317,774 & 318,203 & -- & -- & 635,977 \\
Mixed + healthy (6\%) & 317,774 & 318,203 & 43,000 & -- & 678,977 \\
Mixed + healthy (12\%), main & 317,774 & 318,203 & 86,000 & -- & 721,977 \\
\quad + wrong-target 2.9\% & 317,774 & 318,203 & 86,000 & 21,500 & 743,477 \\
\quad + wrong-target 5.6\% & 317,774 & 318,203 & 86,000 & 43,000 & 764,977 \\
\quad + wrong-target 10.6\% & 317,774 & 318,203 & 86,000 & 86,000 & 807,977 \\
Wrong-target 11.9\% only & 317,774 & 318,203 & -- & 86,000 & 721,977 \\
\bottomrule
\end{tabular}
\caption{\textbf{Training sets of the supervised configurations.} Numbers of training examples by format. Half of the negatives of each kind are in the known-target format and half in the candidate format. The percentages in the names are the shares of the healthy or of the wrong-target negatives in the training set.}
\label{tab:s_train_sets}
\end{table*}

\textbf{Wrong-target negatives.} A wrong-target negative is built from a positive training image whose target is a disease and whose crop is known. The image is queried with diseases that are not its target, and the answer is an empty list. As for a healthy image, there is one negative in the candidate format with four such diseases and one in the known-target format with a fifth. Both negatives use the level of the image. In the largest set, the disease of the known-target negative comes from the crop of the image for 42,186 of the 43,000 images (98.1\%). The three doses use 10,750, 21,500, and 43,000 images. The \emph{wrong-target only} configuration replaces the 86,000 healthy negatives of the main configuration with the 86,000 wrong-target negatives.

\begin{table}[t]
\centering\footnotesize
\setlength{\tabcolsep}{4pt}
\begin{tabular}{@{}ll@{}}
\toprule
\multicolumn{2}{@{}l}{\emph{Supervised fine-tuning}} \\
Base model & Qwen3-VL-2B-Instruct or 8B-Instruct \\
Trained weights & LoRA on the language model \\
LoRA rank, $\alpha$, dropout & 64, 128, 0.05 \\
Vision encoder and projector & frozen \\
Learning rate & $10^{-4}$, cosine decay, 3\% warmup \\
Weight decay, gradient clipping & 0, 1.0 \\
Batch size & 32 \\
Epochs & 1 \\
Image size & at most 50,176 pixels \\
Longest sequence & 4,096 tokens \\
\midrule
\multicolumn{2}{@{}l}{\emph{RL}} \\
Algorithm & GRPO with the GSPO loss \\
Initial model & main model (seed-42 run) \\
Learning rate & $3 \times 10^{-6}$, 10 warmup steps \\
Prompts per step & 128 \\
Sampled answers per prompt & 8 \\
Sampling & temperature 0.7, top-$p$ 0.9, top-$k$ 50 \\
KL coefficient & 0.005 \\
Longest prompt, longest answer & 1,024 and 512 tokens \\
Steps & 292 (two passes over the prompts) \\
Checkpoints & every 10 steps \\
Image size & at most 501,760 pixels \\
\bottomrule
\end{tabular}
\caption{\textbf{Hyperparameters.} The supervised settings are the same for every supervised configuration and for both model sizes. With two GPUs, the batch of 32 is 4 examples per GPU with 4 accumulation steps for the 2B model, and 2 with 8 for the 8B model.}
\label{tab:s_hyper}
\end{table}

\textbf{Supervised fine-tuning.} Table~\ref{tab:s_hyper} lists the settings. The training script is adapted from the fine-tuning code released with the base model, and no value was searched on our data. Every supervised configuration takes one epoch over its training set. The two runs on 9,120 examples of Section~5.2 of the paper take three epochs. The main run uses the default seed of the trainer, 42, and the two further runs use seeds 1 and 2. In training, an image is reduced to at most 50,176 pixels. Evaluation applies no limit. On two NVIDIA RTX A6000 GPUs, the main run of the 2B model took 8.1 hours, and the run of the 8B model took 18.9 hours.

\begin{algorithm}[t]
\caption{Reward of one sampled answer}
\label{alg:s_reward}
\begin{algorithmic}[1]
\small
\STATE \textbf{Input:} answer $y$, automatic boxes $G$ and target of the prompt, weights $(w_f, w_l, w_m, w_r, w_a) = (0.10, 0.35, 0.15, 0.25, 0.15)$
\IF{$y$ is not a JSON list of boxes}
  \RETURN $0$
\ENDIF
\STATE $P \leftarrow$ boxes of $y$, \ $P' \leftarrow$ boxes of $P$ that cover at least 0.1\% of the image
\IF{the prompt shows a healthy image}
  \RETURN $w_f + (1 - w_f)\,[P = \emptyset]$
\ENDIF
\STATE match $P'$ with $G$ one-to-one at IoU 0.5, in descending order of IoU
\STATE $F \leftarrow$ instance F1, with the boxes of $P \setminus P'$ as unmatched boxes
\STATE $m \leftarrow$ instance recall, \ $u \leftarrow$ region IoU of $P'$ and $G$
\STATE $l \leftarrow (F + u)/2$, \ $a \leftarrow [P \neq \emptyset]$
\IF{the prompt is in the candidate format}
  \STATE $r \leftarrow 1$ if the most frequent label of $y$ is the target, $0.25$ if it is another candidate, and $0$ otherwise
\ELSE
  \STATE $r \leftarrow 0$, \ $w_r \leftarrow 0$, \ $w_l \leftarrow 1.5\,w_l$, \ $w_m \leftarrow 1.5\,w_m$
\ENDIF
\RETURN $\min\,(1,\ w_f + w_l\,l + w_m\,m + w_r\,r + w_a\,a)$
\end{algorithmic}
\end{algorithm}

\textbf{Reward.} Algorithm~\ref{alg:s_reward} gives the reward of one sampled answer. An answer that is not a valid JSON list of boxes gets 0. For a healthy image, the reward is 1 for the empty list and 0.10 for any other valid answer. For a positive image, the reward adds five terms. The format term is 0.10. The localization term is the mean of the instance F1 and the region IoU of the answer against the automatic boxes of the example, with weight 0.35. The instance term is the instance recall, with weight 0.15. The name term has weight 0.25 and is 1 if the most frequent label of the answer is the target, 0.25 if it is another candidate, and 0 otherwise. The last term has weight 0.15 and is 1 if the answer has at least one box. A known-target prompt has no name term, and the weights of the localization and instance terms are multiplied by 1.5, so that the weights again sum to 1. Boxes below 0.1\% of the image area cannot be matched and count as wrong boxes. The matching is the matching of the evaluation (\suppref{supp:protocols}), applied to one image.

\begin{table}[t]
\centering\footnotesize
\setlength{\tabcolsep}{5pt}
\begin{tabular}{@{}lrr@{}}
\toprule
Prompts & Requested & Used \\
\midrule
Known target, four or more boxes & 4,000 & 3,941 \\
Known target, fewer boxes & 4,000 & 3,950 \\
Candidate, four or more boxes & 4,000 & 3,960 \\
Candidate, fewer boxes & 4,000 & 3,952 \\
Healthy image, known target & 1,500 & 1,475 \\
Healthy image, candidate & 1,500 & 1,472 \\
\midrule
Total & 19,000 & 18,750 \\
\bottomrule
\end{tabular}
\caption{\textbf{Rollout prompts of the RL stage.} Requested: prompts drawn per group. Used: prompts that remain after the images above 0.5 megapixels are removed.}
\label{tab:s_rl_prompts}
\end{table}

\textbf{Rollout prompts.} The rollout prompts are built with a fixed seed from the supervised training examples, positive examples and healthy negatives, in the six groups of Table~\ref{tab:s_rl_prompts}. Prompts whose image has more than 0.5 megapixels are removed, which leaves 18,750 prompts. Of the 15,803 positive prompts, 7,901 have four or more boxes.

\textbf{RL setup.} RL starts from the main model with its adapter merged into the weights (Table~\ref{tab:s_hyper}). In each step, the model samples 8 answers for each of 128 prompts. Every answer is scored by the reward against the automatic boxes of its example. The model is updated with the GSPO loss and a KL term towards the initial model. The rollout prompts come from the training set, so no benchmark image is among them.

\textbf{Continued supervised fine-tuning (control).} The control starts from the same merged model and uses the supervised settings of Table~\ref{tab:s_hyper}. It trains on the 18,750 rollout prompts, with the automatic boxes of each prompt as its answer and an empty list for the healthy prompts. It runs for 1.16 passes over these prompts. This is 21,760 training examples, as many prompts as RL has seen at step 170 (170 steps of 128 prompts). Its images are limited to 501,760 pixels, as in RL. The control thus shares the prompts, the number of examples seen, and the image size with RL, and differs in the training signal. It was run three times, with the default seed and with seeds 1 and 2.

\textbf{Checkpoints.} The first RL run saves a checkpoint every 10 steps. Two of its checkpoints were kept, step 170 and the final step 292. Both were evaluated on the benchmark, and the choice of step 170 is one of the three analysis choices that Section~3.3 of the paper lists. Table~3 of the paper also reports the final checkpoint. The two further RL runs were trained to step 170.

\begin{table}[t]
\centering\footnotesize
\setlength{\tabcolsep}{5pt}
\begin{tabular}{@{}lccc@{}}
\toprule
Limit at evaluation (pixels) & F1 & Lesion F1 & Dense F1 \\
\midrule
\multicolumn{4}{@{}l}{\emph{Main model}} \\
\quad none & 0.440 & 0.392 & 0.316 \\
\quad 802,816 & 0.444 & 0.396 & 0.322 \\
\quad 200,704 & 0.437 & 0.392 & 0.313 \\
\quad 50,176 & 0.363 & 0.292 & 0.189 \\
\midrule
\multicolumn{4}{@{}l}{\emph{RL, first run}} \\
\quad none & 0.477 & 0.439 & 0.367 \\
\quad 802,816 & 0.481 & 0.444 & 0.373 \\
\quad 200,704 & 0.470 & 0.429 & 0.355 \\
\quad 50,176 & 0.391 & 0.324 & 0.222 \\
\midrule
\multicolumn{4}{@{}l}{\emph{Main mixture trained at 200,704 pixels}} \\
\quad none & 0.425 & 0.367 & 0.278 \\
\quad 200,704 & 0.425 & 0.369 & 0.284 \\
\quad 50,176 & 0.346 & 0.278 & 0.166 \\
\bottomrule
\end{tabular}
\caption{\textbf{Image size at evaluation and in training.} Known-target protocol. Lesion F1: lesion-level images. Dense F1: the 180 dense images. The limit of 50,176 pixels is the limit of supervised training, applied with the preprocessing of training.}
\label{tab:s_resolution}
\end{table}

\textbf{Image size.} Supervised training limits images to 50,176 pixels, and evaluation applies no limit. Table~\ref{tab:s_resolution} tests this difference. Evaluated at the limit of training, the main model falls from F1 0.440 to 0.363 (paired $-$0.077 [$-$0.093, $-$0.059]), and from 0.316 to 0.189 on the dense images. Limits of 200,704 or 802,816 pixels change its F1 by $-$0.003 [$-$0.013, $+$0.006] and $+$0.004 [$-$0.001, $+$0.009]. The main model thus scores higher on images that are larger than those of its training. The first RL run behaves in the same way ($-$0.086 [$-$0.102, $-$0.068] at the limit of training). We also trained the main mixture once with a limit of 200,704 pixels. This model reaches F1 0.425 without a limit at evaluation, against 0.440 for the main model (paired $-$0.015 [$-$0.031, $+$0.001]). Its recognition does not differ from that of the main model (paired $+$0.5 points [$-$1.8, $+$2.8]).

\section{Baselines and Replications}
\label{supp:baselines}

This section gives the details of the baselines and replications of Section~5.1 of the paper. It covers how each baseline is queried, the adapted candidate protocol, the two models fine-tuned on a 20\% subsample, and the models queried through an API.

\begin{table*}[t]
\centering\small
\setlength{\tabcolsep}{5pt}
\begin{tabular}{@{}lp{0.24\textwidth}p{0.35\textwidth}@{}}
\toprule
System & Query & Settings \tabularnewline
\midrule
GroundingDINO, raw prompts & \raggedright lowercased target name & \raggedright box threshold 0.25, text threshold 0.20 \tabularnewline
GroundingDINO, taxonomy prompts & \raggedright detector prompt of the pipeline & \raggedright same thresholds \tabularnewline
GroundingDINO, pipeline configuration (teacher) & \raggedright detector prompt of the pipeline & \raggedright same thresholds, then the geometry filters of the pipeline \tabularnewline
GroundingDINO, fine-tuned on our boxes (20\%) & \raggedright lowercased target name & \raggedright same thresholds \tabularnewline
\addlinespace
VisionReasoner-7B & \raggedright its published template, several boxes & \raggedright 840$\times$840 pixels, greedy, at most 1,024 new tokens \tabularnewline
Seg-Zero-7B & \raggedright its published template, one box & \raggedright 840$\times$840 pixels, greedy, at most 1,024 new tokens \tabularnewline
\addlinespace
Qwen3-VL-2B and Qwen3-VL-8B, zero-shot & \raggedright evaluation instructions & \raggedright greedy, at most 256 new tokens, no cap on the image size \tabularnewline
Qwen3-VL-235B, through an API & \raggedright evaluation instructions & \raggedright greedy, at most 256 new tokens \tabularnewline
Five closed models, through an API & \raggedright evaluation instructions and one sentence on the box format & \raggedright settings of the provider, at most 32,768 output tokens \tabularnewline
\bottomrule
\end{tabular}
\caption{\textbf{How each baseline is queried.} Evaluation instructions: the instructions of \supptab{tab:s_prompts}. The closed models are Gemini 4 Argon, GPT-6 Astra, GPT-6.1 Sol, Claude Opus 5.5, and Claude Fable 5.2.}
\label{tab:s_base_setup}
\end{table*}

\textbf{GroundingDINO.} All configurations use GroundingDINO with the Swin-T backbone~\citep{liu2024groundingdino} at the thresholds of the corpus construction, 0.25 for boxes and 0.20 for text (Table~\ref{tab:s_base_setup}). With \emph{raw prompts}, the query is the lowercased target name. With \emph{taxonomy prompts}, the query is the detector prompt that the pipeline used for the image (\suppref{supp:corpus}). Where the image has none, it is the prompt of its label, and otherwise the raw name. The \emph{pipeline configuration} applies the geometry filters of the pipeline to the boxes of the taxonomy prompts, with one answer per image. Boxes outside 0.1\% to 95\% of the image area are removed. A detector receives no instruction on the level, so the level follows from the boxes. If the largest remaining box covers more than half of the image, this box alone is kept. Otherwise, non-maximum suppression at IoU 0.5 is applied. A healthy image is queried with the raw disease name that the benchmark stores for it, also with taxonomy prompts. An image without a remaining box counts as refused.

\textbf{Reasoning-grounding systems and zero-shot bases.} VisionReasoner-7B~\citep{liu2026visionreasoner} and Seg-Zero-7B~\citep{liu2025segzero} are run with their public weights, their published prompt templates, and their input size of 840$\times$840 pixels. Decoding is greedy with at most 1,024 new tokens. The question inside the template is ``regions affected by \textit{name}'' for lesion-level and healthy images and ``the leaf or plant affected by \textit{name}'' for object-level images. VisionReasoner-7B returns several boxes and Seg-Zero-7B one box. Their boxes are scored, and an answer without a readable box counts as refused. The zero-shot Qwen3-VL-2B and Qwen3-VL-8B receive the instructions and the settings of our fine-tuned models (\suppref{supp:protocols}).

\textbf{Adapted candidate protocol.} The dagger in Table~2 of the paper marks the three rows that are scored in an adapted candidate protocol, because their systems cannot answer the candidate query directly. GroundingDINO is queried once per candidate with the lowercased name. Its answer is the candidate whose boxes reach the highest detector score, and these boxes are its localization. If no candidate returns a box, the answer is a refusal. The fine-tuned detector is queried in the same way. VisionReasoner-7B receives the four candidates inside its question (``the regions affected by the disease present in this plant, which is one of: \textit{name 1}, \textit{name 2}, \textit{name 3}, \textit{name 4}''). Its answer is the candidate that its output names first. Recognition and joint accuracy are then computed as for the other models (\suppref{supp:protocols}). The other GroundingDINO configurations and Seg-Zero-7B were not run in this protocol.

\textbf{The 20\% subsample.} Both replications train on one subsample of the corpus. It holds 20\% of the source samples of every source, drawn separately for the samples with lesion-level examples only, with object-level examples only, and with both (seed 42). This gives 92,552 source samples with 127,192 positive examples, 62,689 at lesion level and 64,503 at object level. The mixed configuration renders 63,628 of them in the candidate format. The healthy negatives are 17,202 examples from 8,601 healthy images, which is 11.9\% of the 144,394 training examples. For comparison, Qwen3-VL-2B is trained on the same subsample with the settings of \suppref{supp:training}.

\textbf{Fine-tuned detector.} GroundingDINO is fine-tuned from its public Swin-T weights on the boxes of the subsample. The boxes of the lesion-level and of the object-level example of an image are pooled, because a detector takes no instruction on the level. This gives 71,998 images with boxes. The images without boxes are healthy images of the subsample. The phrase of a box is the lowercased target label. Training uses a public training implementation of GroundingDINO with its reference optimizer settings and no hyperparameter search: AdamW with a learning rate of $10^{-4}$ ($10^{-5}$ for the backbone). The text encoder is frozen, the batch size is 16, and training takes two epochs, with the learning rate divided by 10 for the second. The final checkpoint is used. The row of Table~2 of the paper queries it with raw prompts at the thresholds above.

\begin{figure*}[t]
\centering
\includegraphics[width=\linewidth]{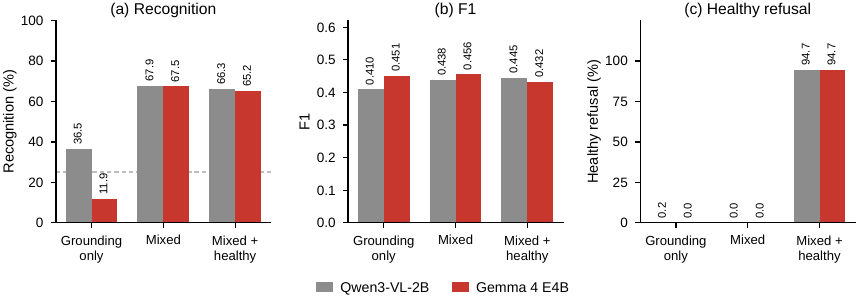}
\caption{\textbf{The three trained configurations in two model families on the 20\% subsample.} (a) Recognition in the candidate protocol. Dashed line: chance with four candidates. (b) F1 and (c) healthy refusal in the known-target protocol. Numbers as in Table~\ref{tab:s_family}.}
\label{fig:s_family}
\end{figure*}

\begin{table}[t]
\centering\footnotesize
\setlength{\tabcolsep}{4pt}
\begin{tabular}{@{}lcccc@{}}
\toprule
 & & Recog. & Joint & Healthy \\
Configuration & F1 & (\%) & (\%) & ref.\ (\%) \\
\midrule
\multicolumn{5}{@{}l}{\emph{Qwen3-VL-2B}} \\
\quad Grounding only & 0.410 & 36.5 & 22.8 & 0.2 \\
\quad Mixed & 0.438 & 67.9 & 43.2 & 0.0 \\
\quad Mixed + healthy & 0.445 & 66.3 & 41.4 & 94.7 \\
\midrule
\multicolumn{5}{@{}l}{\emph{Gemma 4 E4B}} \\
\quad Grounding only & 0.451 & 11.9 & 7.3 & 0.0 \\
\quad Mixed & 0.456 & 67.5 & 45.4 & 0.0 \\
\quad Mixed + healthy & 0.432 & 65.2 & 42.8 & 94.7 \\
\bottomrule
\end{tabular}
\caption{\textbf{Two model families on the 20\% subsample.} Benchmark of 1,478 images. F1 and healthy refusal: known-target protocol. Recog.\ and Joint: recognition and joint accuracy in the candidate protocol. All configurations are trained on the 20\% subsample, with 11.9\% healthy negatives in mixed + healthy. Single runs.}
\label{tab:s_family}
\end{table}

\textbf{Second model family.} Gemma~4 E4B~\citep{gemma4_2026} is trained on the subsample in three configurations: grounding only, mixed, and mixed + healthy. The LoRA settings, the learning rate, the schedule, the batch size, and the single epoch are those of \suppref{supp:training}, and the vision encoder is frozen. The training answers use the box format of this model family, and its answers are converted to the format of the benchmark before scoring. Evaluation uses the instructions of \suppref{supp:protocols} with greedy decoding and at most 512 new tokens. Each configuration is a single run.

\textbf{Results in two families.} Table~\ref{tab:s_family} and Figure~\ref{fig:s_family} give both families on the benchmark. After grounding-only training, Gemma~4 E4B names the correct target on 11.9\% of the positive images, and 71.9\% of its answers list all four candidates. With mixed instructions it reaches 67.5\%. Its healthy refusal is 0.0\% without healthy negatives and 94.7\% with them. These are the values that the paper reports. On this subsample, Qwen3-VL-2B reaches 36.5\% after grounding-only training and 67.9\% with mixed instructions. Its mixed + healthy configuration is the model that the paper compares with the fine-tuned detector (F1 0.445, recognition 66.3\%, joint accuracy 41.4\%).

\begin{table*}[t]
\centering\small
\setlength{\tabcolsep}{5pt}
\begin{tabular}{lp{0.70\textwidth}}
\toprule
Models & Sentence added to the instruction \\
\midrule
Gemini 4 Argon & The JSON is a list. Each entry has \texttt{"box\_2d"}: [ymin, xmin, ymax, xmax], normalized to 0-1000, and \texttt{"label"}. Answer with the JSON only. \\
\addlinespace
GPT-6 Astra, GPT-6.1 Sol & The JSON is a list. Each entry has \texttt{"bbox\_2d"}: [xmin, ymin, xmax, ymax], normalized to [0,1], and \texttt{"label"}. Answer with the JSON only. \\
\addlinespace
Claude Opus 5.5, Claude Fable 5.2 & The JSON is a list. Each entry has \texttt{"bbox\_2d"}: [x1, y1, x2, y2] in pixel coordinates of this image, which is \textit{w} pixels wide and \textit{h} pixels high, and \texttt{"label"}. Answer with the JSON only. \\
\bottomrule
\end{tabular}
\caption{\textbf{Box format of the closed models.} One sentence is added to each evaluation instruction. \textit{w} and \textit{h}: width and height of the image as sent. Qwen3-VL-235B answers in the format of the smaller Qwen3-VL models and receives no added sentence.}
\label{tab:s_api_format}
\end{table*}

\textbf{Models queried through an API.} Qwen3-VL-235B and five closed models are queried zero-shot through an API in November 2026, and each model is run once. A model receives 3,929 queries: the known-target and the candidate query for all 1,478 images, and the wrong-name query for the 973 positive images. The instructions and the wrong names are those of our own models. Each query is one message with the image followed by the instruction, without a system message. Qwen3-VL-235B answers in the box format of the smaller Qwen3-VL models. It is decoded greedily with at most 256 new tokens, and its answers are read with the same parser. A closed model has no fixed box format, so one sentence that states the format is added to the instruction (Table~\ref{tab:s_api_format}). The format of each model was fixed before the run. The closed models run with the settings of their providers and an output limit of 32,768 tokens. Images are sent as JPEG files of quality 95. The 51 images above the input limit of one provider are reduced once, and every model receives the same file.

\begin{table*}[t]
\centering\small
\setlength{\tabcolsep}{6pt}
\begin{tabular}{lcccccc}
\toprule
 & & & & & Healthy & False \\
Model & F1 [95\% CI] & R & P & Lesion F1 & ref.\ (\%) & ref.\ (\%) \\
\midrule
Qwen3-VL-235B & 0.411 [0.376, 0.449] & 0.301 & 0.650 & 0.341 & 56.4 & 1.4 \\
\midrule
Gemini 4 Argon & 0.499 [0.466, 0.533] & 0.397 & 0.674 & 0.452 & 92.5 & 10.2 \\
GPT-6 Astra & 0.520 [0.481, 0.562] & 0.443 & 0.630 & 0.473 & 43.4 & 1.2 \\
GPT-6.1 Sol & 0.531 [0.493, 0.571] & 0.462 & 0.623 & 0.480 & 39.0 & 0.9 \\
Claude Opus 5.5 & 0.561 [0.531, 0.593] & 0.546 & 0.577 & 0.535 & 92.7 & 7.2 \\
Claude Fable 5.2 & 0.597 [0.570, 0.625] & 0.615 & 0.579 & 0.578 & 89.1 & 4.6 \\
\bottomrule
\end{tabular}
\caption{\textbf{Models queried through an API, known-target protocol.} Zero-shot on the 1,478 benchmark images (973 positive, 505 healthy). First row: open model. Other rows: closed models. R and P: recall and precision at IoU 0.5. Lesion F1: the 508 lesion-level images. Healthy ref.: healthy refusal. False ref.: false refusal. F1, its interval, lesion F1, and the two refusals are the values of Table~2 of the paper.}
\label{tab:s_api_given}
\end{table*}

\begin{table}[t]
\centering\footnotesize
\setlength{\tabcolsep}{4.5pt}
\begin{tabular}{@{}lrrrr@{}}
\toprule
 & & With & & No \\
Query & Queries & boxes & Empty & answer \\
\midrule
\multicolumn{5}{@{}l}{\emph{Qwen3-VL-235B}} \\
\quad Known target & 1,478 & 1,179 & 299 & 0 \\
\quad Candidate & 1,478 & 1,305 & 173 & 0 \\
\quad Wrong name & 973 & 722 & 251 & 0 \\
\addlinespace[2pt]
\multicolumn{5}{@{}l}{\emph{Gemini 4 Argon}} \\
\quad Known target & 1,478 & 908 & 561 & 9 \\
\quad Candidate & 1,478 & 926 & 552 & 0 \\
\quad Wrong name & 973 & 207 & 753 & 13 \\
\addlinespace[2pt]
\multicolumn{5}{@{}l}{\emph{GPT-6 Astra}} \\
\quad Known target & 1,478 & 1,247 & 231 & 0 \\
\quad Candidate & 1,478 & 1,288 & 190 & 0 \\
\quad Wrong name & 973 & 715 & 258 & 0 \\
\addlinespace[2pt]
\multicolumn{5}{@{}l}{\emph{GPT-6.1 Sol}} \\
\quad Known target & 1,478 & 1,272 & 206 & 0 \\
\quad Candidate & 1,478 & 1,329 & 149 & 0 \\
\quad Wrong name & 973 & 744 & 229 & 0 \\
\addlinespace[2pt]
\multicolumn{5}{@{}l}{\emph{Claude Opus 5.5}} \\
\quad Known target & 1,478 & 940 & 538 & 0 \\
\quad Candidate & 1,478 & 1,028 & 450 & 0 \\
\quad Wrong name & 973 & 242 & 731 & 0 \\
\addlinespace[2pt]
\multicolumn{5}{@{}l}{\emph{Claude Fable 5.2}} \\
\quad Known target & 1,478 & 983 & 495 & 0 \\
\quad Candidate & 1,478 & 1,130 & 348 & 0 \\
\quad Wrong name & 973 & 313 & 660 & 0 \\
\bottomrule
\end{tabular}
\caption{\textbf{Answers of the models queried through an API.} Number of queries by kind of answer. With boxes: an answer with at least one box. Empty: no box. No answer: the provider blocked the answer, which counts against the model.}
\label{tab:s_api_requests}
\end{table}

\textbf{Scoring of the API models.} The boxes are converted to the 0--1000 grid and scored by the rules of \suppref{supp:protocols}. In the candidate protocol, the name of an answer of a closed model is the candidate that its label equals or contains. Requiring the label to equal a candidate gives the same recognition for all five closed models. A query without an answer counts against the model. On a positive image it counts as a refusal of the true target, and on a healthy image or with a wrong name it counts as not refused. Table~\ref{tab:s_api_requests} gives the answers of each model.

\textbf{Results of the API models.} Table~\ref{tab:s_api_given} adds recall and precision to the values of Table~2 of the paper. The candidate protocol of these models is in \suppref{supp:results}, and their refusal of wrong names is in \suppref{supp:refusal}.

\section{Detailed Results}
\label{supp:results}

This section adds detail to Tables~2 and~3 of the paper (Section~5.2 of the paper): the candidate protocol with intervals, the share of healthy negatives, and the 2B and 8B models. The metrics are defined in \suppref{supp:protocols}, and the training settings are in \suppref{supp:training}.

\begin{table*}[t]
\centering\small
\setlength{\tabcolsep}{8pt}
\begin{tabular}{lccc}
\toprule
Model & Recognition (\%) [95\% CI] & Pos.-1 (\%) & Joint (\%) [95\% CI] \\
\midrule
GroundingDINO, raw prompts$^\dagger$~\citep{liu2024groundingdino} & 27.5 [24.9, 30.5] & -- & 17.5 [15.2, 20.0] \\
VisionReasoner-7B$^\dagger$~\citep{liu2026visionreasoner} & 40.7 [37.7, 43.9] & -- & 26.8 [24.1, 29.8] \\
Qwen3-VL-2B, zero-shot & 51.7 [48.5, 54.8] & 33.8 & 34.3 [31.4, 37.2] \\
Qwen3-VL-8B, zero-shot & 64.3 [61.3, 67.4] & 26.5 & 44.5 [41.4, 47.6] \\
Qwen3-VL-235B, zero-shot$^\S$ & 73.6 [70.7, 76.3] & 27.9 & 52.7 [49.6, 55.8] \\
\midrule
Ours 2B, grounding only & 29.2 [26.5, 32.2] & 88.8 & 19.2 [16.8, 21.9] \\
Ours 2B, mixed & 72.7 [69.6, 75.4] & 26.7 & 43.3 [40.3, 46.4] \\
Ours 2B, mixed + healthy \\
\quad mean $\pm$ std, three runs & 73.7 $\pm$ 1.1 & 24.9 $\pm$ 0.7 & 43.5 $\pm$ 1.2 \\
\quad seed-42 run & 74.7 [71.9, 77.5] & 25.1 & 44.8 [41.6, 48.0] \\
Ours 2B, + RL \\
\quad mean $\pm$ std, three runs & 71.0 $\pm$ 1.5 & 25.0 $\pm$ 1.4 & 43.6 $\pm$ 1.4 \\
\quad first RL run & 69.3 [66.4, 72.3] & 24.6 & 43.1 [40.0, 46.2] \\
Ours 8B, mixed + healthy & 78.2 [75.7, 80.7] & 25.7 & 48.6 [45.5, 51.7] \\
\midrule
Gemini 4 Argon$^\S$ & 80.8 [78.5, 83.3] & 26.8 & 60.5 [57.5, 63.7] \\
GPT-6 Astra$^\S$ & 78.8 [76.4, 81.3] & 27.0 & 56.0 [53.1, 59.3] \\
GPT-6.1 Sol$^\S$ & 79.5 [77.0, 82.0] & 27.5 & 57.7 [54.6, 61.0] \\
Claude Opus 5.5$^\S$ & 80.2 [77.7, 82.9] & 26.8 & 64.9 [61.9, 68.1] \\
Claude Fable 5.2$^\S$ & 84.8 [82.6, 87.1] & 26.8 & 70.7 [67.9, 73.6] \\
\bottomrule
\end{tabular}
\caption{\textbf{Results in the candidate protocol.} The 973 positive images of the benchmark, with four candidates per image (chance 25\%). Recognition and joint accuracy have 95\% bootstrap intervals over images, except in the two mean rows. Pos.-1: share of in-list answers that name the first listed candidate. Mean $\pm$ std: mean and sample standard deviation of three training runs. All other rows are single runs. Ours 2B and Ours 8B, mixed + healthy use 12\% healthy negatives. $^\dagger$Adapted protocol (\suppref{supp:baselines}). $^\S$Zero-shot through an API. --: Pos.-1 is not given for the adapted protocol.}
\label{tab:s_recog}
\end{table*}

\textbf{Candidate protocol.} Table~\ref{tab:s_recog} gives recognition, joint accuracy, and the share of in-list answers on the first candidate (Pos.-1). Its rows are models of Table~2 of the paper, including those queried through an API, and the configurations Grounding only and Mixed of Table~3 of the paper. It adds intervals and Pos.-1 to the values of Table~2 of the paper. Two rows give the mean and the sample standard deviation of three training runs, the main model (seeds 42, 1, and 2) and RL. The seed-42 run and the first RL run are the single runs that Table~2 of the paper gives below the two means. Two rows use the adapted protocol of Table~2 of the paper and have no Pos.-1: GroundingDINO with raw prompts and VisionReasoner-7B. The adaptation of each of them is described in \suppref{supp:baselines}. Recognition, Pos.-1, and joint accuracy of the control and of the final RL checkpoint are in \suppref{supp:rl}.

\begin{table*}[t]
\centering\small
\setlength{\tabcolsep}{6pt}
\begin{tabular}{lcccccc}
\toprule
 & \multicolumn{4}{c}{Known-target protocol} & \multicolumn{2}{c}{Candidate protocol} \\
\cmidrule(lr){2-5}\cmidrule(l){6-7}
Configuration & Healthy ref.\ (\%) [95\% CI] & False ref.\ (\%) & F1 & RIoU & Recog.\ (\%) & Joint (\%) \\
\midrule
Mixed & 0.2 [0.0, 0.6] & 0.0 & 0.435 & 0.616 & 72.7 & 43.3 \\
Mixed + healthy (6\%) & 93.3 [90.9, 95.4] & 5.1 & 0.433 & 0.585 & 73.1 & 44.3 \\
Mixed + healthy (12\%) & 95.0 [93.0, 96.8] & 4.5 & 0.440 & 0.578 & 74.7 & 44.8 \\
\bottomrule
\end{tabular}
\caption{\textbf{Share of healthy negatives.} The mixed configuration trained with 0\% (Mixed), 6\%, and 12\% healthy negatives among its training examples, on the benchmark of 1,478 images (973 positive, 505 healthy). Single runs. Mixed + healthy (12\%) is the seed-42 run of the main model. Healthy ref.: healthy refusal. False ref.: false refusal. RIoU: region IoU. Recog.: recognition.}
\label{tab:s_healthyneg}
\end{table*}

\textbf{Share of healthy negatives.} Table~\ref{tab:s_healthyneg} gives the mixed configuration with 0\%, 6\%, and 12\% healthy negatives among the training examples. Each row is a single run. In the known-target protocol, healthy refusal is 0.2\% without healthy negatives, 93.3\% with 6\%, and 95.0\% with 12\%. False refusal is 0.0\%, 5.1\%, and 4.5\%. For the seed-42 run, as the paper states, instance F1 does not fall (0.440 against 0.435 without negatives), but region IoU does (0.616 to 0.578, paired $-$0.039 [$-$0.053, $-$0.027]).

\begin{table*}[t]
\centering\small
\setlength{\tabcolsep}{5pt}
\begin{tabular}{lccccc}
\toprule
 & \multicolumn{3}{c}{Known-target protocol} & \multicolumn{2}{c}{Candidate protocol} \\
\cmidrule(lr){2-4}\cmidrule(l){5-6}
 & & Healthy & False & Recognition (\%) & Joint (\%) \\
Model & F1 [95\% CI] & ref.\ (\%) & ref.\ (\%) & [95\% CI] & [95\% CI] \\
\midrule
Ours 2B, mixed + healthy, seed-42 run & 0.440 [0.412, 0.474] & 95.0 & 4.5 & 74.7 [71.9, 77.5] & 44.8 [41.6, 48.0] \\
Ours 8B, mixed + healthy & 0.429 [0.399, 0.460] & 96.0 & 2.8 & 78.2 [75.7, 80.7] & 48.6 [45.5, 51.7] \\
\midrule
8B minus 2B, paired & $-$0.011 [$-$0.027, +0.005] & -- & -- & +3.5 [+0.6, +6.1] & +3.8 [+1.2, +6.5] \\
\bottomrule
\end{tabular}
\caption{\textbf{The same recipe on Qwen3-VL-2B and Qwen3-VL-8B.} Mixed + healthy with 12\% healthy negatives, on the benchmark of 1,478 images (973 positive, 505 healthy). The 2B row is the seed-42 run of the main model. Healthy ref.: healthy refusal. False ref.: false refusal. Paired: difference on the same images, with a 95\% bootstrap interval that resamples all benchmark images. Recognition and joint accuracy differences are in points. --: not computed.}
\label{tab:s_8b}
\end{table*}

\textbf{2B and 8B.} Table~\ref{tab:s_8b} applies the same recipe, mixed + healthy with 12\% healthy negatives, to Qwen3-VL-2B and Qwen3-VL-8B. The 2B row is the seed-42 run of the main model, and the 8B row is a single run. As the paper states, the 8B model recognizes more than the 2B (78.2\% against 74.7\%, paired +3.5 points [+0.6, +6.1]) at a similar F1 ($-$0.011 [$-$0.027, +0.005]). Joint accuracy is 48.6\% against 44.8\% (paired +3.8 points [+1.2, +6.5]).

\section{Automatic Boxes and Expert Labels}
\label{supp:scale}

This section gives the details of two paragraphs of Section~5.2 of the paper: the fine-tuned model against its teacher, and automatic boxes against expert labels. The metrics are those of \suppref{supp:protocols}, and the training settings are those of \suppref{supp:training}.

\begin{table*}[t]
\centering\small
\setlength{\tabcolsep}{6pt}
\begin{tabular}{@{}lccc@{}}
\toprule
Model & F1 [95\% CI] & Lesion F1 & Object F1 \\
\midrule
GroundingDINO \\
\quad Raw prompts & 0.400 [0.369, 0.436] & 0.323 & 0.607 \\
\quad Taxonomy prompts & 0.395 [0.375, 0.418] & 0.358 & 0.476 \\
\quad Pipeline configuration (teacher) & 0.392 [0.362, 0.427] & 0.282 & 0.701 \\
\midrule
Ours 2B, main model (seed-42 run) & 0.440 [0.412, 0.474] & 0.392 & 0.613 \\
Ours 2B, + RL (first RL run) & 0.477 [0.447, 0.509] & 0.439 & 0.617 \\
\bottomrule
\end{tabular}
\caption{\textbf{The teacher and the fine-tuned models on the benchmark.} Known-target protocol on the 973 positive images. F1 counts instances matched at IoU 0.5, pooled over images, with a 95\% bootstrap interval over images. Lesion F1 and Object F1: the 508 lesion-level and the 465 object-level images. Main model: mixed + healthy (12\%). The pipeline configuration is the teacher, the configuration that labeled the corpus.}
\label{tab:s_teacher}
\end{table*}

\textbf{The teacher and the fine-tuned models.} Table~\ref{tab:s_teacher} compares the three GroundingDINO configurations of Table~2 of the paper with the main model (seed-42 run) and the first RL run. The teacher is the pipeline configuration: the taxonomy prompts with the geometry filters of the pipeline, which is the configuration that labeled the corpus. The three configurations reach F1 0.400, 0.395, and 0.392, within 0.008 of each other. The main model reaches 0.440, and the first RL run 0.477.

\begin{table}[t]
\centering\footnotesize
\setlength{\tabcolsep}{4pt}
\begin{tabular}{@{}lc@{}}
\toprule
Difference in F1 & [95\% CI] \\
\midrule
\multicolumn{2}{@{}l}{\emph{Main model (seed-42 run) minus GroundingDINO}} \\
\quad Raw prompts & +0.040 [+0.017, +0.063] \\
\quad Taxonomy prompts & +0.045 [+0.021, +0.068] \\
\quad Pipeline configuration, teacher & +0.048 [+0.021, +0.075] \\
\multicolumn{2}{@{}l}{\emph{First RL run minus GroundingDINO}} \\
\quad Raw prompts & +0.077 [+0.052, +0.103] \\
\quad Taxonomy prompts & +0.082 [+0.059, +0.104] \\
\quad Pipeline configuration, teacher & +0.085 [+0.055, +0.113] \\
\bottomrule
\end{tabular}
\caption{\textbf{Fine-tuned models minus each GroundingDINO configuration.} Known-target protocol, paired differences in F1 on the same images with 95\% bootstrap intervals. For the differences against the teacher the bootstrap resamples the positive images, and for the others all benchmark images.}
\label{tab:s_teacher_paired}
\end{table}

\textbf{Paired differences.} Table~\ref{tab:s_teacher_paired} gives the paired differences in F1. The main model exceeds the three configurations by +0.040 to +0.048, and the first RL run by +0.077 to +0.085. Every interval lies above zero.

\begin{table}[t]
\centering\footnotesize
\setlength{\tabcolsep}{3pt}
\begin{tabular}{@{}lrccccc@{}}
\toprule
 & & & & & Empty & \\
Images & $n$ & P & R & F1 & (\%) & Boxes \\
\midrule
All positive images & 973 & 0.584 & 0.295 & 0.392 & 0.5 & 1.80 \\
\quad Lesion-level & 508 & 0.517 & 0.194 & 0.282 & 0.4 & 2.06 \\
\quad\quad Dense & 180 & 0.575 & 0.135 & 0.218 & 0.6 & 2.86 \\
\quad Object-level & 465 & 0.685 & 0.718 & 0.701 & 0.6 & 1.51 \\
\bottomrule
\end{tabular}
\caption{\textbf{The teacher by level.} GroundingDINO, pipeline configuration (teacher), scored against the boxes of the benchmark. Known-target protocol on the positive images, instances matched at IoU 0.5 and pooled over the images of a group. P and R: precision and recall. Dense: lesion-level images with five or more reference boxes. Empty: share of images on which the teacher returns no box. Boxes: mean number of teacher boxes per image.}
\label{tab:s_teacher_levels}
\end{table}

\textbf{Lesion level and object level.} As the paper states, the margin comes from lesion-level images, where the labels of the teacher are incomplete. On the 508 lesion-level images, the main model reaches F1 0.392 and the teacher 0.282. There the teacher has recall 0.194 at precision 0.517 (Table~\ref{tab:s_teacher_levels}). On the 465 object-level images the teacher leads (0.701 against 0.613). The values of the paper on the dense images, on the other images, and at IoU 0.75 are in \supptab{tab:s_variants}.

\textbf{Expert-label training set.} The expert masks come from the training split of PlantSeg \citep{wei2024plantseg}. Training images of PlantSeg within perceptual-hash distance 6 of a benchmark image are removed first, and images with an empty mask are skipped. Each mask is split into connected components, and a component is dropped when its bounding box covers less than 0.1\% of the image. An image without a remaining component is skipped. If the largest remaining box covers more than half of the image, the example is at object level and has one box around the whole mask. Otherwise it is at lesion level and keeps the boxes of the components, at most the 30 largest. Boxes are written on the 0--1000 grid of the model, with the disease name of PlantSeg as label. This gives 9,120 examples in the known-target format.

\textbf{Equal-size subset of our boxes.} The subset is drawn at random (seed 11) from the 635,977 examples of the grounding-only configuration, without a condition on source or level. The examples are not changed. As the paper states, the benchmark shares its sources with our boxes.

\textbf{Training.} Both models on 9,120 examples are Qwen3-VL-2B with the LoRA adapter and the supervised settings of the main runs (\suppref{supp:training}), except that they take three epochs instead of one. Each of the two is a single training run.

\begin{table}[t]
\centering\footnotesize
\setlength{\tabcolsep}{4pt}
\begin{tabular}{@{}lrcc@{}}
\toprule
Training data & Examples & Epochs & F1 [95\% CI] \\
\midrule
PlantSeg expert masks & 9,120 & 3 & 0.375 [0.343, 0.409] \\
Our boxes, random subset & 9,120 & 3 & 0.371 [0.339, 0.406] \\
\midrule
Our boxes, full corpus & 635,977 & 1 & 0.425 [0.395, 0.458] \\
\bottomrule
\end{tabular}
\caption{\textbf{Expert labels against automatic boxes at equal size, and the full corpus.} Known-target protocol on the 973 positive images of the benchmark. Every row is Qwen3-VL-2B trained with the grounding-only recipe in a single training run. Our boxes: the automatic boxes of the corpus. Full corpus: the grounding-only configuration of Table~3 of the paper. F1 counts instances matched at IoU 0.5, with a 95\% bootstrap interval over images.}
\label{tab:s_scaling}
\end{table}

\textbf{Results at equal size.} Table~\ref{tab:s_scaling} gives the results. Trained on 9,120 examples, the grounding-only recipe reaches F1 0.375 with the expert masks and 0.371 with the automatic boxes. The paired difference, automatic boxes minus expert masks, is $-$0.004 [$-$0.023, +0.015]. As the paper reports, the full corpus adds 0.054 [0.036, 0.072] to the subset of 9,120 examples. The model trained on the expert masks is also scored on test images of PlantSeg in \suppref{supp:external}.

\clearpage\clearpage
\section{Recognition}
\label{supp:recognition}

This section gives the details of the recognition findings of Section~5.2 of the paper. The candidate protocol, the candidate lists, and the rule that reads the name of an answer are defined in \suppref{supp:protocols}. Recognition is measured on the 973 positive images of the benchmark, where chance is 25\%.

\begin{table}[t]
\centering\footnotesize
\setlength{\tabcolsep}{3pt}
\begin{tabular}{@{}lrrrrrrrr@{}}
\toprule
 & Recog. & \multicolumn{4}{c}{Answers at position} & No & Off- & Pos.-1 \\
\cmidrule(lr){3-6}
Model & (\%) & 1 & 2 & 3 & 4 & name & list & (\%) \\
\midrule
Zero-shot 2B & 51.7 & 301 & 164 & 250 & 175 & 83 & 0 & 33.8 \\
Grounding only & 29.2 & 790 & 36 & 40 & 24 & 6 & 77 & 88.8 \\
Mixed & 72.7 & 258 & 233 & 237 & 237 & 1 & 7 & 26.7 \\
Main model & 74.7 & 240 & 241 & 235 & 241 & 8 & 8 & 25.1 \\
+ RL & 69.3 & 233 & 238 & 239 & 239 & 13 & 11 & 24.6 \\
\midrule
True name & & 248 & 235 & 262 & 228 & & & \\
\bottomrule
\end{tabular}
\caption{\textbf{Answers by position of the named candidate.} Candidate protocol on the 973 positive images. Recog.: recognition. Columns 1 to 4: answers that name the candidate at this position of the list. No name: the other answers, that is, empty lists and answers from which no name can be read. Off-list: answers that name a target outside the list. Pos.-1: answers at position 1 as a share of the answers at positions 1 to 4. Last row: number of lists with the true name at each position. Main model: seed-42 run. + RL: first RL run.}
\label{tab:s_recog_positions}
\end{table}

\textbf{Answers by position.} Table~\ref{tab:s_recog_positions} counts the answers of five models in the candidate protocol by the position of the named candidate in the list. The true name is at each of the four positions in 228 to 262 of the 973 lists. The share of first-candidate answers in Table~3 and Figure~4 of the paper is the count at position 1 divided by the number of answers at positions 1 to 4 (in-list answers). For the grounding-only model this is 790 of 890 answers (88.8\%), and for the zero-shot model 301 of 890 (33.8\%). For the mixed model it is 258 of 965 (26.7\%). For the five models of the table, the shares of the other positions in Figure~4 of the paper follow in the same way. The grounding-only model names a target outside the list in 77 answers, and the mixed model in 7.

\begin{table*}[t]
\centering\small
\setlength{\tabcolsep}{4.5pt}
\begin{tabular}{@{}lcccccccccc@{}}
\toprule
 & \multicolumn{2}{c}{All 973 images} & \multicolumn{8}{c}{By position of the true name in the list} \\
\cmidrule(lr){2-3}\cmidrule(l){4-11}
 & & & \multicolumn{2}{c}{1 (248)} & \multicolumn{2}{c}{2 (235)} & \multicolumn{2}{c}{3 (262)} & \multicolumn{2}{c}{4 (228)} \\
\cmidrule(lr){4-5}\cmidrule(lr){6-7}\cmidrule(lr){8-9}\cmidrule(l){10-11}
Model & LR [95\% CI] & CP & LR & CP & LR & CP & LR & CP & LR & CP \\
\midrule
Zero-shot 2B & 57.1 [54.0, 60.2] & 51.7 & 55.6 & 61.3 & 54.9 & 39.1 & 59.2 & 56.9 & 58.8 & 48.2 \\
Grounding only & 43.0 [40.0, 45.9] & 29.2 & 46.2 & 89.5 & 39.9 & 7.2 & 41.0 & 11.5 & 45.0 & 6.6 \\
Mixed & 61.4 [58.3, 64.5] & 72.7 & 64.9 & 77.4 & 56.7 & 71.1 & 61.8 & 71.4 & 61.8 & 70.6 \\
\bottomrule
\end{tabular}
\caption{\textbf{Names queried one at a time.} Recognition (\%) on the 973 positive images. LR: least refused name. Each of the four names of the candidate list is asked alone in the known-target wording, and the name with the lowest empty-answer probability is the choice of the model. CP: candidate protocol, with the four names listed in one query. Both read-outs use the same images and the same four names. The right part groups the images by the position of their true name in the list (number of images in brackets). Chance is 25\%.}
\label{tab:s_candchoice}
\end{table*}

\textbf{Names queried one at a time.} The candidate protocol asks for one choice among four listed names. A second read-out needs no list. Each of the four names of an image is asked alone in the known-target wording, and the empty-answer probability of each query is read as in \suppref{supp:protocols}. The name with the lowest empty-answer probability is the choice of the model (least refused name). The names are those of the candidate list, and all 973 positive images have their true name and three distractors. Table~\ref{tab:s_candchoice} gives both read-outs on the same images for the three models that the paper compares. As the paper states, recognition also falls when the four names are queried one at a time and the least refused one is taken. The grounding-only model reaches 43.0\% [40.0, 45.9], against 57.1\% for the zero-shot model and 61.4\% for the mixed model. In the candidate protocol, the same three models reach 29.2\%, 51.7\%, and 72.7\%. Paired on the same images, the least refused name of the grounding-only model is right 14.1 points less often than that of the zero-shot model (95\% CI [$-$18.0, $-$10.6]). That of the mixed model is right 18.4 points more often than that of the grounding-only model (95\% CI [+14.9, +21.9]).

\begin{figure}[t]
\centering
\includegraphics[width=\linewidth]{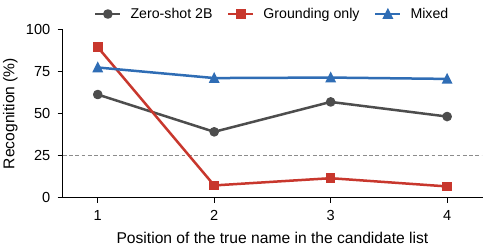}
\caption{\textbf{Recognition in the candidate protocol by the position of the true name.} The 973 positive images are grouped by the position of their true name in the candidate list. Dashed line: chance (25\%). Numbers as in the CP columns of the right part of Table~\ref{tab:s_candchoice}.}
\label{fig:s_recognition}
\end{figure}

\textbf{Position of the true name.} The right part of Table~\ref{tab:s_candchoice} splits both read-outs by the position of the true name in the candidate list, and Figure~\ref{fig:s_recognition} shows the candidate protocol. In the candidate protocol, the grounding-only model is right on 89.5\% of the images whose true name is listed first and on 6.6\% to 11.5\% of the other images. With the least refused name, for which each name is asked alone without a list, it is right on 39.9\% to 46.2\% at each position.

\begin{table*}[t]
\centering\small
\setlength{\tabcolsep}{6pt}
\begin{tabular}{@{}lcccc@{}}
\toprule
 & Real image & Image tiles shuffled & \multicolumn{2}{c}{Blank image} \\
\cmidrule(lr){2-2}\cmidrule(lr){3-3}\cmidrule(l){4-5}
Model & Recognition (\%) & Recognition (\%) & Recognition (\%) & No boxes (\%) \\
\midrule
Mixed & 72.7 [69.6, 75.4] & 69.8 [66.7, 72.6] & 34.0 [31.1, 37.2] & 0.0 \\
Main model & 74.7 [71.9, 77.5] & 71.2 [68.3, 74.1] & 5.7 [4.2, 7.0] & 85.5 \\
+ RL & 69.3 [66.4, 72.3] & 65.0 [61.9, 68.0] & 4.6 [3.3, 6.1] & 88.0 \\
\bottomrule
\end{tabular}
\caption{\textbf{Image controls of the candidate protocol.} Recognition [95\% CI] on the 973 positive images in three conditions. Real image: the image itself. Image tiles shuffled: the image cut into a grid of 4 by 4 tiles in random order. Blank image: a uniform gray image of the same size. The instruction is unchanged. No boxes: share of answers without boxes. Chance is 25\%. Main model: seed-42 run. + RL: first RL run.}
\label{tab:s_image_controls}
\end{table*}

\textbf{Image controls.} Two controls change the image of the candidate query and keep the instruction. With shuffled tiles, the image is cut into a grid of 4 by 4 tiles whose order is permuted at random. With a blank image, the image is replaced by a uniform gray image of the same size. Table~\ref{tab:s_image_controls} gives recognition under both controls. With shuffled tiles, the mixed model reaches 69.8\%, against 72.7\% with the real image, so the global layout is not needed, as the paper states. With a blank image, the mixed model is still right on 34.0\% [31.1, 37.2] of the queries. As the paper states, up to nine points above chance can thus come from the candidate lists alone. The main model and the RL model return no boxes on 85.5\% and 88.0\% of the blank-image queries. Their recognition on the blank image is 5.7\% and 4.6\%.

\textbf{Naming without a candidate list.} In the open-set query, the model names the target itself and boxes it. The instruction is ``What specific disease, pest, or disorder affects this plant? Answer with its common name (for example `septoria leaf spot' or `aphids') and locate all affected regions. Output bounding boxes in JSON format with the specific name as the label. If the plant is healthy, output an empty list.'' The name of an answer is the most frequent label of its boxes, compared with the true target after lowercasing and removing extra spaces. The lenient match also accepts a name that contains the true target or is contained in it. For the five 2B models (zero-shot, grounding only, mixed, the seed-42 run of the main model, and the first RL run), the lenient match is at most 4.6\% of the 973 positive images. As the paper states, the 2B models do not name the disease without a candidate list, so recognition is evaluated with candidates.

\section{Refusal}
\label{supp:refusal}

This section gives the details of the refusal results of Section~5.2 of the paper. The queries, the wrong and random names, refusal, pair accuracy, the empty-answer probability, and the AUROC are defined in \suppref{supp:protocols}. The healthy and the wrong-target negatives and the training sets are described in \suppref{supp:training}.

\begin{table*}[t]
\centering\footnotesize
\setlength{\tabcolsep}{4pt}
\begin{tabular}{@{}lcccccccc@{}}
\toprule
 & \multicolumn{3}{c}{Empty answers (\%)} & & \multicolumn{3}{c}{AUROC, wrong vs.\ true name} & \\
\cmidrule(lr){2-4}\cmidrule(lr){6-8}
Configuration & Healthy & True & Wrong & Pair (\%) [95\% CI] & All [95\% CI] & Same & Rand. & F1 \\
\midrule
\multicolumn{9}{@{}l}{\emph{Zero-shot}} \\
\quad Qwen3-VL-2B & 1.2 & 0.0 & 0.0 & 0.0 [0.0, 0.0] & 0.706 [0.689, 0.721] & -- & -- & 0.319 \\
\quad Qwen3-VL-8B & 30.9 & 2.5 & 15.5 & 9.7 [7.7, 11.5] & 0.771 [0.754, 0.787] & -- & -- & 0.379 \\
\midrule
Grounding only & 0.0 & 0.0 & 0.0 & 0.0 [0.0, 0.0] & 0.637 [0.618, 0.656] & -- & -- & 0.425 \\
Mixed & 0.2 & 0.0 & 0.0 & 0.0 [0.0, 0.0] & 0.666 [0.651, 0.681] & 0.647 & 0.697 & 0.435 \\
Mixed + healthy (6\%) & 93.3 & 5.1 & 25.7 & 17.2 [14.8, 19.7] & 0.795 [0.779, 0.812] & 0.786 & 0.869 & 0.433 \\
\midrule
\multicolumn{9}{@{}l}{Mixed + healthy (12\%), main model} \\
\quad seed-42 run & 95.0 & 4.5 & 30.2 & 19.5 [17.2, 22.0] & 0.823 [0.807, 0.839] & 0.804 & 0.893 & 0.440 \\
\quad seed 1 & 93.3 & 4.0 & 27.4 & 18.0 [15.6, 20.5] & 0.809 [0.793, 0.824] & 0.800 & 0.897 & 0.436 \\
\quad seed 2 & 94.1 & 6.3 & 28.1 & 16.9 [14.6, 19.4] & 0.805 [0.788, 0.821] & 0.798 & 0.881 & 0.432 \\
\quad mean $\pm$ std & 94.1$\pm$0.9 & 4.9$\pm$1.2 & 28.6$\pm$1.5 & 18.1$\pm$1.3 & 0.812$\pm$0.010 & -- & -- & 0.436$\pm$0.004 \\
\quad + wrong-target 2.9\% & 97.0 & 10.4 & 52.1 & 29.9 [26.8, 32.9] & 0.854 [0.838, 0.869] & 0.865 & 0.914 & 0.412 \\
\quad + wrong-target 5.6\% & 96.8 & 13.2 & 59.7 & 31.7 [29.0, 34.5] & 0.844 [0.826, 0.860] & 0.853 & 0.899 & 0.409 \\
\quad + wrong-target 5.6\%, second run & 97.6 & 13.8 & 59.5 & 31.1 [28.3, 34.1] & 0.841 [0.824, 0.856] & 0.863 & 0.892 & 0.410 \\
\quad + wrong-target 10.6\% & 95.8 & 15.7 & 62.0 & 30.3 [27.4, 33.3] & 0.826 [0.810, 0.842] & 0.865 & 0.882 & 0.392 \\
Wrong-target 11.9\% only & 84.8 & 13.8 & 61.9 & 31.6 [28.7, 34.3] & 0.846 [0.830, 0.862] & 0.872 & 0.910 & 0.395 \\
\midrule
+ RL, first RL run & 87.7 & 0.9 & 8.1 & 6.7 [5.3, 8.1] & 0.720 [0.704, 0.736] & 0.707 & 0.794 & 0.477 \\
+ continued SFT (control), first control run & 96.6 & 5.1 & 36.5 & 22.2 [19.8, 25.0] & 0.828 [0.812, 0.844] & 0.808 & 0.887 & 0.458 \\
\midrule
Ours 8B, mixed + healthy & 96.0 & 2.8 & 27.9 & 17.7 [15.5, 20.1] & 0.849 [0.833, 0.865] & 0.826 & 0.907 & 0.429 \\
\bottomrule
\end{tabular}
\caption{\textbf{Refusal of the configurations.} Known-target protocol on the 1,478 benchmark images. Empty answers: share of the 505 healthy images (Healthy), and of the 973 positive images queried with the true target (True, false refusal) or with a wrong name in the same wording (Wrong). Pair: pair accuracy. AUROC, wrong vs.\ true name: AUROC of the empty-answer probability on all 973 positive images (All), and on the 496 images with a same-crop alternative for a same-crop (Same) and a random (Rand.) name. F1: instance F1 at IoU 0.5. Mean $\pm$ std: mean and sample standard deviation of the three runs of the main model. Seed 1 and seed 2: further training runs of the main model. The control and RL start from the seed-42 run. Ours 8B: 12\% healthy negatives. A dash marks a value that was not computed.}
\label{tab:s_refusal_all}
\end{table*}

\textbf{Healthy images.} Table~\ref{tab:s_refusal_all} gives the zero-shot Qwen3-VL-2B and Qwen3-VL-8B, the supervised configurations of \supptab{tab:s_train_sets} with their further runs, Ours 8B, and the first runs of RL and of the control. With 12\% healthy negatives, the main model returns an empty list on 94.1\% of the 505 healthy images (mean of three runs, $\pm$0.9) and on 95.0\% [93.0, 96.8] in the seed-42 run. It refuses the true target on 4.9\% ($\pm$1.2) of the 973 positive images, and on 4.5\% in the seed-42 run. For healthy against positive images, the AUROC of the empty-answer probability is 0.996 on the lesion-level and 0.990 on the object-level images in the seed-42 run. As the paper states, this is about 0.99.

\textbf{Wrong names.} On the wrong-name queries, the main model returns an empty list on 28.6\% ($\pm$1.5) of the positive images, and on 30.2\% [27.5, 33.1] in the seed-42 run. Its pair accuracy is 18.1\% ($\pm$1.3), and 19.5\% in the seed-42 run. The AUROC for wrong against true names is 0.812 ($\pm$0.010), and 0.823 [0.807, 0.839] in the seed-42 run. As the paper states, refusing a wrong name is a different ability. The other runs of RL and of the control, and the other RL configurations, are in \suppref{supp:rl}.

\begin{table}[t]
\centering\footnotesize
\setlength{\tabcolsep}{3pt}
\begin{tabular}{@{}lccccc@{}}
\toprule
 & & & & \multicolumn{2}{c}{Wrong, by kind} \\
\cmidrule(l){5-6}
Model & Healthy & True & Wrong & Same & Other \\
\midrule
Qwen3-VL-235B & 56.4 & 1.4 & 25.8 & 9.5 & 42.8 \\
Gemini 4 Argon & 92.5 & 10.2 & 77.4 & 64.9 & 90.4 \\
GPT-6 Astra & 43.4 & 1.2 & 26.5 & 7.9 & 45.9 \\
GPT-6.1 Sol & 39.0 & 0.9 & 23.5 & 7.1 & 40.7 \\
Claude Opus 5.5 & 92.7 & 7.2 & 75.1 & 58.9 & 92.0 \\
Claude Fable 5.2 & 89.1 & 4.6 & 67.8 & 51.4 & 84.9 \\
\bottomrule
\end{tabular}
\caption{\textbf{Refusal of the models queried through an API.} Zero-shot, known-target protocol, in percent. Healthy, True, and Wrong as in Table~\ref{tab:s_refusal_all}. Same: wrong-name refusal on the 496 positive images with a same-crop wrong name. Other: on the other 477 positive images. Wrong-name refusal is counted on generated answers, and these models have no AUROC (\suppref{supp:protocols}).}
\label{tab:s_refusal_api}
\end{table}

\textbf{Models queried through an API.} Table~\ref{tab:s_refusal_api} gives the refusal of the five closed models and of Qwen3-VL-235B, which are queried through an API without training. Three closed models refuse 89.1\% to 92.7\% of the healthy images and 67.8\% to 77.4\% of the wrong names. The other two refuse 39.0\% to 43.4\% and 23.5\% to 26.5\%. All five and the 235B open model refuse the wrong name less often for the 496 same-crop names (7.1\% to 64.9\%) than for the other 477 (40.7\% to 92.0\%). The queries and the scoring of these models are described in \suppref{supp:baselines}.

\begin{table*}[t]
\centering\small
\setlength{\tabcolsep}{5pt}
\begin{tabular}{@{}lccc@{}}
\toprule
Wrong-name refusal (\%) & Same-crop name & Random name & Random $-$ same-crop \\
\midrule
Mixed & 0.0 [0.0, 0.0] & 0.0 [0.0, 0.0] & +0.0 [+0.0, +0.0] \\
Mixed + healthy (6\%) & 27.0 [23.0, 31.2] & 30.8 [26.8, 35.1] & +3.8 [$-$0.4, +8.3] \\
Main, seed-42 run & 25.4 [21.8, 29.4] & 36.7 [32.7, 40.7] & +11.3 [+6.5, +16.1] \\
Main, seed 1 & 23.4 [19.6, 27.2] & 33.7 [29.6, 37.7] & +10.3 [+5.6, +15.3] \\
Main, seed 2 & 27.2 [23.4, 31.2] & 36.1 [31.9, 40.1] & +8.9 [+4.4, +13.5] \\
Main + wrong-target 2.9\% & 58.3 [53.6, 62.3] & 68.3 [64.3, 72.4] & +10.1 [+4.4, +15.5] \\
Main + wrong-target 5.6\% & 65.9 [61.7, 70.0] & 71.8 [67.7, 75.4] & +5.8 [+0.6, +10.9] \\
Main + wrong-target 5.6\%, second run & 67.1 [63.1, 71.0] & 68.5 [64.3, 72.4] & +1.4 [$-$4.0, +6.7] \\
Main + wrong-target 10.6\% & 73.0 [69.0, 76.8] & 72.2 [68.1, 76.0] & $-$0.8 [$-$6.7, +4.6] \\
Wrong-target 11.9\% only & 68.3 [64.3, 72.0] & 77.4 [73.4, 81.0] & +9.1 [+4.0, +14.1] \\
Main + RL, first RL run & 9.5 [6.9, 12.3] & 11.1 [8.5, 13.7] & +1.6 [$-$1.4, +4.6] \\
Main + continued SFT, first control run & 34.7 [30.4, 38.9] & 42.9 [38.9, 47.2] & +8.3 [+3.6, +13.1] \\
Ours 8B, mixed + healthy & 26.2 [22.4, 30.0] & 38.3 [34.1, 42.7] & +12.1 [+7.3, +17.1] \\
\bottomrule
\end{tabular}

\vspace{4pt}
\begin{tabular}{@{}lccc@{}}
\toprule
AUROC, wrong vs.\ true name & Same-crop name & Random name & Random $-$ same-crop \\
\midrule
Mixed & 0.647 [0.625, 0.668] & 0.697 [0.676, 0.717] & +0.050 [+0.035, +0.068] \\
Mixed + healthy (6\%) & 0.786 [0.759, 0.809] & 0.869 [0.847, 0.888] & +0.083 [+0.062, +0.105] \\
Main, seed-42 run & 0.804 [0.777, 0.827] & 0.893 [0.873, 0.911] & +0.089 [+0.069, +0.111] \\
Main, seed 1 & 0.800 [0.773, 0.822] & 0.897 [0.876, 0.914] & +0.097 [+0.076, +0.121] \\
Main, seed 2 & 0.798 [0.772, 0.821] & 0.881 [0.860, 0.898] & +0.083 [+0.064, +0.104] \\
Main + wrong-target 2.9\% & 0.865 [0.839, 0.885] & 0.914 [0.896, 0.930] & +0.049 [+0.031, +0.071] \\
Main + wrong-target 5.6\% & 0.853 [0.826, 0.874] & 0.899 [0.879, 0.916] & +0.047 [+0.027, +0.067] \\
Main + wrong-target 5.6\%, second run & 0.863 [0.837, 0.884] & 0.892 [0.870, 0.911] & +0.030 [+0.010, +0.051] \\
Main + wrong-target 10.6\% & 0.865 [0.840, 0.885] & 0.882 [0.860, 0.901] & +0.017 [$-$0.004, +0.039] \\
Wrong-target 11.9\% only & 0.872 [0.845, 0.891] & 0.910 [0.890, 0.927] & +0.038 [+0.021, +0.056] \\
Main + RL, first RL run & 0.707 [0.682, 0.732] & 0.794 [0.771, 0.816] & +0.087 [+0.063, +0.112] \\
Main + continued SFT, first control run & 0.808 [0.782, 0.831] & 0.887 [0.866, 0.904] & +0.078 [+0.057, +0.099] \\
Ours 8B, mixed + healthy & 0.826 [0.800, 0.848] & 0.907 [0.888, 0.924] & +0.080 [+0.062, +0.101] \\
\bottomrule
\end{tabular}
\caption{\textbf{Refusal by kind of wrong name.} Known-target protocol on the 496 positive images with a same-crop alternative, with 95\% bootstrap intervals over images. Same-crop name: another benchmark target of the crop of the image. Random name: a second query of the same images with a benchmark target that is not a target of the crop. Random $-$ same-crop: paired difference on these images, in points or in AUROC. AUROC: empty-answer probability for the wrong against the true name on the same images. Main: mixed + healthy (12\%).}
\label{tab:s_refusal_kind}
\end{table*}

\begin{figure*}[t]
\centering
\includegraphics[width=\linewidth]{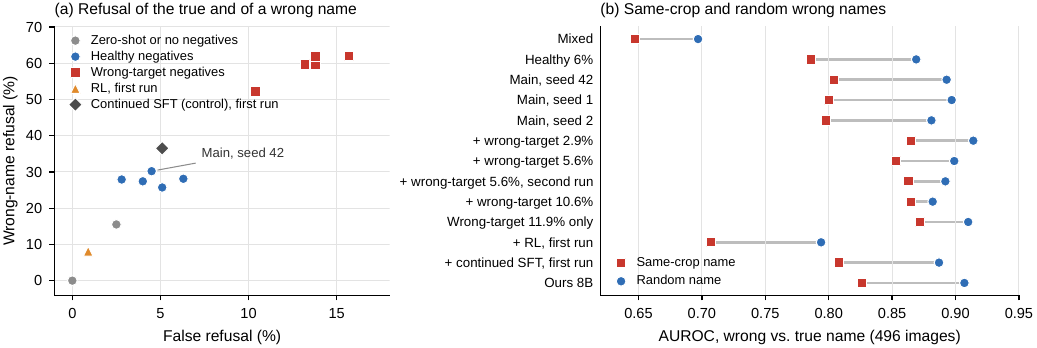}
\caption{\textbf{Refusal of the true and of a wrong name.} (a) Wrong-name refusal against false refusal for the 16 configurations of Table~\ref{tab:s_refusal_all}. The gray point at zero holds three configurations: Qwen3-VL-2B zero-shot, grounding only, and mixed. (b) AUROC of the empty-answer probability for a same-crop and for a random wrong name against the true name, on the same 496 positive images. Main: mixed + healthy (12\%). Healthy 6\%: mixed + healthy (6\%). Numbers as in Tables~\ref{tab:s_refusal_all} and~\ref{tab:s_refusal_kind}.}
\label{fig:s_refusal}
\end{figure*}

\textbf{Kind of wrong name.} The 496 images with a same-crop alternative are also queried with a random name. The two kinds of wrong name are thus compared on the same images (Table~\ref{tab:s_refusal_kind} and Figure~\ref{fig:s_refusal}b). The seed-42 run refuses the same-crop name on 25.4\% and the random name on 36.7\% of these images (paired $+$11.3 points [$+$6.5, $+$16.1]). Its AUROC is 0.804 for same-crop and 0.893 for random names (paired $+$0.089 [$+$0.069, $+$0.111]). On the other 477 positive images, whose wrong name is drawn from all benchmark targets (\suppref{supp:protocols}), it refuses 35.2\% [30.8, 39.6] of the wrong names, with AUROC 0.858 [0.834, 0.880]. Across the configurations of Table~\ref{tab:s_refusal_kind}, the AUROC is 0.017 to 0.097 higher for random than for same-crop names. The interval of this difference excludes zero for every configuration except the 10.6\% configuration. As the paper states, a plausible wrong name is the hard case.

\begin{table*}[t]
\centering\small
\setlength{\tabcolsep}{6pt}
\begin{tabular}{@{}lccc@{}}
\toprule
 & \multicolumn{3}{c}{AUROC gain} \\
\cmidrule(l){2-4}
Wrong-target & Same-crop & Random & Difference \\
\midrule
2.9\% & +0.061 [+0.047, +0.076] & +0.021 [+0.007, +0.035] & +0.040 [+0.025, +0.055] \\
5.6\% & +0.049 [+0.033, +0.066] & +0.006 [$-$0.009, +0.019] & +0.043 [+0.027, +0.059] \\
5.6\%, second run & +0.059 [+0.044, +0.075] & $-$0.001 [$-$0.016, +0.014] & +0.060 [+0.043, +0.076] \\
10.6\% & +0.061 [+0.045, +0.078] & $-$0.011 [$-$0.028, +0.005] & +0.072 [+0.053, +0.091] \\
11.9\% only & +0.067 [+0.052, +0.085] & +0.016 [+0.000, +0.032] & +0.051 [+0.035, +0.068] \\
\bottomrule
\end{tabular}
\caption{\textbf{Gains of the wrong-target negatives in AUROC by kind of wrong name.} Paired differences in the AUROC for wrong against true names, against the main model (seed-42 run), on the same images, with 95\% bootstrap intervals over images. Wrong-target: the main mixture with 2.9\%, 5.6\%, or 10.6\% wrong-target negatives, the second run of the 5.6\% configuration, and wrong-target 11.9\% only. Same-crop and Random: the 496 images with a same-crop alternative. Difference: same-crop gain minus random gain on these images.}
\label{tab:s_refusal_kind_gain}
\end{table*}

\textbf{Wrong-target negatives.} At 2.9\%, 5.6\%, and 10.6\% of the training set, wrong-target negatives raise wrong-name refusal to 52.1\%, 59.7\%, and 62.0\% and pair accuracy to 29.9\%, 31.7\%, and 30.3\% (Table~\ref{tab:s_refusal_all}). The true name is then refused on 10.4\%, 13.2\%, and 15.7\% of the positive images. Against the seed-42 run, F1 changes by $-$0.028 [$-$0.043, $-$0.013], $-$0.032 [$-$0.047, $-$0.017], and $-$0.048 [$-$0.065, $-$0.032]. Recognition changes by $-$2.7 [$-$4.9, $-$0.3], $-$6.7 [$-$8.9, $-$4.6], and $-$7.2 [$-$9.6, $-$4.8] points. On the 496 images, the AUROC of the three configurations rises by 0.061, 0.049, and 0.061 for same-crop names and changes by $+$0.021, $+$0.006, and $-$0.011 for random names (Table~\ref{tab:s_refusal_kind_gain}). For these three configurations, the gain for same-crop names exceeds the gain for random names by 0.040 to 0.072, with intervals that exclude zero.

\textbf{Second run of the 5.6\% configuration.} The 5.6\% configuration was trained once more with another seed. This one further run refuses 59.5\% of the wrong names and 13.8\% of the true names, with pair accuracy 31.1\%, AUROC 0.841, and F1 0.410. The first run has 59.7\%, 13.2\%, 31.7\%, 0.844, and 0.409. On the 496 images, its AUROC gain for same-crop names is $+$0.059, against $+$0.049 for the first run. As the paper states, a second seed of the 5.6\% dose agrees.

\begin{table*}[t]
\centering\small
\setlength{\tabcolsep}{8pt}
\begin{tabular}{@{}lcc@{}}
\toprule
AUROC, wrong vs.\ true name & Six other sources (392 images) & MIRAGE, AgroBench (104 images) \\
\midrule
Main, seed-42 run & 0.851 [0.826, 0.875] & 0.649 [0.592, 0.709] \\
Main + wrong-target 2.9\% & 0.918 [0.898, 0.937] & 0.640 [0.585, 0.699] \\
Main + wrong-target 5.6\% & 0.906 [0.885, 0.925] & 0.626 [0.566, 0.689] \\
Main + wrong-target 5.6\%, second run & 0.920 [0.900, 0.939] & 0.628 [0.574, 0.685] \\
Main + wrong-target 10.6\% & 0.929 [0.910, 0.946] & 0.625 [0.574, 0.679] \\
Wrong-target 11.9\% only & 0.932 [0.913, 0.947] & 0.622 [0.569, 0.677] \\
\bottomrule
\end{tabular}
\caption{\textbf{Separation of same-crop wrong names from true names by source group.} Known-target protocol. AUROC of the empty-answer probability for the same-crop wrong name against the true name, with 95\% bootstrap intervals over images. MIRAGE, AgroBench: the two sources of in-the-wild images. Six other sources: all other benchmark sources. In parentheses: the number of images of each group among the 496 images with a same-crop alternative. Main: mixed + healthy (12\%).}
\label{tab:s_refusal_source}
\end{table*}

\textbf{Source groups.} Table~\ref{tab:s_refusal_source} splits the 496 images with a same-crop alternative into MIRAGE and AgroBench, the two sources of in-the-wild images, and the six other sources. For same-crop names, the AUROC of the seed-42 run is 0.851 on the 392 images of the six other sources and 0.649 on the 104 images of MIRAGE and AgroBench. With the three configurations of wrong-target negatives, it is 0.918, 0.906, and 0.929 on the six other sources, and 0.640, 0.626, and 0.625 on MIRAGE and AgroBench. As the paper states, these six sources supply 98\% of the training samples.

\begin{table}[t]
\centering\footnotesize
\setlength{\tabcolsep}{3pt}
\begin{tabular}{@{}lccccc@{}}
\toprule
 & \multicolumn{3}{c}{Empty answers (\%)} & & \\
\cmidrule(lr){2-4}
Run, system & Healthy & True & Wrong & Pair (\%) [95\% CI] & F1 \\
\midrule
\multicolumn{6}{@{}l}{Seed-42 run} \\
\quad own answers & 95.0 & 4.5 & 30.2 & 19.5 [17.2, 22.0] & 0.440 \\
\quad threshold & 97.6 & 9.9 & 53.9 & 29.4 [26.7, 32.5] & 0.431 \\
\multicolumn{6}{@{}l}{Seed 1} \\
\quad own answers & 93.3 & 4.0 & 27.4 & 18.0 [15.6, 20.5] & 0.436 \\
\quad threshold & 97.2 & 10.4 & 50.2 & 26.6 [23.7, 29.4] & 0.423 \\
\multicolumn{6}{@{}l}{Seed 2} \\
\quad own answers & 94.1 & 6.3 & 28.1 & 16.9 [14.6, 19.4] & 0.432 \\
\quad threshold & 97.0 & 9.8 & 39.1 & 20.6 [18.1, 23.1] & 0.425 \\
\bottomrule
\end{tabular}

\caption{\textbf{Threshold on the empty-answer probability, without retraining.} Known-target protocol, the three runs of the main model. Own answers: the answers of the model. Threshold: the same answers, with every query whose empty-answer probability is above a cross-fitted threshold answered with an empty list (rule in the text, false-refusal budget 10\%). Healthy: 505 healthy images. True, Wrong, Pair, F1: 973 positive images, as in Table~\ref{tab:s_refusal_all}.}
\label{tab:s_threshold}
\end{table}

\textbf{Threshold on the empty-answer probability.} The first alternative without retraining puts a threshold on the empty-answer probability of each known-target query of the main model. A query whose probability is above the threshold is answered with an empty list. A query that the model already answers with an empty list keeps this answer. The threshold is cross-fitted over the positive images. The 973 positive images are split at random (fixed seed) into five folds. For each fold, the threshold is chosen on the other four folds and applied to the true-name and the wrong-name queries of the images of that fold. The only criterion is false refusal: the threshold is the lowest value at which at most 10\% of the true-name queries of the four folds are refused. The candidate values are the empty-answer probabilities of the true-name and the wrong-name queries of these four folds. If the model already refuses more than 10\% of the true names of the four folds, no threshold is applied. The healthy images use the threshold chosen by the same rule on all 973 positive images. A refused true-name query has no boxes, so its reference boxes count as missed in F1.

\textbf{Threshold results.} Table~\ref{tab:s_threshold} gives the three runs of the main model. For the seed-42 run, the threshold raises pair accuracy from 19.5\% to 29.4\% [26.7, 32.5] at F1 0.431, with wrong-name refusal 53.9\% and healthy refusal 97.6\%. For seeds 1 and 2, pair accuracy is 26.6\% and 20.6\%. Each fold uses a threshold chosen on the other folds, so the false refusal of the three runs is 9.9\%, 10.4\%, and 9.8\%. After retraining with wrong-target negatives, pair accuracy is 29.9\% to 31.7\% at F1 0.392 to 0.412 (Table~\ref{tab:s_refusal_all}).

\textbf{Name check.} The second alternative uses the candidate question as a check on each known-target answer. For the true-name query, the model is asked the candidate question on a list of four names that contains the true target and the wrong name of the image. The answer is kept only if the model picks the true target. Otherwise the answer becomes an empty list. For the wrong-name query, the list holds four names without the true target, and the answer is kept only if the model picks the wrong name. Answers that are already empty stay empty. Both queries are in the same wording, and the wrong-name answer is read from the token probabilities (\suppref{supp:protocols}). The name check is not applied to the healthy images. For the seed-42 run, it reaches pair accuracy 37.7\% [34.7, 40.6] and refuses 77.9\% of the wrong names, but it also refuses 26.2\% of the true names, at F1 0.373. This is the row Main + name check of Table~4 of the paper.

\textbf{Wording.} The wrong name can also be asked in another wording than the true name, with the instruction ``Locate the \textit{name} in this image. Output the bounding boxes of the affected regions in JSON format. If it is not present, output an empty list.'' The true-name query keeps the known-target wording of \supptab{tab:s_prompts}. In this other wording, the seed-42 run refuses 15.0\% [12.8, 17.4] of the wrong names, against 30.2\% in the wording of the true-name query. As the paper states, both names are therefore asked in one wording.

\section{RL Results}
\label{supp:rl}

This section gives the single runs behind the RL results of Section~5.2 of the paper, the comparison with the control run by run, and the other RL configurations. The training setup is in \suppref{supp:training}.

\begin{figure*}[t]
\centering
\includegraphics[width=\linewidth]{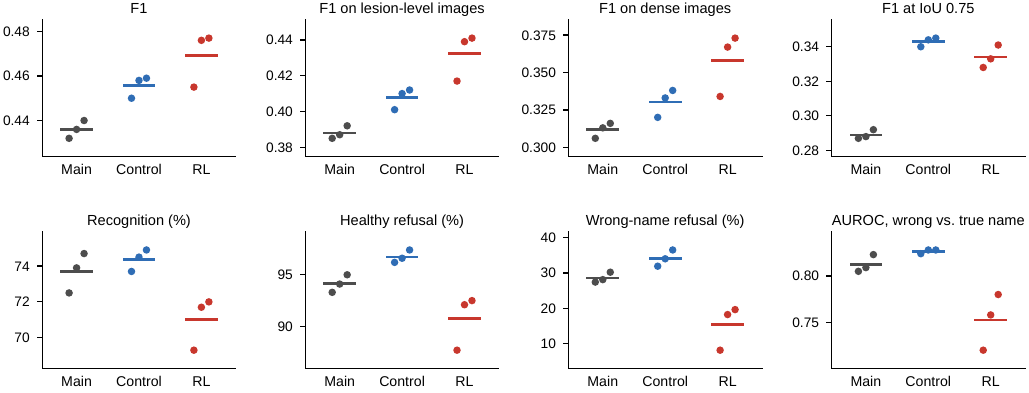}
\caption{\textbf{Every run of the main model, of the control, and of RL.} Each dot is one training run, and the line is the mean of the three runs of a group (Table~\ref{tab:s_rl_groups}). Within a group, the dots are ordered by value. Numbers of the dots as in Table~\ref{tab:s_rl_runs}.}
\label{fig:s_rl_runs}
\end{figure*}

\begin{table*}[t]
\centering\small
\setlength{\tabcolsep}{6pt}
\begin{tabular}{lccccc}
\toprule
Run & F1 [95\% CI] & Lesion F1 & Dense F1 & F1@0.75 & Region IoU \\
\midrule
Mixed + healthy (12\%), main model \\
\quad seed-42 run & 0.440 [0.412, 0.474] & 0.392 & 0.316 & 0.288 & 0.578 \\
\quad seed 1 & 0.436 [0.406, 0.468] & 0.385 & 0.306 & 0.292 & 0.591 \\
\quad seed 2 & 0.432 [0.404, 0.463] & 0.387 & 0.313 & 0.287 & 0.565 \\
\midrule
+ continued SFT (control) \\
\quad first control run & 0.458 [0.429, 0.487] & 0.410 & 0.333 & 0.340 & 0.588 \\
\quad seed 1 & 0.459 [0.430, 0.489] & 0.412 & 0.338 & 0.344 & 0.591 \\
\quad seed 2 & 0.450 [0.422, 0.480] & 0.401 & 0.320 & 0.345 & 0.594 \\
\midrule
+ RL \\
\quad first RL run & 0.477 [0.447, 0.509] & 0.439 & 0.367 & 0.341 & 0.618 \\
\quad seed 1 & 0.455 [0.426, 0.486] & 0.417 & 0.334 & 0.328 & 0.609 \\
\quad seed 2 & 0.476 [0.449, 0.505] & 0.441 & 0.373 & 0.333 & 0.593 \\
\bottomrule
\end{tabular}

\vspace{4pt}
\begin{tabular}{lcccccc}
\toprule
 & \multicolumn{2}{c}{Refusal (\%)} & \multicolumn{2}{c}{Recognition (\%)} & \multicolumn{2}{c}{Wrong names} \\
\cmidrule(lr){2-3}\cmidrule(lr){4-5}\cmidrule(lr){6-7}
Run & Healthy & False & Recog. & Joint & Refusal (\%) & AUROC \\
\midrule
Mixed + healthy (12\%), main model \\
\quad seed-42 run & 95.0 & 4.5 & 74.7 & 44.8 & 30.2 & 0.823 \\
\quad seed 1 & 93.3 & 4.0 & 73.9 & 43.3 & 27.4 & 0.809 \\
\quad seed 2 & 94.1 & 6.3 & 72.5 & 42.4 & 28.1 & 0.805 \\
\midrule
+ continued SFT (control) \\
\quad first control run & 96.6 & 5.1 & 73.7 & 45.2 & 36.5 & 0.828 \\
\quad seed 1 & 96.2 & 4.8 & 74.9 & 44.9 & 31.9 & 0.828 \\
\quad seed 2 & 97.4 & 5.2 & 74.5 & 45.3 & 34.0 & 0.824 \\
\midrule
+ RL \\
\quad first RL run & 87.7 & 0.9 & 69.3 & 43.1 & 8.1 & 0.720 \\
\quad seed 1 & 92.5 & 3.2 & 71.7 & 45.2 & 19.6 & 0.758 \\
\quad seed 2 & 92.1 & 1.8 & 72.0 & 42.5 & 18.2 & 0.780 \\
\bottomrule
\end{tabular}
\caption{\textbf{Every run of the main model, of the continued-SFT control, and of RL.} Benchmark of 1,478 images (973 positive, 505 healthy). Localization and refusal: known-target protocol. Recognition (Recog.) and joint accuracy: candidate protocol. F1: instances matched at IoU 0.5 on the 973 positive images, with a 95\% bootstrap interval over images. Lesion F1: the 508 lesion-level images. Dense F1: the 180 lesion-level images with five or more boxes. F1@0.75: instances matched at IoU 0.75. Healthy: healthy refusal. False: false refusal. Wrong names, Refusal: positive images refused when queried with a wrong name. AUROC: AUROC of the empty-answer probability for wrong against true names (973 positive images). Seed 1 and seed 2: further runs with other training seeds. The control and RL start from the seed-42 run. The first RL run is the RL run of Tables 2 to 5 of the paper, and the first control run is the control of Tables 3 and 4.}
\label{tab:s_rl_runs}
\end{table*}

\textbf{Runs.} Table~\ref{tab:s_rl_runs} and Figure~\ref{fig:s_rl_runs} give every run of the main model, of the control, and of RL. The control and RL start from the seed-42 run of the main model, and their further runs use other training seeds. The first RL run is the RL run of Tables~2 to~5 of the paper, and the first control run is the control of Tables~3 and~4 of the paper.

\begin{table*}[t]
\centering\small
\setlength{\tabcolsep}{8pt}
\begin{tabular}{@{}lcccc@{}}
\toprule
 & \multicolumn{3}{c}{Mean $\pm$ std of three runs} & RL $-$ control \\
\cmidrule(lr){2-4}
 & Main model & Control & RL & [95\% CI] \\
\midrule
F1 & 0.436 $\pm$ 0.004 & 0.456 $\pm$ 0.005 & 0.469 $\pm$ 0.012 & +0.014 [+0.003, +0.025] \\
Lesion F1 & 0.388 $\pm$ 0.003 & 0.408 $\pm$ 0.006 & 0.432 $\pm$ 0.013 & +0.025 [+0.013, +0.037] \\
Dense F1 & 0.312 $\pm$ 0.005 & 0.330 $\pm$ 0.009 & 0.358 $\pm$ 0.022 & +0.028 [+0.013, +0.045] \\
F1 at IoU 0.75 & 0.289 $\pm$ 0.003 & 0.343 $\pm$ 0.003 & 0.334 $\pm$ 0.007 & $-$0.009 [$-$0.019, +0.001] \\
Region IoU & 0.578 $\pm$ 0.013 & 0.591 $\pm$ 0.003 & 0.607 $\pm$ 0.013 & +0.016 [+0.003, +0.028] \\
\midrule
Healthy refusal (\%) & 94.1 $\pm$ 0.9 & 96.8 $\pm$ 0.6 & 90.8 $\pm$ 2.6 & $-$6.0 [$-$8.0, $-$4.3] \\
False refusal (\%) & 4.9 $\pm$ 1.2 & 5.1 $\pm$ 0.2 & 2.0 $\pm$ 1.1 & $-$3.1 [$-$4.3, $-$2.0] \\
Recognition (\%) & 73.7 $\pm$ 1.1 & 74.4 $\pm$ 0.6 & 71.0 $\pm$ 1.5 & $-$3.4 [$-$4.9, $-$1.8] \\
Joint accuracy (\%) & 43.5 $\pm$ 1.2 & 45.2 $\pm$ 0.2 & 43.6 $\pm$ 1.4 & $-$1.5 [$-$3.4, +0.4] \\
\midrule
Wrong-name refusal (\%) & 28.6 $\pm$ 1.5 & 34.1 $\pm$ 2.3 & 15.3 $\pm$ 6.3 & -- \\
AUROC, wrong vs.\ true name & 0.812 $\pm$ 0.010 & 0.827 $\pm$ 0.002 & 0.753 $\pm$ 0.030 & -- \\
\bottomrule
\end{tabular}
\caption{\textbf{Three runs per group.} Main model: mixed + healthy (12\%), seeds 42, 1 and 2. Control and RL: three runs each, all started from the seed-42 run. Mean $\pm$ std: mean and sample standard deviation over the three runs (variation between training runs). RL $-$ control: difference of the group means with a 95\% bootstrap interval over images (every run scored on the same resampled images). Differences of percentages are in points. Means, standard deviations, and differences are computed from unrounded values, except the AUROC of the control and of RL, which is computed from the three values of Table~\ref{tab:s_rl_runs}. --: not computed. Metrics and protocols as in Table~\ref{tab:s_rl_runs}.}
\label{tab:s_rl_groups}
\end{table*}

\textbf{Three runs per group.} Table~\ref{tab:s_rl_groups} gives the mean and the standard deviation of the three runs of each group, and the difference between RL and the control with a bootstrap interval over images. Against the control, RL adds 0.014 F1 [$+$0.003, $+$0.025], 0.025 on lesion-level images, and 0.028 on dense images. At IoU 0.75 the two are level ($-$0.009 [$-$0.019, $+$0.001]). RL loses 3.4 recognition points and 6.0 points of healthy refusal against the control, and its false refusal is 3.1 points lower. Joint accuracy is not resolved ($-$1.5 points [$-$3.4, $+$0.4]). As the paper states, the first control run is above the starting model in F1 (paired $+$0.017 [$+$0.003, $+$0.031]). Part of the gain of RL over the starting model is therefore additional training on the rollout prompts.

\begin{table*}[t]
\centering\small
\setlength{\tabcolsep}{10pt}
\begin{tabular}{lccc}
\toprule
RL run $\downarrow$ \, minus \, control run $\rightarrow$ & First control run & Seed 1 & Seed 2 \\
\midrule
\multicolumn{4}{l}{\emph{F1}} \\
\quad First RL run & +0.019 [+0.004, +0.034] & +0.018 [+0.004, +0.033] & +0.027 [+0.011, +0.041] \\
\quad Seed 1 & $-$0.002 [$-$0.015, +0.011] & $-$0.004 [$-$0.017, +0.010] & +0.005 [$-$0.009, +0.019] \\
\quad Seed 2 & +0.019 [+0.004, +0.033] & +0.017 [+0.004, +0.031] & +0.026 [+0.011, +0.040] \\
\midrule
\multicolumn{4}{l}{\emph{Recognition (points)}} \\
\quad First RL run & $-$4.4 [$-$6.6, $-$2.2] & $-$5.7 [$-$7.8, $-$3.5] & $-$5.2 [$-$7.3, $-$3.1] \\
\quad Seed 1 & $-$2.0 [$-$4.2, +0.3] & $-$3.2 [$-$5.3, $-$0.9] & $-$2.8 [$-$4.9, $-$0.7] \\
\quad Seed 2 & $-$1.6 [$-$3.9, +0.5] & $-$2.9 [$-$5.1, $-$0.7] & $-$2.5 [$-$4.6, $-$0.3] \\
\midrule
\multicolumn{4}{l}{\emph{Healthy refusal (points)}} \\
\quad First RL run & $-$8.9 [$-$11.5, $-$6.4] & $-$8.5 [$-$11.2, $-$6.0] & $-$9.7 [$-$12.6, $-$7.0] \\
\quad Seed 1 & $-$4.2 [$-$6.5, $-$2.1] & $-$3.8 [$-$6.1, $-$1.6] & $-$5.0 [$-$7.4, $-$2.9] \\
\quad Seed 2 & $-$4.6 [$-$6.6, $-$2.8] & $-$4.2 [$-$6.2, $-$2.4] & $-$5.3 [$-$7.5, $-$3.5] \\
\bottomrule
\end{tabular}
\caption{\textbf{Every RL run minus every control run.} Paired differences on the same benchmark images with 95\% bootstrap intervals over images. F1: known-target protocol (973 positive images). Recognition: candidate protocol (973 positive images). Healthy refusal: 505 healthy images.}
\label{tab:s_rl_pairs}
\end{table*}

\textbf{Run by run.} Table~\ref{tab:s_rl_pairs} compares every RL run with every control run. In F1, RL is ahead in seven of the nine pairs, and in six of them the interval excludes zero. One RL run is level with the three control runs ($-$0.004 to $+$0.005). In recognition and in healthy refusal, every RL run is below every control run. This difference is resolved in seven of the nine pairs for recognition and in all nine for healthy refusal.

\textbf{Wrong names.} The three RL runs refuse 8.1\%, 19.6\%, and 18.2\% of the wrong names, against 30.2\% for the model they start from and 31.9\% to 36.5\% for the control runs (Table~\ref{tab:s_rl_runs}). Their AUROC for wrong against true names is 0.720 to 0.780, against 0.823 for the starting model and 0.824 to 0.828 for the control runs.

\begin{table*}[t]
\centering\small
\setlength{\tabcolsep}{5pt}
\begin{tabular}{@{}lcccccccc@{}}
\toprule
 & & Lesion & Recog. & Pos.-1 & Joint & \multicolumn{2}{c}{Refusal (\%)} & \\
\cmidrule(lr){7-8}
Configuration & F1 & F1 & (\%) & (\%) & (\%) & Healthy & False & AUROC \\
\midrule
Starting model (main model, seed-42 run) & 0.440 & 0.392 & 74.7 & 25.1 & 44.8 & 95.0 & 4.5 & 0.823 \\
\midrule
+ RL, first RL run (step 170) & 0.477 & 0.439 & 69.3 & 24.6 & 43.1 & 87.7 & 0.9 & 0.720 \\
+ RL, final checkpoint (step 292) & 0.467 & 0.440 & 67.2 & 24.7 & 39.0 & 89.3 & 6.4 & 0.686 \\
+ RL, recognition-weighted reward & 0.480 & 0.436 & 70.2 & 24.2 & 44.1 & 84.2 & 2.9 & 0.756 \\
+ RL, uniform rollout sampling & 0.440 & 0.409 & 70.2 & 26.1 & 40.9 & 88.1 & 4.8 & 0.733 \\
\midrule
+ continued SFT (control), first run & 0.458 & 0.410 & 73.7 & 24.3 & 45.2 & 96.6 & 5.1 & 0.828 \\
\bottomrule
\end{tabular}
\caption{\textbf{RL configurations.} Every RL configuration and the control start from the seed-42 run of the main model (first row). Each row is one training run. Columns as in Table 3 of the paper, plus the AUROC of the empty-answer probability for wrong against true names (973 positive images). Healthy and False: healthy refusal and false refusal. Pos.-1: in-list answers on the first candidate. Final checkpoint (step 292): the first RL run at its last training step.}
\label{tab:s_rl_configs}
\end{table*}

\textbf{Other RL configurations.} Table~\ref{tab:s_rl_configs} adds three configurations to the main one. The final checkpoint is the first RL run at the end of its second pass over the prompts. The recognition-weighted reward moves weight from the localization and instance terms to the name term (0.30, 0.10, and 0.35 in place of 0.35, 0.15, and 0.25) and is trained for one pass. Uniform rollout sampling draws the positive prompts without the preference for images with four or more boxes and is also trained for one pass. With uniform sampling, F1 stays at 0.440, the value of the starting model. With the recognition-weighted reward, recognition is 70.2\%, against 69.3\% for the first RL run and 74.7\% for the starting model. The AUROC of these three configurations lies between 0.686 and 0.756.

\textbf{Joint accuracy.} Every RL configuration starts from the seed-42 run of the main model, which has joint accuracy 44.8\%. The main RL configuration reaches 43.1\% in its first run, the run that the comparisons of the paper use, and 43.6\% on average over its three runs (Table~\ref{tab:s_rl_groups}). The final checkpoint reaches 39.0\%, the recognition-weighted reward 44.1\%, and uniform sampling 40.9\%. None of these values exceeds the 44.8\% of the starting model, as the paper states.

\clearpage\clearpage
\section{Held-Out Sources}
\label{supp:heldout}

This section gives the details of the held-out-source experiment of Section~5.3 of the paper. The protocols and the metrics are defined in \suppref{supp:protocols}.

\begin{table}[t]
\centering\footnotesize
\setlength{\tabcolsep}{6pt}
\begin{tabular}{@{}lrrr@{}}
\toprule
 & \multicolumn{3}{c}{Benchmark images} \\
\cmidrule(l){2-4}
Held-out source & All & Positive & Healthy \\
\midrule
CDDM & 299 & 79 & 220 \\
LeafNet & 512 & 342 & 170 \\
MIRAGE & 289 & 289 & 0 \\
AgroBench & 174 & 174 & 0 \\
LeafBench & 122 & 20 & 102 \\
AgroMind & 42 & 32 & 10 \\
AgMMU & 35 & 35 & 0 \\
\midrule
Seven folds & 1,473 & 971 & 502 \\
\bottomrule
\end{tabular}
\caption{\textbf{Held-out-source folds.} Each fold retrains the main model without any training example of one source. Benchmark images: the images of the held-out source, on which the fold is scored.}
\label{tab:s_heldout_folds}
\end{table}

\textbf{Design.} To measure transfer beyond the training sources, the main model is retrained seven times, each time without one source (Table~\ref{tab:s_heldout_folds}). A fold removes from the training mixture of the main model every example of its source, including the healthy negatives that come from this source. All other examples are kept. The eighth source, AgroCoT, has only five benchmark images and therefore no fold. Each fold is one training run with the recipe of the main model (\suppref{supp:training}), and the final weights of each run are used.

\textbf{Scoring.} Each fold model is evaluated with both protocols on the benchmark images of its held-out source only. The seven sources have 1,473 benchmark images, 971 positive and 502 healthy. Every pooled value of the held-out models scores each of these images with the fold model that never saw its source. The wrong names are the same as for the main model. The comparison is the seed-42 run of the main model trained on all sources (in source), on the same images, and zero-shot Qwen3-VL-2B. Intervals are 95\% bootstrap intervals over images (1,000 resamples), and differences between models are paired.

\textbf{Comparison values on 1,473 images.} The 1,473 images are the benchmark without the five AgroCoT images. The values of the two comparison models therefore differ slightly from those of Table~2 of the paper, which uses all 1,478 images. For the seed-42 run of the main model, F1 is 0.440 on both sets. On the 1,473 images, recognition is 74.9\% against 74.7\%, and joint accuracy is 44.9\% against 44.8\%. Healthy refusal is 95.2\% against 95.0\%, false refusal 4.4\% against 4.5\%, and wrong-name refusal 30.1\% against 30.2\%. The AUROC for wrong against true names is 0.824 against 0.823 (Table~4 of the paper). For zero-shot Qwen3-VL-2B, F1 is 0.319 and healthy refusal 1.2\% on both sets. Its recognition is 51.6\% against 51.7\%, and its joint accuracy 34.2\% against 34.3\%.

\begin{table*}[t]
\centering\footnotesize
\setlength{\tabcolsep}{4pt}
\begin{tabular}{@{}lccc@{}}
\toprule
Model & F1 [95\% CI] & Recognition (\%) & Joint (\%) \\
\midrule
Qwen3-VL-2B, zero-shot & 0.319 [0.288, 0.354] & 51.6 & 34.2 \\
Main model (all sources) & 0.440 [0.411, 0.473] & 74.9 & 44.9 \\
Held-out models & 0.424 [0.395, 0.456] & 56.1 & 35.8 \\
\midrule
\emph{Held out $-$ main model} & $-$0.016 [$-$0.031, $-$0.001] & $-$18.7 [$-$21.6, $-$15.7] & $-$9.1 [$-$11.8, $-$6.2] \\
\bottomrule
\end{tabular}

\vspace{4pt}
\begin{tabular}{@{}lcccc@{}}
\toprule
Model & Healthy ref.\ (\%) & False ref.\ (\%) & Wrong-name ref.\ (\%) & AUROC \\
\midrule
Qwen3-VL-2B, zero-shot & 1.2 [0.4, 2.2] & 0.0 [0.0, 0.0] & 0.0 [0.0, 0.0] & 0.706 [0.690, 0.721] \\
Main model (all sources) & 95.2 [93.2, 97.0] & 4.4 [3.2, 5.9] & 30.1 [27.3, 33.0] & 0.824 [0.808, 0.840] \\
Held-out models & 91.4 [89.0, 93.8] & 12.2 [10.2, 14.3] & 39.1 [36.4, 42.1] & 0.725 [0.707, 0.743] \\
\midrule
\emph{Held out $-$ main model} & $-$3.8 [$-$6.0, $-$1.6] & +7.7 [+5.9, +9.7] & +9.1 [+6.1, +12.0] & $-$0.098 [$-$0.112, $-$0.086] \\
\bottomrule
\end{tabular}
\caption{\textbf{Held-out sources, pooled over the seven folds.} The 1,473 benchmark images of the seven sources, which are the benchmark without the five AgroCoT images. Held-out models: every image is scored by the model that was trained without its source. Main model: the seed-42 run. The values of the main model and of the zero-shot model are recomputed on these images, so some differ slightly from Tables~2 and~4 of the paper. Healthy refusal (Healthy ref.): the 502 healthy images. All other columns: the 971 positive images. F1, healthy refusal, and false refusal (False ref.): known-target protocol. Recognition and joint accuracy: candidate protocol. Wrong-name ref.: positive images refused when queried with a wrong name. AUROC: empty-answer probability, wrong against true names. Intervals are 95\% bootstrap intervals over images. Last row: paired difference on the same images, in points for the percentages.}
\label{tab:s_heldout_pooled}
\end{table*}

\textbf{Results.} Table~\ref{tab:s_heldout_pooled} gives the pooled results. Pooled over the 1,473 images, F1 is 0.424 for the held-out models against 0.440 in source (paired $-$0.016 [$-$0.031, $-$0.001]) and 0.319 zero-shot. As the paper states, this is about seven eighths of the gain over zero-shot. Recognition is 56.1\% against 74.9\% (paired $-$18.7 [$-$21.6, $-$15.7]) and 51.6\% zero-shot. As the paper states, recognition keeps about one fifth of its gain. Joint accuracy is 35.8\% against 44.9\% (paired $-$9.1 [$-$11.8, $-$6.2]) and 34.2\% zero-shot. The held-out models refuse 91.4\% of the healthy images, against 95.2\% in source (paired $-$3.8 [$-$6.0, $-$1.6]) and 1.2\% zero-shot. The true name is refused on 12.2\% of the positive images, against 4.4\% (paired +7.7 [+5.9, +9.7]). The wrong name is refused on 39.1\%, against 30.1\% (paired +9.1 [+6.1, +12.0]). The AUROC of the empty-answer probability for wrong against true names is 0.725 against 0.824 (paired $-$0.098 [$-$0.112, $-$0.086]) and 0.706 zero-shot. As the paper states, the two names are separated about as well as zero-shot.

\clearpage\clearpage
\section{External Benchmarks}
\label{supp:external}

This section gives the details of the external benchmarks of Section~5.3 of the paper: PlantSeg, PlantDoc, AgroVG, and GroundingME. It covers Table~5 of the paper and the PlantSeg column of Table~2 of the paper. The held-out sources of the same section of the paper are in \suppref{supp:heldout}.

\textbf{Models.} Every benchmark is scored with checkpoints of Table~2 of the paper: Qwen3-VL-2B and Qwen3-VL-8B zero-shot, the main model (mixed + healthy (12\%), seed-42 run), the first RL run, and Ours 8B. In Table~5 of the paper, Ours is the main model. Some tables add further configurations.

\textbf{PlantSeg: data.} PlantSeg \citep{wei2024plantseg} is an expert-reviewed localization set with fixed classes and no language. We score the images of its released test split. Each expert mask is split into connected components, and components whose box covers less than 0.1\% of the image are dropped. If the largest remaining box covers more than half of the image, the image is at object level and gets one box around the whole mask. Otherwise it is at lesion level and keeps the boxes of its largest components, at most 30. Images with an empty mask or without a remaining component are skipped. Images whose 64-bit perceptual hash lies within Hamming distance 6 of a training image of our corpus are removed as near-duplicates. This leaves 2,165 test images. All of them are positive images, so refusal is not measured on PlantSeg.

\textbf{PlantSeg: protocol.} The models are queried in the known-target protocol with PlantSeg's disease name and the instruction of the image's level (\suppref{supp:protocols}). GroundingDINO uses the thresholds of the annotation run (box 0.25, text 0.20). It is prompted with the raw disease name, or with the taxonomy prompt of the disease where one exists (363 images). The teacher row applies the box filters of the annotation pipeline to the predictions with taxonomy prompts. Scores use our scorer and the metrics of the benchmark (\suppref{supp:protocols}), with 95\% bootstrap intervals over images (1,000 resamples).

\begin{table}[t]
\centering\footnotesize
\setlength{\tabcolsep}{4pt}
\begin{tabular}{@{}lc@{}}
\toprule
Model & F1 [95\% CI] \\
\midrule
GroundingDINO, raw prompts & 0.182 [0.172, 0.194] \\
GroundingDINO, taxonomy prompts & 0.183 [0.173, 0.194] \\
GroundingDINO, pipeline configuration (teacher) & 0.161 [0.150, 0.173] \\
\midrule
Qwen3-VL-2B, zero-shot & 0.164 [0.152, 0.177] \\
Qwen3-VL-8B, zero-shot & 0.215 [0.201, 0.231] \\
\midrule
Ours 2B, main model & 0.231 [0.219, 0.245] \\
Ours 2B, + RL, first RL run & 0.282 [0.269, 0.296] \\
Ours 8B, mixed + healthy & 0.237 [0.224, 0.251] \\
\midrule
2B, PlantSeg expert masks & 0.251 [0.238, 0.266] \\
\bottomrule
\end{tabular}
\caption{\textbf{PlantSeg test set, 2,165 images.} Known-target protocol with PlantSeg's disease names. Reference boxes come from the connected components of the expert masks. F1: instance F1 at IoU 0.5 pooled over the images, with a 95\% bootstrap interval over images. Main model: mixed + healthy (12\%), seed-42 run. The last row is a 2B model trained on 9,120 examples of PlantSeg's training split with its expert masks (\suppref{supp:scale}). Paired differences in F1 with 95\% bootstrap intervals: RL minus main model +0.051 [+0.042, +0.059]. RL minus teacher +0.121 [+0.109, +0.133]. RL minus the model trained on the expert masks +0.031 [+0.018, +0.042]. RL minus GroundingDINO with taxonomy prompts +0.099 [+0.089, +0.110].}
\label{tab:s_plantseg}
\end{table}

\textbf{PlantSeg: results.} Table~\ref{tab:s_plantseg} gives the scores. Its F1 values for the models of Table~2 of the paper are those of the PlantSeg column of that table. As the paper states, instance F1 transfers: the RL model reaches 0.282 against 0.161 to 0.183 for GroundingDINO and 0.164 and 0.215 for the zero-shot models. A 2B model trained on PlantSeg's own masks reaches 0.251 (RL minus this model: paired +0.031 [+0.018, +0.042]), and the main model 0.231. The region IoU of our models (0.367 to 0.463) stays below that of the zero-shot models (0.497 and 0.547). On our benchmark, the model trained on the expert masks is compared with a model trained on an equal-size subset of our boxes in \suppref{supp:scale}.

\textbf{PlantDoc: data.} PlantDoc \citep{singh2020plantdoc} consists of web photographs with boxes drawn around whole leaves. Its classes are 18 diseases and 11 healthy leaf classes. We use both official splits. Images with two classes or with a recorded size that differs from the image file are dropped, and boxes are clipped to the image. An image with a disease becomes an object-level item whose reference boxes are all its leaf boxes. A healthy image becomes a target-absent item and is queried with a disease associated with its crop. The diseases carry their standard names in place of the PlantDoc class strings. Near-duplicates of training images and of other PlantDoc images are removed with the rule of the benchmark (64-bit perceptual hash, Hamming distance at most 6). After this, 986 images remain: 431 positive images and 555 healthy images. Of the positive images, 273 show one leaf and 158 several leaves, with 2.16 leaf boxes per image on average.

\textbf{PlantDoc: protocol.} Localization and refusal use the known-target protocol with our object-level instruction, as Table~5 of the paper states. Recognition and joint accuracy use the candidate protocol with our object-level candidate instruction (\suppref{supp:protocols}). The three other candidates are drawn in this order: PlantDoc diseases of the same crop, our training targets of that crop with at least 100 examples, and PlantDoc diseases of other crops. The true disease is at positions 1 to 4 on 121, 107, 93, and 110 images. All values use our scorer, with 95\% bootstrap intervals over images (1,000 resamples).

\begin{table*}[t]
\centering\small
\setlength{\tabcolsep}{5pt}
\begin{tabular}{@{}lccccc@{}}
\toprule
 & & Healthy & False & Recognition (\%) & Joint (\%) \\
Model & F1 [95\% CI] & refusal (\%) & refusal (\%) & [95\% CI] & [95\% CI] \\
\midrule
Qwen3-VL-2B, zero-shot & 0.512 [0.461, 0.560] & 0.5 & 0.0 & 44.1 [39.2, 48.7] & 28.8 [24.4, 33.0] \\
Ours 2B, grounding only & 0.418 [0.369, 0.471] & 0.0 & 0.0 & 27.4 [23.2, 31.6] & 16.9 [13.5, 20.6] \\
Ours 2B, mixed & 0.431 [0.383, 0.483] & 0.0 & 0.0 & 52.7 [48.0, 57.3] & 39.0 [34.3, 43.4] \\
Ours 2B, main model & 0.413 [0.365, 0.463] & 68.8 & 1.9 & 52.0 [47.6, 57.1] & 38.3 [33.9, 42.9] \\
Ours 2B, + RL & 0.449 [0.402, 0.502] & 41.8 & 0.7 & 48.0 [43.6, 52.9] & 36.9 [32.7, 41.5] \\
\midrule
Qwen3-VL-8B, zero-shot & 0.570 [0.516, 0.623] & 51.5 & 0.2 & 57.5 [52.7, 61.9] & 46.9 [42.0, 51.5] \\
Ours 8B, mixed + healthy & 0.450 [0.400, 0.503] & 71.4 & 3.5 & 50.1 [45.7, 55.0] & 37.6 [33.2, 42.2] \\
\bottomrule
\end{tabular}
\caption{\textbf{PlantDoc, 986 images (431 positive, 555 healthy).} F1: known-target protocol with the object-level instruction, instance F1 at IoU 0.5 pooled over the 431 positive images. Healthy refusal: share of the 555 healthy images answered with an empty list. False refusal: share of the positive images on which the true target is refused. Recognition and Joint: candidate protocol on the 431 positive images. Intervals are 95\% bootstrap intervals over images. Main model: mixed + healthy (12\%), seed-42 run. + RL: first RL run.}
\label{tab:s_plantdoc}
\end{table*}

\textbf{PlantDoc: results.} Table~\ref{tab:s_plantdoc} gives the scores. As the paper states, the fine-tuned models localize worse than their zero-shot bases. The paired difference in F1 is $-$0.099 [$-$0.135, $-$0.066] for the main model and $-$0.121 [$-$0.158, $-$0.086] for Ours 8B. They answer with one box per image, as the object-level training examples do, and lose on images with several leaves. On the 158 images with several leaves, the grounding-only, main, RL, and 8B models give 0.97 to 1.06 boxes per image, against 1.68 and 1.83 for the zero-shot models. The refusal of healthy images largely transfers: 68.8\% with the main model against 0.5\% zero-shot, and 41.8\% after RL. Candidate recognition rises at 2B only (+7.9 points [1.6, 14.8] for the main model over its base), as does joint accuracy (paired +9.5 [3.7, 15.5]). The grounding-only model reaches 27.4\%, against 52.7\% for the mixed model and 44.1\% zero-shot. It names the first listed candidate in 354 of its 374 answers that name a listed candidate.

\textbf{AgroVG: data.} AgroVG \citep{li2026agrovg} is an evaluation benchmark without a training split. We use the test split of its task T1, with 1,288 queries from six target families. Of these, 621 queries have one target, 468 several targets, and 199 no target. The families are crop and weed (233 queries), fruit (245), wheat head (202), pest (251), plant disease (242), and forest and tree cover (115). A target-absent query asks for a target of another family on the same image.

\textbf{AgroVG: protocol.} Every model receives the official instruction of the benchmark. Its answers are converted and scored with the official code of the benchmark, unchanged. Decoding is greedy with at most 256 new tokens. The scorer matches predicted and reference boxes by maximum-cardinality bipartite matching at IoU 0.5 and computes a Set-F1 per query. A target-absent query answered with no box scores 1. All@0.5 is the mean over the queries.

\begin{table*}[t]
\centering\footnotesize
\setlength{\tabcolsep}{4pt}
\begin{tabular}{@{}lcccc@{}}
\toprule
 & All@0.5 & Present & Absent & No box, \\
Model & [95\% CI] & @0.5 & acc. & present \\
\midrule
Qwen3-VL-2B, zero-shot & 0.248 [0.227, 0.269] & 0.170 & 0.673 & 0.089 \\
\midrule
Grounding only & 0.208 [0.190, 0.228] & 0.245 & 0.010 & 0.007 \\
\midrule
Mixed + healthy (6\%) & 0.260 [0.239, 0.283] & 0.220 & 0.482 & 0.071 \\
Mixed + healthy (12\%), main model, seed 42 & 0.256 [0.233, 0.280] & 0.207 & 0.523 & 0.085 \\
\quad main model, seed 1 & 0.239 [0.218, 0.260] & 0.224 & 0.322 & 0.058 \\
\quad main model, seed 2 & 0.202 [0.183, 0.223] & 0.211 & 0.156 & 0.032 \\
\midrule
\quad + wrong-target 2.9\% & 0.271 [0.249, 0.293] & 0.205 & 0.633 & 0.096 \\
\quad + wrong-target 5.6\% & 0.258 [0.235, 0.280] & 0.155 & 0.819 & 0.466 \\
\quad + wrong-target 10.6\% & 0.291 [0.269, 0.314] & 0.205 & 0.764 & 0.153 \\
Wrong-target 11.9\% only & 0.269 [0.246, 0.292] & 0.149 & 0.925 & 0.519 \\
\midrule
+ continued SFT (control), first run & 0.248 [0.227, 0.270] & 0.213 & 0.442 & 0.086 \\
\quad control, seed 1 & 0.252 [0.231, 0.275] & 0.221 & 0.422 & 0.078 \\
\quad control, seed 2 & 0.256 [0.235, 0.278] & 0.227 & 0.417 & 0.081 \\
\midrule
+ RL, first run (step 170) & 0.226 [0.204, 0.246] & 0.243 & 0.131 & 0.017 \\
\quad + RL, seed 1 & 0.235 [0.213, 0.256] & 0.223 & 0.296 & 0.049 \\
\quad + RL, seed 2 & 0.225 [0.206, 0.245] & 0.236 & 0.161 & 0.017 \\
\midrule
Qwen3-VL-8B, zero-shot & 0.329 [0.305, 0.353] & 0.226 & 0.889 & 0.271 \\
Ours 8B, mixed + healthy & 0.205 [0.186, 0.226] & 0.217 & 0.141 & 0.023 \\
\bottomrule
\end{tabular}

\caption{\textbf{AgroVG task T1, test split (1,288 queries, 1,089 with the target present and 199 with the target absent).} All@0.5: macro Set-F1 at IoU 0.5 over all queries, with a 95\% bootstrap interval over queries. Present@0.5: the same on the queries whose target is present. Absent acc.: share of the 199 target-absent queries answered with no box. No box, present: share of the 1,089 target-present queries answered with no box. The answers are converted and scored with the official code of the benchmark. Wrong-target rows: the configurations with wrong-target negatives of Table~4 of the paper. The rows between the two zero-shot rows are our 2B models. Main model: mixed + healthy (12\%), seed-42 run. Seed 1 and seed 2: further training runs (\suppref{supp:rl}). Shares are given as fractions.}
\label{tab:s_agrovg_t1}
\end{table*}

\begin{figure}[t]
\centering
\includegraphics[width=\linewidth]{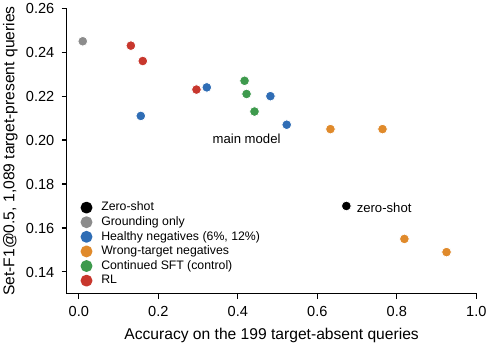}
\caption{\textbf{AgroVG T1, the runs of the 2B models.} Set-F1 at IoU 0.5 on the 1,089 target-present queries against accuracy on the 199 target-absent queries, for the zero-shot Qwen3-VL-2B and the 15 runs of our 2B models in Table~\ref{tab:s_agrovg_t1}. Numbers as in that table.}
\label{fig:s_external}
\end{figure}

\textbf{AgroVG: results.} Table~\ref{tab:s_agrovg_t1} gives the runs on T1. The main model is at most level with its base (paired +0.008 [$-$0.012, +0.027]), and its other two seeds are at $-$0.009 and $-$0.045. Ours 8B is 0.123 below its base [$-$0.149, $-$0.099]. The three runs of the main model gain on present targets (paired +0.037 to +0.054) and lose on absent ones. As the paper states, accuracy on absent targets follows the training mixture (Figure~\ref{fig:s_external}). It is 0.673 zero-shot, 0.010 with grounding only, 0.156 to 0.523 with healthy negatives, 0.633 to 0.925 with wrong-name negatives, and 0.131 to 0.296 after RL (control: 0.417 to 0.442). The wrong-name negatives are those of the wrong-target rows of Table~\ref{tab:s_agrovg_t1}. The 10.6\% wrong-name model gains on both (overall +0.044 [+0.025, +0.061]). Two of the other three return no box on 47 to 52\% of the present targets: + wrong-target 5.6\% on 507 of the 1,089 (0.466) and wrong-target 11.9\% only on 565 (0.519). The third, + wrong-target 2.9\%, returns no box on 104 (0.096).

\textbf{GroundingME: data and protocol.} GroundingME \citep{li2025groundingme} is a general-domain grounding benchmark of 1,005 samples. Its Discriminative (204), Spatial (300), and Limited (300) samples have a target. Its 201 Rejection samples have no target. Every model receives the official prompt template, and its answers are scored with the official functions of the benchmark, unchanged. Decoding is greedy with at most 2,048 new tokens. The metric is accuracy at IoU 0.5.

\begin{table}[t]
\centering\footnotesize
\setlength{\tabcolsep}{4pt}
\begin{tabular}{@{}lcc@{}}
\toprule
 & Acc@0.5 (\%) & $\Delta$ to base \\
Model & [95\% CI] & [95\% CI] \\
\midrule
Qwen3-VL-2B, zero-shot & 21.2 [18.9, 23.8] & -- \\
Ours 2B, main model & 8.7 [7.0, 10.6] & $-$12.5 [$-$14.9, $-$10.1] \\
Ours 2B, + RL & 8.4 [6.7, 10.1] & $-$12.8 [$-$15.4, $-$10.3] \\
\midrule
Qwen3-VL-8B, zero-shot & 31.0 [28.1, 33.9] & -- \\
Ours 8B, mixed + healthy & 14.6 [12.5, 16.8] & $-$16.4 [$-$19.0, $-$13.9] \\
\bottomrule
\end{tabular}
\caption{\textbf{GroundingME, 1,005 samples.} Accuracy at IoU 0.5 with the official prompt template and the official scoring functions. $\Delta$ to base: paired difference in points to the zero-shot model of the same size, with a 95\% interval. The zero-shot models are the bases. Main model: mixed + healthy (12\%), seed-42 run. + RL: first RL run.}
\label{tab:s_groundingme}
\end{table}

\textbf{GroundingME: results.} Table~\ref{tab:s_groundingme} gives the accuracies. Accuracy falls at 2B (paired $-$12.5 points [$-$14.9, $-$10.1] for the main model) and at 8B ($-$16.4 [$-$19.0, $-$13.9]).

\end{document}